\documentclass{article}
\pdfoutput=1
     \PassOptionsToPackage{sort,compress,numbers}{natbib}
     \usepackage[final]{neurips_2022}

\usepackage[utf8]{inputenc} % allow utf-8 input
\usepackage[T1]{fontenc}    % use 8-bit T1 fonts
\usepackage{url}            % simple URL typesetting
\usepackage{booktabs}       % professional-quality tables
\usepackage{amsfonts}       % blackboard math symbols
\usepackage{nicefrac}       % compact symbols for 1/2, etc.
\usepackage{microtype}      % microtypography
\usepackage[usenames,dvipsnames]{xcolor}         % colors
\definecolor{citecolor}{HTML}{0071BC}
\definecolor{linkcolor}{HTML}{ED1C24}
\usepackage[pagebackref=false, breaklinks=true, colorlinks, citecolor=citecolor, linkcolor=linkcolor, bookmarks=false]{hyperref}
\usepackage{todonotes}

\usepackage{graphicx}
\usepackage{custom_math}
\usepackage{subcaption}
\usepackage{cleveref}
\usepackage{enumitem}
\usepackage{multirow}
\usepackage{wrapfig}
\usepackage{thmtools}
\usepackage{thm-restate}
\usepackage{titletoc}
\usepackage[page,header]{appendix}
\usepackage{thmtools}
\usepackage{thm-restate}
\usepackage{placeins}
\usepackage[usestackEOL]{stackengine}

\newcommand{\BaseClassifier}{f}
\newcommand{\DecisionClassifier}{F}
\newcommand{\RandomParameter}{\bm{\theta}}
\newcommand{\InputTransformation}{t}
\newcommand{\DataDimension}{d}
\newcommand{\NumberOfClasses}{C}
\newcommand{\PredictionTime}{n}
\newcommand{\EOTSamples}{m}
\newcommand{\PGDSteps}{k}
\newcommand{\DataSpace}{\mathcal{X}}
\newcommand{\DataSpaceExpand}{[0,1]^\DataDimension}
\newcommand{\LabelSpace}{\mathcal{Y}}
\newcommand{\LabelSpaceExpand}{[\NumberOfClasses]}
\newcommand{\LogitsSpace}{\RR^\NumberOfClasses}
\newcommand{\RandomizationSpace}{\Theta}
\newcommand{\bx}{\bm{x}}
\newcommand{\y}{y}
\newcommand{\AdversarialPerturbation}{\bm{\delta}}
\newcommand{\TargetClass}{y^\prime}
\newcommand{\RandomizedClassifier}[1][\RandomParameter]{\BaseClassifier_{#1}}
\newcommand{\RandomizedClassifierOutputOf}[2][\RandomParameter]{\BaseClassifier_{{#1}, {#2}}}

\newcommand{\RandomizedDecisionClassifier}[1][\RandomParameter]{\DecisionClassifier_{#1}}

\newtheorem{observation}{Observation}
\newtheorem{assumption}{Assumption}
\title{On the Limitations of \\
Stochastic Pre-processing Defenses}

\author{%
  Yue Gao \\
  University of Wisconsin--Madison \\
  \texttt{gy@cs.wisc.edu} \\
  \And
  Ilia Shumailov \\
  University of Cambridge \&
  Vector Institute \\
  \texttt{ilia.shumailov@cl.cam.ac.uk} \\
  \And
  Kassem Fawaz \\
  University of Wisconsin--Madison \\
  \texttt{kfawaz@wisc.edu} \\
  \And
  Nicolas Papernot \\
  University of Toronto \&
  Vector Institute \\
  \texttt{nicolas.papernot@utoronto.ca}
}

\begin{document}

\maketitle

\addtocontents{toc}{\protect\setcounter{tocdepth}{0}}  % disable toc for main body

\begin{abstract}

Defending against adversarial examples remains an open problem. A common belief is that randomness at inference increases the cost of finding adversarial inputs. An example of such a defense is to apply a random transformation to inputs prior to feeding them to the model. In this paper, we empirically and theoretically investigate such stochastic pre-processing defenses and demonstrate that they are flawed. First, we show that most stochastic defenses are weaker than previously thought; they lack sufficient randomness to withstand even standard attacks like projected gradient descent. This casts doubt on a long-held assumption that stochastic defenses invalidate attacks designed to evade deterministic defenses and force attackers to integrate the Expectation over Transformation (EOT) concept. Second, we show that stochastic defenses confront a trade-off between adversarial robustness and model invariance; they become less effective as the defended model acquires more invariance to their randomization. Future work will need to decouple these two effects. We also discuss implications and guidance for future research.

\end{abstract}

\section{Introduction}
\label{sec:intro}

% adversarial example, defenses, hard to evaluate
Machine learning models are vulnerable to adversarial examples~\cite{adversarial-example-1,adversarial-example-2}, where an adversary can add imperceptible perturbations to the input of a model and change its prediction~\cite{pgd,cw}. Their discovery has motivated a wide variety of defense approaches~\cite{input-transformation,random-rescaling,activation-pruning,kwinners,odds,smoothing} along with the evaluation of their adversarial robustness~\cite{bpda,adaptive,bart}. Current evaluations mostly rely on adaptive attacks~\cite{bpda,adaptive}, which require significant modeling and computational efforts. However, even when the attack succeeds, such evaluations may not always reveal the fundamental weaknesses of an examined defense. Without awareness of the underlying weaknesses, subsequent defenses may still conduct inadvertently weak adaptive attacks; this leads to overestimated robustness.

% stochastic defenses, attack-driven, hard to evaluate, require effort, bart
One popular class of defenses that demonstrates the above is the stochastic pre-processing defense, which relies on applying randomized transformations to inputs to provide robustness~\cite{input-transformation,random-rescaling}. Despite existing attack techniques designed to handle randomness~\cite{bpda,eot}, there is an increasing effort to improve these defenses through a larger randomization space or more complicated transformations. For example, BaRT~\cite{bart} employs 25 transformations, where the parameters of each transformation are further randomized. Due to the complexity of this defense, it was only broken recently (three years later) by \citet{aggmopgd} with a complicated adaptive attack. Still, it is unclear how future defenses can avoid the pitfalls of existing defenses, largely because these pitfalls remain unknown.

% different step, fundamental limitations
% \paragraph{Contributions.}
In this paper, we investigate stochastic pre-processing defenses and explain their limitations both empirically and theoretically. First, we revisit previous stochastic pre-processing defenses and explain why such defenses are broken. We show that most stochastic defenses are not sufficiently randomized to invalidate standard attacks designed for deterministic defenses. Second, we study recent stochastic defenses that exhibit more randomness and show that they also face key limitations. In particular, we identify a trade-off between their robustness and the model's invariance to their transformations. These defenses achieve a notion of robustness that results from reducing the model's invariance to the applied transformations. We outline our findings below. These findings suggest future work to find new ways of using randomness that decouples these two effects.

% first, not random enough
\textbf{Most stochastic defenses lack sufficient randomness.}
While \citet{bpda} and \citet{adaptive} have demonstrated the ineffectiveness of several stochastic defenses with techniques like Expectation over Transformation (EOT)~\cite{eot}, it remains unclear whether and why EOT is required (or at least as a ``standard technique'') to break them. A commonly accepted explanation is that EOT computes the ``correct gradients'' of models with randomized components~\cite{bpda,adaptive}, yet the necessity of such correct gradients has not been explicitly discussed. To fill this gap, we examine a long-held assumption that stochastic defenses invalidate standard attacks designed for deterministic defenses.

Specifically, we revisit stochastic pre-processing defenses previously broken by EOT and examine their robustness \emph{without} applying EOT. Interestingly, we find that most stochastic defenses lack sufficient randomness to withstand even standard attacks (that do not integrate any strategy to capture model randomness) like projected gradient descent (PGD)~\cite{pgd}. We then conduct a systematic evaluation to show that applying EOT is only beneficial when the defense is sufficiently randomized. Otherwise, standard attacks already perform well and the randomization's robustness is overestimated.

% second, trade-off, (differ from known trade-off), theoretical model
\textbf{Trade-off between adversarial robustness and model invariance.}
When stochastic pre-processing defenses do have sufficient randomness, they must fine-tune the model using augmented training data to preserve utility in the face of randomness added. We characterize this procedure by the model's \emph{invariance} to the applied defense, where we identify a trade-off between the model's robustness (provided by the defense) and its invariance to the applied defense. Stochastic pre-processing defenses become less effective when their defended model acquires more invariance to their transformations.

On the theoretical front, we present a theoretical setting where this trade-off provably exists. We show from this trade-off that stochastic pre-processing defenses provide robustness by inducing variance on the defended model, and must take back such variance to recover utility. We verify this trade-off with empirical evaluations on realistic datasets, models, and defenses. We observe that robustness drops when the defended model is fine-tuned on data processed by its defense to acquire higher invariance.

\section{Related Work}
\label{sec:related}

\textbf{Stochastic Pre-processing Defenses.}
Defending against adversarial examples remains an open problem, where a common belief is that inference-time randomness increases the cost of finding adversarial inputs. Early examples of such stochastic defenses include input transformations~\cite{input-transformation} and rescaling~\cite{random-rescaling}. These defenses were broken by \citet{bpda} using techniques like EOT~\cite{eot} to capture randomness. After that, more stochastic defenses were proposed but with inadvertently weak evaluations~\cite{kwinners,odds,mixup,menet}, which were found ineffective by \citet{adaptive}. Subsequent stochastic defenses resort to larger randomization space like BaRT~\cite{bart}, which was only broken recently by \citet{aggmopgd}. In parallel to our work, DiffPure~\cite{diffpure} adopts a complicated stochastic diffusion process to purify the inputs. As we will discuss in \Cref{app:discuss:diffpure}, this defense belongs to an existing line of research that leverages generative models to pre-process input images~\cite{defense-gan,pixeldefend,gen-robustness}, hence it matches the settings in our work. On the other hand, randomized smoothing~\cite{smoothing,smoothing2,smoothing3} leverages randomness to certify the inherent robustness of a given decision. In this work, instead of designing adaptive attacks for individual defenses, which is a well-known challenging progress~\cite{bpda,adaptive,aggmopgd}, we focus on the general stochastic pre-processing defenses and demonstrate their limitations.

\textbf{Trade-offs for Adversarial Robustness.}
The trade-offs associated with adversarial robustness have been widely discussed in the literature. For example, prior work identified trade-offs between robustness and accuracy~\cite{tradeoff-rob-acc-1,tradeoff-rob-acc-2} for deterministic classifiers. \citet{pinot2022robustness} generalize this trade-off to randomized classifiers with a similar form as randomized smoothing. Compared with these results, our work provides a deeper understanding that stochastic pre-processing defenses \emph{explicitly} control such trade-offs to provide robustness. Recent work also investigated the trade-off between the model's robustness and invariance to input transformations, such as circular shifts~\cite{shift-invariance} and rotations~\cite{spatial-tradeoff}. These trade-offs characterize a standalone model's own property --- the model itself is less robust to adversarial examples when it becomes more invariant to certain transformations, without any defense. Our setting, however, is orthogonal to such analysis --- the model that we consider is protected by a stochastic pre-processing defense, and what we really aim to characterize is the performance of that pre-processing defense, not the inherent robustness of the model itself.

%and even between the robustness against different types of adversarial examples~\cite{invariance-adv1,invariance-adv2}. 

\section{Preliminaries}
\label{sec:prelim}

\paragraph{Notations.} Let $\BaseClassifier:\DataSpace\to\LogitsSpace$ denote the classifier with pre-softmax outputs, where $\DataSpace=\DataSpaceExpand$ is the input space with $\DataDimension$ dimensions and $\NumberOfClasses$ is the number of classes. We then consider a stochastic pre-processing defense $\InputTransformation_{\RandomParameter}:\DataSpace\to\DataSpace$, where $\RandomParameter$ is the random variable drawn from some randomization space $\RandomizationSpace$ that parameterizes the defense. The defended classifier can be written as $\RandomizedClassifier(\bx)\coloneqq\BaseClassifier(\InputTransformation_{\RandomParameter}(\bx))$.

Let $\DecisionClassifier(\bx)\coloneqq\argmax_{i\in\LabelSpace}\BaseClassifier_i(\bx)$ denote the classifier that returns the predicted label, where $\BaseClassifier_i$ is the output of the $i$-th class and $\LabelSpace=\LabelSpaceExpand$ is the label space. Similarly, we use $\RandomizedDecisionClassifier$ and $\RandomizedClassifierOutputOf{i}$ to denote the prediction and class-output of the stochastic classifier $\RandomizedClassifier$. Since this classifier returns varied outputs for a fixed input, it determines the final prediction by aggregating $\PredictionTime$ independent inferences with strategies like majority vote. We discuss these strategies and the choice of $\PredictionTime$ in \Cref{app:prelim:aggregation}.

\paragraph{Adversarial Examples.} Given an image $\bx\in\DataSpace$ and a classifier $\DecisionClassifier$, the adversarial example $\bx^\prime\coloneqq\bx+\AdversarialPerturbation$ is visually similar to $\bx$ but either misclassified (i.e., $\DecisionClassifier(\bx^\prime)\neq\DecisionClassifier(\bx)$) or classified as a target class $\TargetClass$ chosen by the attacker (i.e., $\DecisionClassifier(\bx^\prime)=\TargetClass$). Attack algorithms generate adversarial examples by searching for $\AdversarialPerturbation$ such that $\bx^\prime$ fools the classifier while minimizing $\AdversarialPerturbation$ under some distance metrics; for instance, the \LL{p} norm constraint $\|\AdversarialPerturbation\|_p\leq\epsilon$ for a perturbation budget $\epsilon$.

\paragraph{Projected Gradient Descent (PGD).} PGD~\cite{pgd} is one of the most established attacks to evaluate adversarial example defenses. Given a benign example $\bx^0$ and its ground-truth label $\y$, each iteration of the untargeted PGD attack (with \LL{\infty} norm budget $\epsilon$) can be formulated as
\begin{equation}
	\bx^{i+1} \gets \bx^{i} + \alpha \cdot \sgn\s[\big]{\grad{\mathcal{L}\p[\big]{\RandomizedClassifier(\bx^{i}), y}}},
\end{equation}
where $\alpha$ is the step size, $\mathcal{L}$ is the loss function, and each iteration is projected to the $\ell_\infty$ ball around $\bx^0$ of radius $\epsilon$. We use PGD-$\PGDSteps$ to denote the PGD attack with $\PGDSteps$ steps. We outline formulations for other settings and norms in \Cref{app:prelim:attack}.

\paragraph{Expectation over Transformation (EOT).} Since the classifier $\RandomizedClassifier$ is stochastic, the defense evaluation literature~\cite{bpda,adaptive} argues that attacks should target the \emph{expectation} of the gradient using Expectation over Transformation (EOT)~\cite{eot}, which reformulates the PGD attack as
\begin{equation}
	\bx^{i+1} \gets \bx^{i} + \alpha \cdot \sgn\s[\Big]{\EE_{\RandomParameter\sim\RandomizationSpace} \br[\Big]{\grad{\mathcal{L}\p[\big]{\RandomizedClassifier(\bx^{i}), y}}}} 
	\approx \bx^{i} + \alpha \cdot \sgn\s[\Big]{\frac{1}{m}\sum_{j=1}^\EOTSamples \grad{\mathcal{L}\p[\big]{\RandomizedClassifier[\RandomParameter_j](\bx^{i}), y}}},
\end{equation}
where $\EOTSamples$ is the number of samples to estimate the expectation and $\RandomParameter_j\iid\RandomizationSpace$ are sampled parameters for the defense. We use EOT-$\EOTSamples$ to denote the EOT technique with $\EOTSamples$ samples at each PGD step. 
%Note that EOT computes the gradient with lower variance (due to averaging over $m$ samples) and effectively reduces to PGD when $m=1$.

In addition, for a fair comparison among attacks with different PGD steps and EOT samples, we quantify the attack's strength by its total number of gradient computations. For example, attacks using PGD-$\PGDSteps$ and EOT-$\EOTSamples$ will have strength $\PGDSteps\times\EOTSamples$. Although white-box attacks are typically not constrained in this way, it allows for a fair comparison when attacks have finite computing resources (e.g., when EOT is not parallelizable). We discuss more about this quantification in \Cref{app:prelim:strength}.

\section{Most Stochastic Defenses Lack Sufficient Randomness}
\label{sec:revisit}

\citet{bpda} and \citet{adaptive} demonstrate adaptive evaluation of stochastic defenses with the application of EOT. However, it remains unclear why EOT is required (or at least as a ``standard technique'') to break these stochastic defenses. While a commonly accepted explanation is that EOT computes the ``correct gradients'' of models with randomized components~\cite{bpda,adaptive}, the necessity of such correct gradients has not been explicitly discussed. To fill this gap, we revisit stochastic defenses previously broken by EOT and examine their robustness \emph{without} applying EOT. Interestingly, we find that applying EOT is mostly \emph{unnecessary} when evaluating existing stochastic defenses.

\begin{wraptable}{r}{0.35\columnwidth}
\centering
\caption{\centering Evaluation of the random rotation with PGD-$k$ and EOT-$m$.\vspace{-0.3em}}
\label{tab:rotation}
\resizebox{0.35\columnwidth}{!}{
\begin{tabular}{@{}cccc@{}}
\toprule
Attacks                     & $k$ & $m$ & Success Rate \\ \midrule
\multirow{2}{*}{Untargeted} & 10      & 5       & 100\%        \\
                            & 50      & 1       & 100\%        \\ \midrule
\multirow{2}{*}{Targeted}   & 10      & 5       & 99.0\%       \\
                            & 50      & 1       & 99.0\%       \\ \bottomrule
\end{tabular}
}
\vspace{-1em}
\end{wraptable} 
\textbf{Case Study: Random Rotation.}
We start with a simple stochastic defense that randomly rotates the input image for $\RandomParameter\in[-90, 90]$ degrees (chosen at uniform) before classification. This defense is representative for most pre-processing defenses~\cite{input-transformation,random-rescaling,bart}. 
We evaluate this defense on 1,000 ImageNet images with PGD-$\PGDSteps$ and EOT-$\EOTSamples$ under the constraint $\PGDSteps\times\EOTSamples=50$, as discussed in \Cref{sec:prelim}. All attacks use maximum \LL{\infty} perturbation $\epsilon=8/255$ with step size chosen from $\alpha\in\s{1/255, 2/255}$. 
The results are shown in \Cref{tab:rotation}, where PGD-50 performs equally well as PGD-10 combined with EOT-5. This observation suggests that \emph{some stochastic defenses are already breakable without applying EOT}, casting doubt on a long-held assumption that stochastic defenses simply invalidate attacks designed for deterministic defenses.

\textbf{Comprehensive Evaluations.}
We then extend the above case study to other stochastic defenses evaluated in the literature. Specifically, we replicate the (untargeted) adaptive evaluation of stochastic defenses from \citet{bpda} and \citet{adaptive} with their official implementation. We only change the attack's hyper-parameters (e.g., number of iterations and learning rate) and disable EOT by setting its number of samples to one ($\EOTSamples=1$), which avoids potential implementation flaws if removed from the source code. The comparison between evaluations with and without applying EOT is summarized in \Cref{tab:previous-defenses}, which serves as a missing ablation study of adaptive evaluations in the literature. The experimental settings are identical within each row (detailed in \Cref{app:revisit}). 

Interestingly, we find it \emph{unnecessary} to break these defenses with EOT, as long as the standard attack runs for more iterations with a smaller learning rate. For such defenses, standard iterative attacks already contain an \emph{implicit expectation} across iterations to capture the limited randomness. This observation implies that most stochastic defenses lack sufficient randomness to withstand even standard attacks designed for deterministic defenses.
Therefore, increasing randomness becomes a promising approach to enhancing stochastic defenses, as adopted by recent defenses~\cite{smoothing,bart}. 
Note that this ablation study only aims to inspire potential ways of enhancing stochastic defenses; it does not invalidate EOT for stronger adaptive evaluations of stochastic defenses.

\begin{table}[tb]
\caption{The missing ablation study of adaptive evaluations of stochastic defenses in the literature. Notations: attack iterations $k$, EOT samples $m$, learning rate $\alpha$, number of gradient queries $k\times m$. The details of these defenses and their evaluation settings are in \Cref{app:revisit}.}
\label{tab:previous-defenses}
\resizebox{\columnwidth}{!}{%
\begin{tabular}{@{}l|rrcrc|rccrc@{}}
\toprule
\multicolumn{1}{c|}{\multirow{2}{*}{Defenses}} & \multicolumn{5}{c|}{Original Adaptive Evaluation (w/ EOT)} & \multicolumn{5}{c}{Our Ablation Study (w/o EOT)}  \\
\multicolumn{1}{c|}{}                                     & \multicolumn{1}{c}{$k$} & \multicolumn{1}{c}{$m$} & \multicolumn{1}{c}{$\alpha$} & \multicolumn{1}{c}{$k\times m$} & Success Rate & \multicolumn{1}{c}{$k$} & \multicolumn{1}{c}{$m$} & \multicolumn{1}{c}{$\alpha$} & \multicolumn{1}{c}{$k\times m$} & Success Rate  \\ \midrule
\citet{input-transformation}    &    1,000       & 30      & 0.1      & 30,000   & 100\%   &   1,000 & 1 & 0.001   &   1,000  & 99.0\%  \\
\citet{random-rescaling}        &    1,000       & 30      & 0.1      & 30,000   & 100\%   &     200 & 1 & 0.1     &     200  & 100\%   \\
\citet{activation-pruning}      &      500       & 10      & 0.1      & 5,000    & 100\%   &     500 & 1 & 0.1     &     500  & 100\%   \\
\citet{kwinners}                &      100       & 1,000   & 0.01     & 100,000  & 100\%   &  40,000 & 1 & 0.1/255 &  40,000  & 98.4\%  \\
\citet{odds}                    &      100       & 40      & 0.2/255  & 4,000    & 100\%   &   4,000 & 1 & 0.1/255 &   4,000  & 96.1\%  \\ \bottomrule
\end{tabular}
}
\end{table}

\section{Trade-offs between Robustness and Invariance}
\label{sec:tradeoff}

When stochastic pre-processing defenses {\em do have} sufficient randomness, they must ensure that the utility of the defended model is preserved in the face of randomness. To achieve high utility, existing defenses mostly rely on augmentation invariance through \emph{trained invariance}~\cite{lyle2020benefits}. In such a case, the invariance is achieved by applying the defense's randomness to the training data so as to guide the model in learning their transformations. For defenses based on stochastic pre-processor $\InputTransformation_{\RandomParameter}$, each data sample from the dataset gets augmented with $\InputTransformation_{\RandomParameter}$ sampled from the randomization space $\RandomizationSpace$, and the risk is minimized over such augmented data.

\begin{figure}[t]
\centering
\begin{subfigure}{0.325\linewidth}
	\includegraphics[width=\linewidth]{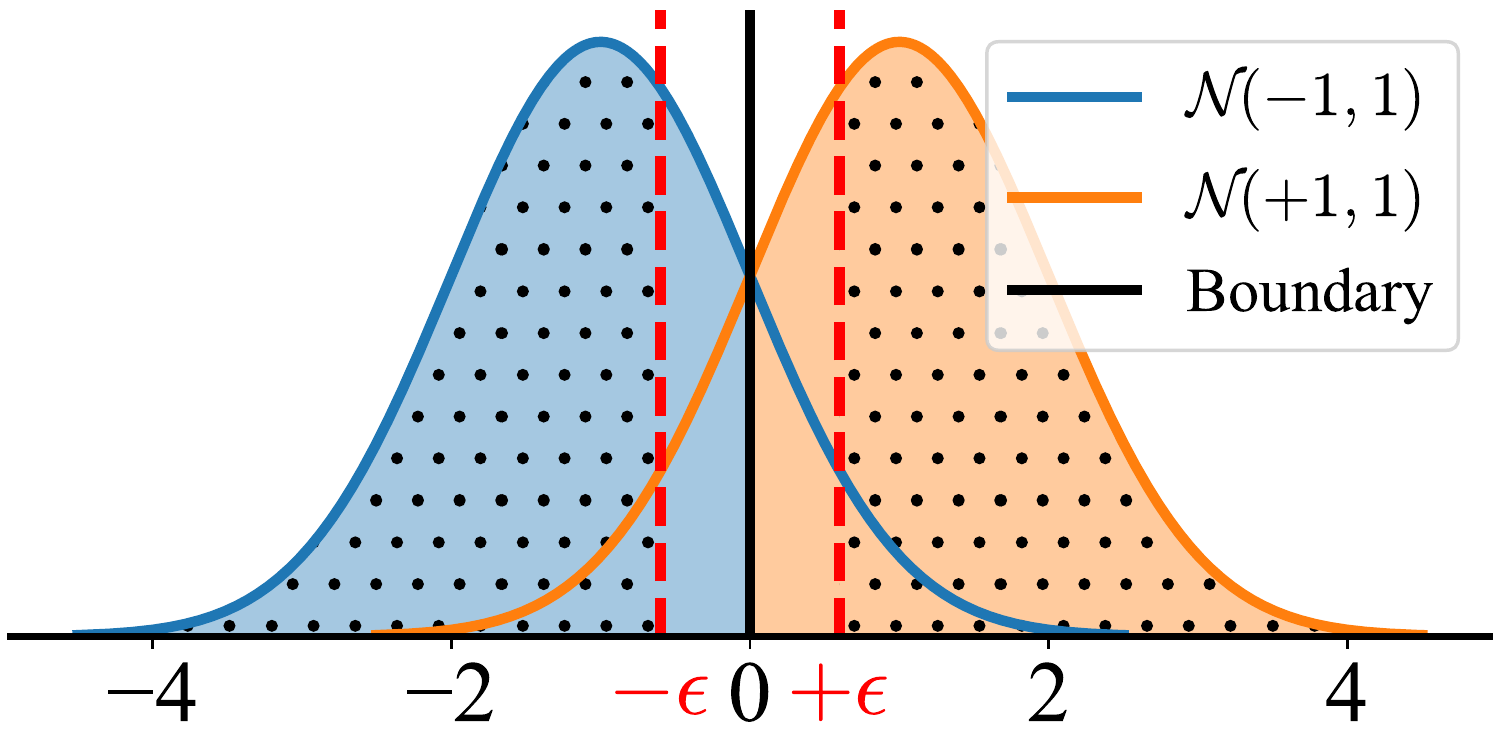}
	\caption{Undefended}
	\label{fig:demo:1}
\end{subfigure}
\begin{subfigure}{0.325\linewidth}
	\includegraphics[width=\linewidth]{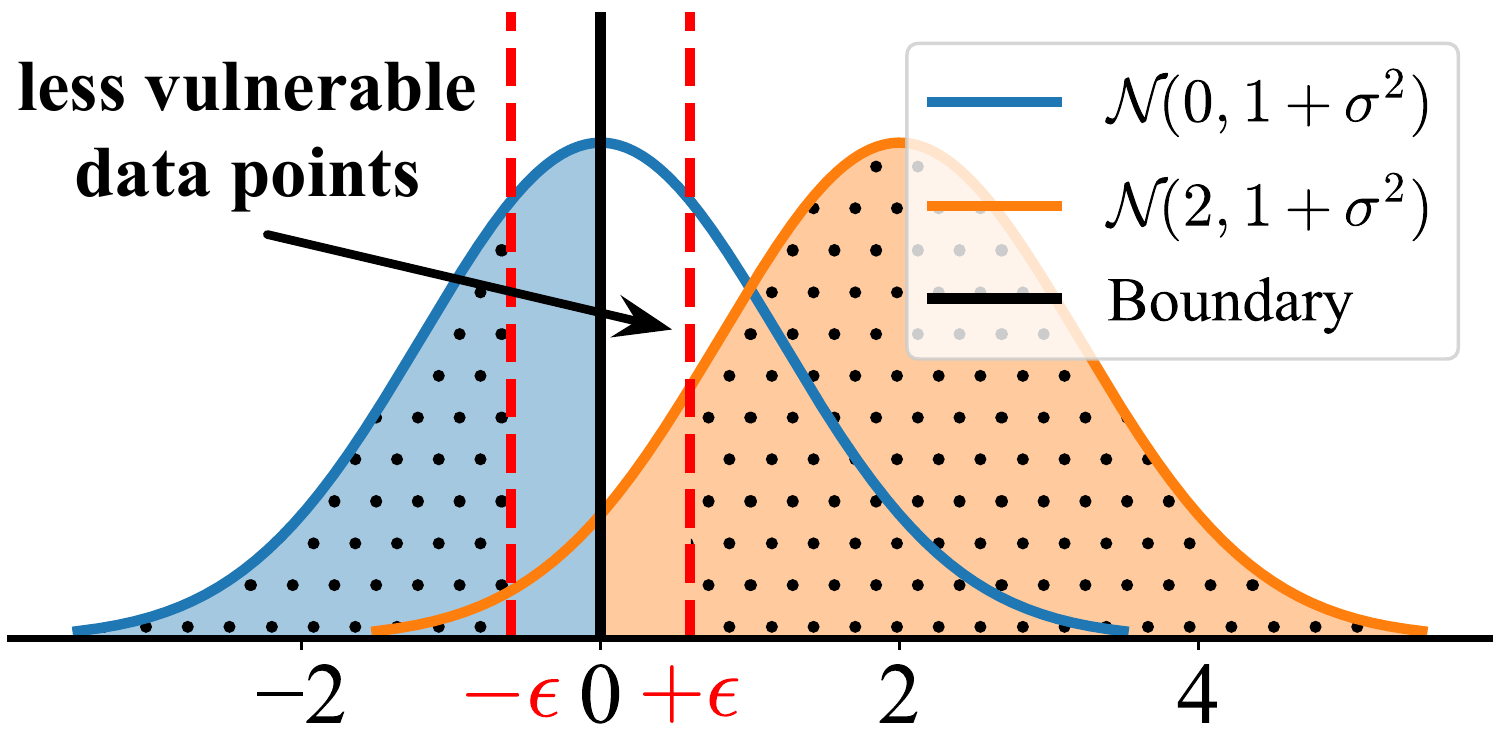}
	\caption{Defended (Lack of Invariance)}
	\label{fig:demo:2}
\end{subfigure}
\begin{subfigure}{0.325\linewidth}
	\includegraphics[width=\linewidth]{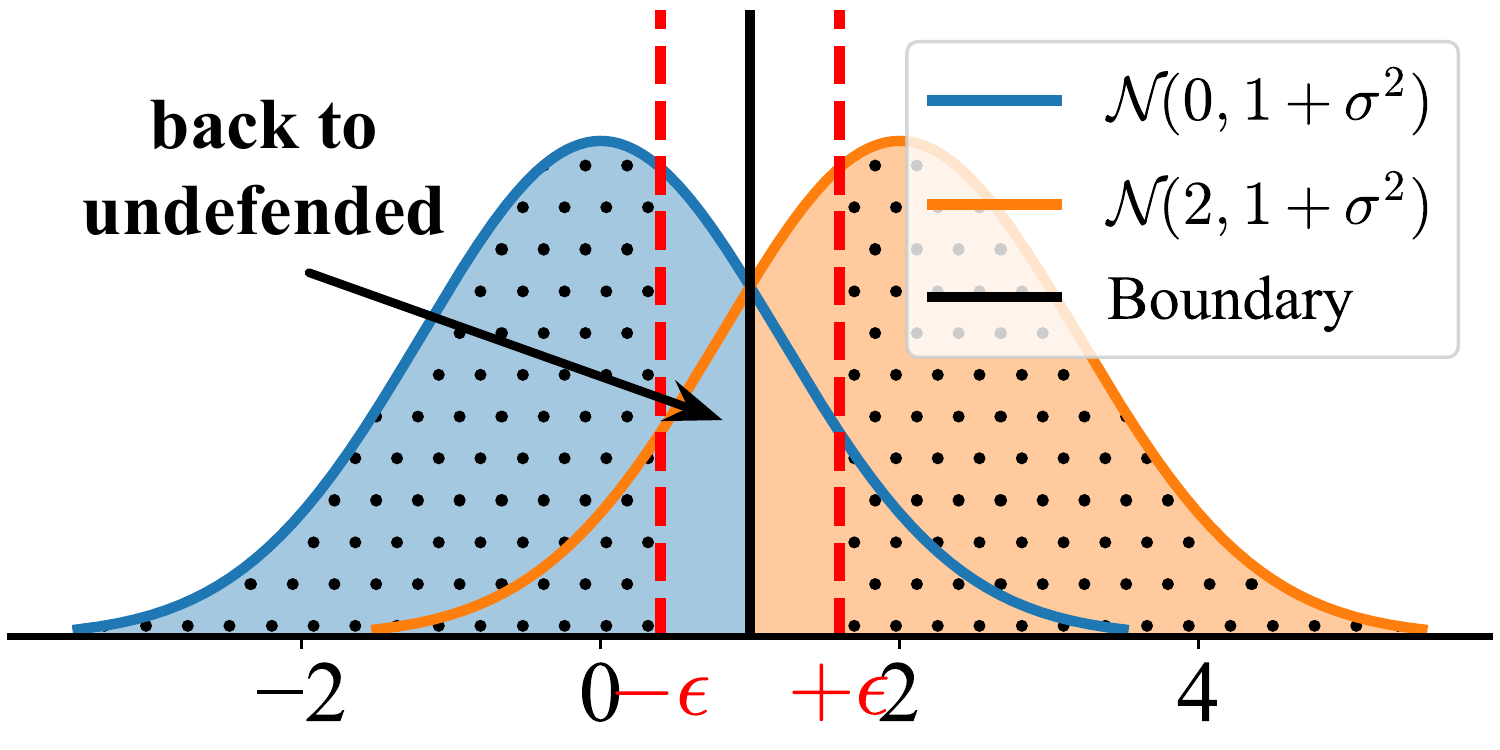}
	\caption{Defended (Trained Invariance)}
	\label{fig:demo:3}
\end{subfigure}
\caption{Illustration of the binary classification task we consider. The curves are the probability density function of two classes of data. Shadowed area denotes correct classification. Dotted area denotes robustly correct classification under the \LL{\infty}-bounded adversary with perturbation budget $\epsilon$.}
\label{fig:demo}
\end{figure}

The defended classifier $\RandomizedDecisionClassifier(\bx) \coloneqq \DecisionClassifier(\InputTransformation_{\RandomParameter}(\bx))$ is invariant under the randomization space $\RandomizationSpace$ if
\begin{equation}
\label{eq:inv}
    \DecisionClassifier(\InputTransformation_{\RandomParameter}(\bx)) = \DecisionClassifier(\bx), \quad \forall\ \RandomParameter\in\RandomizationSpace, \bx \in \DataSpace.
\end{equation}
As we can observe from the definition, invariance has direct implications on the performance of stochastic pre-processing defenses. If the classifier is invariant under the defense's randomization space $\RandomizationSpace$ as is defined in~\Cref{eq:inv}, then the defense should not work -- computing the model and its gradients over randomization $\RandomParameter\in\RandomizationSpace$ is the same as if $\InputTransformation_{\RandomParameter}$ was not applied at all. This observation suggests a direct coupling between invariance and performance of the defense: the more invariant, hence performant, the model is under a given randomization space, the less protection such a defense would provide. In this section, we present a simple theoretical setting where this coupling provably exists, as illustrated in \Cref{fig:demo}. Detailed arguments are deferred to \Cref{app:tradeoff:analysis}.

\textbf{Binary Classification Task.}
We consider a class-balanced dataset $\mathcal{D}$ consisting of input-label pairs $(x, y)$ with $y\in\s{-1,+1}$ and $x|y\sim\mathcal{N}(y, 1)$, where $\mathcal{N}(\mu, \sigma^2)$ is a normal distribution with mean $\mu$ and variance $\sigma^2$. Moreover, an \LL{\infty}-bounded adversary perturbs the input with a small $\delta$ to fool the classifier for $\norm{\delta}_{\infty}\leq\epsilon$. We quantify the classifier's robustness by its robust accuracy, i.e., the ratio of correctly classified samples that remain correct after being perturbed by the adversary.

\textit{Undefended Classification.}
We start with the optimal linear classifier $\DecisionClassifier(x)\coloneqq\sgn(x)$ without any defense in \Cref{fig:demo:1}. This classifier attains robust accuracy
\begin{equation}
\label{eq:linear-natural}
\Pr\br[\big]{
  \DecisionClassifier( x + \delta ) = y \ |\ \DecisionClassifier( x ) = y
}
=
\frac{
  \Pr\br[\big]{
    \DecisionClassifier( x + \delta ) = y \land \DecisionClassifier( x ) = y
  }
}{
  \Pr\br[\big]{
    \DecisionClassifier( x ) = y
  }
}
=
\frac{\Phi(1-\epsilon)}{\Phi(1)},
\end{equation}
where $\Phi$ is the cumulative distribution function of $\mathcal{N}(0, 1)$.

\textit{Defended Classification.}
We then try to improve adversarial robustness by introducing a stochastic pre-processing defense $\InputTransformation_{\theta}(x) \coloneqq x + \theta$, where $\theta\sim\mathcal{N}(1, \sigma^2)$ is the random variable parameterizing the defense. This defense characterizes common pre-processing defenses that enforce randomness while shifting the input distribution. Here, the processed input follows a shifted distribution $\InputTransformation_{\theta}(x)\sim\mathcal{N}(y+1, 1+\sigma^2)$ in \Cref{fig:demo:2}. The defended classifier $\DecisionClassifier_{\theta}(x)=\sgn(x+\theta)$ has robust accuracy
\begin{equation}
\label{eq:linear-defended}
\Pr\br[\big]{
  \DecisionClassifier_{\theta}( x + \delta ) = y \ |\ \DecisionClassifier_{\theta}( x ) = y
}
=
\frac{
  \Pr\br[\big]{
    \DecisionClassifier_{\theta}( x + \delta ) = y \land \DecisionClassifier_{\theta}( x ) = y
  }
}{
  \Pr\br[\big]{
    \DecisionClassifier_{\theta}( x ) = y
  }
}
=
\frac{\Phi^\prime(-\epsilon) + \Phi^\prime(2-\epsilon)}{\Phi^\prime(0) + \Phi^\prime(2)},
\end{equation}
where $\Phi^\prime(x)\coloneqq\Phi(x/\sqrt{1+\sigma^2})$ is the cumulative distribution function of $\mathcal{N}(0, 1+\sigma^2)$. At this point, we have not fit the classifier on processed inputs. Due to its lack of invariance, the defended classifier has low utility yet higher robust accuracy than the undefended one in \Cref{eq:linear-natural}.

\textit{Defended Classification (Trained Invariance).}
As discussed above, one critical step of stochastic pre-processing defenses is to preserve the defended model's utility by minimizing the risk over augmented data $\InputTransformation_\theta(x)$, which leads to a new defended classifier $\DecisionClassifier^+_{\theta}(x)=\sgn(x+\theta-1)$ in \Cref{fig:demo:3}. As a result, this new defended classifier achieves higher invariance with robust accuracy
\begin{equation}
\label{eq:linear-trained}
\Pr\br[\big]{
  \DecisionClassifier^+_{\theta}( x + \delta ) = y \ |\ \DecisionClassifier^+_{\theta}( x ) = y
}
=
\frac{
  \Pr\br[\big]{
    \DecisionClassifier^+_{\theta}( x + \delta ) = y \land \DecisionClassifier^+_{\theta}( x ) = y
  }
}{
  \Pr\br[\big]{
    \DecisionClassifier^+_{\theta}( x ) = y
  }
}
=
\frac{\Phi^\prime(1-\epsilon)}{\Phi^\prime(1)},
\end{equation}
which is less robust than the previous less-invariant classifier $\DecisionClassifier_{\theta}$ in \Cref{eq:linear-defended}. However, one may observe that this classifier, though loses some robustness compared with $\DecisionClassifier_{\theta}$, is still more robust than the original undefended classifier $\DecisionClassifier$ in \Cref{eq:linear-natural}. This part of robustness comes from the changed data distribution due to the defense's randomness. It shows that we have not achieved perfect invariance to the defense's randomness, thus gaining some robustness at the cost of utility.

\textit{Defended Classification (Perfect Invariance).}
Furthermore, these defenses usually leverage majority vote to obtain stable predictions, which finally produces a perfectly invariant defended classifier
\begin{equation}
\label{eq:linear-reduction}
    \DecisionClassifier_{\theta}^*(x)
    =
    \sgn\s*{\frac{1}{n}\sum_{i=1}^n \DecisionClassifier^+_{\theta_i}(x)}
    =
    \sgn\s*{\frac{1}{n}\sum_{i=1}^n \sgn(x+\theta_i-1)}
    \to
    \sgn(x)
    =
    F(x),
\end{equation}
where $\theta_i\iid\mathcal{N}(1, \sigma^2)$ are sampled parameters. In such a case, the defended classifier reduces to the original undefended classifier with the original robust accuracy:
\begin{equation}
\label{eq:linear-invariant}
\Pr\br[\big]{
  \DecisionClassifier_{\theta}^*( x + \delta ) = y
  \ |\ 
  \DecisionClassifier_{\theta}^*( x ) = y
}
=
\Pr\br[\big]{
  \DecisionClassifier( x + \delta ) = y
  \ |\ 
  \DecisionClassifier( x ) = y
}
=
\frac{\Phi(1-\epsilon)}{\Phi(1)}.
\end{equation}

\emph{Summary.}
The above theoretical setting illustrates how stochastic pre-processing defenses first induce variance on the binary classifier we consider to provide adversarial robustness in \Cref{eq:linear-defended}, and how they finally take back such variance in \Cref{eq:linear-trained,eq:linear-invariant} to recover utility. We then extend the above coupling between robustness and invariance to a general trade-off in the following theorem, whose detailed descriptions and proofs are deferred to \Cref{app:tradeoff:theorem,app:proofs}, respectively.
\begin{restatable}[Trade-off between Robustness and Invariance]{theorem}{theoremTradeoff}
\label{theorem}
	Given the above theoretical setting and assumptions, when the defended classifier $F_\theta(x)$ achieves higher invariance $R(k)$ under the defense's randomization space to preserve utility, the adversarial robustness provided by the defense strictly decreases.
\end{restatable}

In a nutshell, we prove the strictly opposite monotonic behavior of robustness and invariance when the classifier shifts its decision boundary and employs majority vote to preserve utility. It shows that stochastic pre-processing defenses provide robustness by explicitly reducing the model's invariance to added randomized transformations, and the robustness disappears once the invariance is recovered.

\section{Experiments}
\label{sec:exp}

Our experiments are designed to answer the following two questions.

\textbf{Q1: What properties make applying EOT beneficial when evaluating stochastic defenses?}

We show that applying EOT is only beneficial when the defense is sufficiently randomized; otherwise standard attacks already perform well and leave no room for EOT to improve.

\textbf{Q2: What is the limitation of stochastic defenses when they do have sufficient randomness?}

We show a trade-off between the stochastic defense's robustness and the model's invariance to the defense itself. Such defenses become less effective when the defended model achieves higher invariance to their randomness, as required to preserve utility under the defense.

\subsection{Experimental Settings}
\label{sec:exp:settings}

\textbf{Datasets \& Models.}
We conduct all experiments on ImageNet~\cite{imagenet} and ImageNette~\cite{imagenette}. For ImageNet, our test data consists of 1,000 images randomly sampled from the validation set.
ImageNette is a ten-class subset of ImageNet, and we test on its validation set.
We adopt various ResNet~\cite{resnet} models. For defenses with low randomness, we evaluate them on ImageNet with a pre-trained ResNet-50 with Top-1 accuracy 75.9\%. For defenses with higher randomness (thus requiring fine-tuning), we switch to ImageNette and a pre-trained ResNet-34 with Top-1 accuracy 96.9\% to reduce the training cost like previous work~\cite{aggmopgd}. These models are fine-tuned on the training data processed by tested defenses. As a special case, we also evaluate randomized smoothing on ImageNet using the ResNet-50 models from \citet{smoothing}. More details of datasets and models can be found in \Cref{app:exp:datasets,app:exp:models}.

\textbf{Defenses \& Metrics.}
We focus on stochastic defenses allowing us to increase randomness: randomized smoothing~\cite{smoothing} and BaRT~\cite{bart}. For randomized smoothing, we vary the variance of the added Gaussian noise. For BaRT, we vary the number $\kappa$ of applied randomized transformations. Note that we have evaluated other stochastic defenses and discussed their low randomness in \Cref{sec:revisit}. We measure the defense's performance by the defended model's \emph{benign accuracy} and the attack's \emph{success rate}, all evaluated with majority vote over $n=500$ predictions. The attack's success rate is the ratio of samples that do not satisfy the attack's objective prior to the attack but satisfy it after the attack. For example, we discard samples that were misclassified before being perturbed in untargeted attacks. Details of the evaluated defenses can be found in \Cref{app:exp:defenses}. 

\textbf{Attacks.}
We evaluate defenses with standard PGD combined with EOT and focus on the \LL{\infty}-bounded adversary with a perturbation budget $\epsilon=8/255$ in both untargeted and targeted settings. We only use constant step sizes and no random restarts for PGD. We only conduct adaptive evaluations, where the defense is included in the attack loop with non-differentiable components captured by BPDA~\cite{bpda}.
We also utilize AutoPGD~\cite{autopgd} to avoid selecting the step size when it is computationally expensive to repeat experiments in \Cref{sec:exp:tradeoff}. More details and intuitions of the attack's settings and implementation can be found in \Cref{app:exp:attacks}. Our code is available at \url{https://github.com/wi-pi/stochastic-preprocessing-defenses}.

\subsection{Q1: Evaluate the Benefits of Applying EOT under Different Settings}
\label{sec:exp:eot}

In \Cref{sec:revisit}, we showed that standard attacks are sufficient to break most stochastic defenses due to their lack of randomness. Here, we aim to understand what properties make applying EOT beneficial when evaluating stochastic defenses. We design a systematic evaluation of stochastic defenses with different levels of randomness and check if applying EOT improves the attack.

\begin{figure}[tb]
	\centering
	\begin{subfigure}[t]{0.328\textwidth}
		\includegraphics[width=\linewidth]{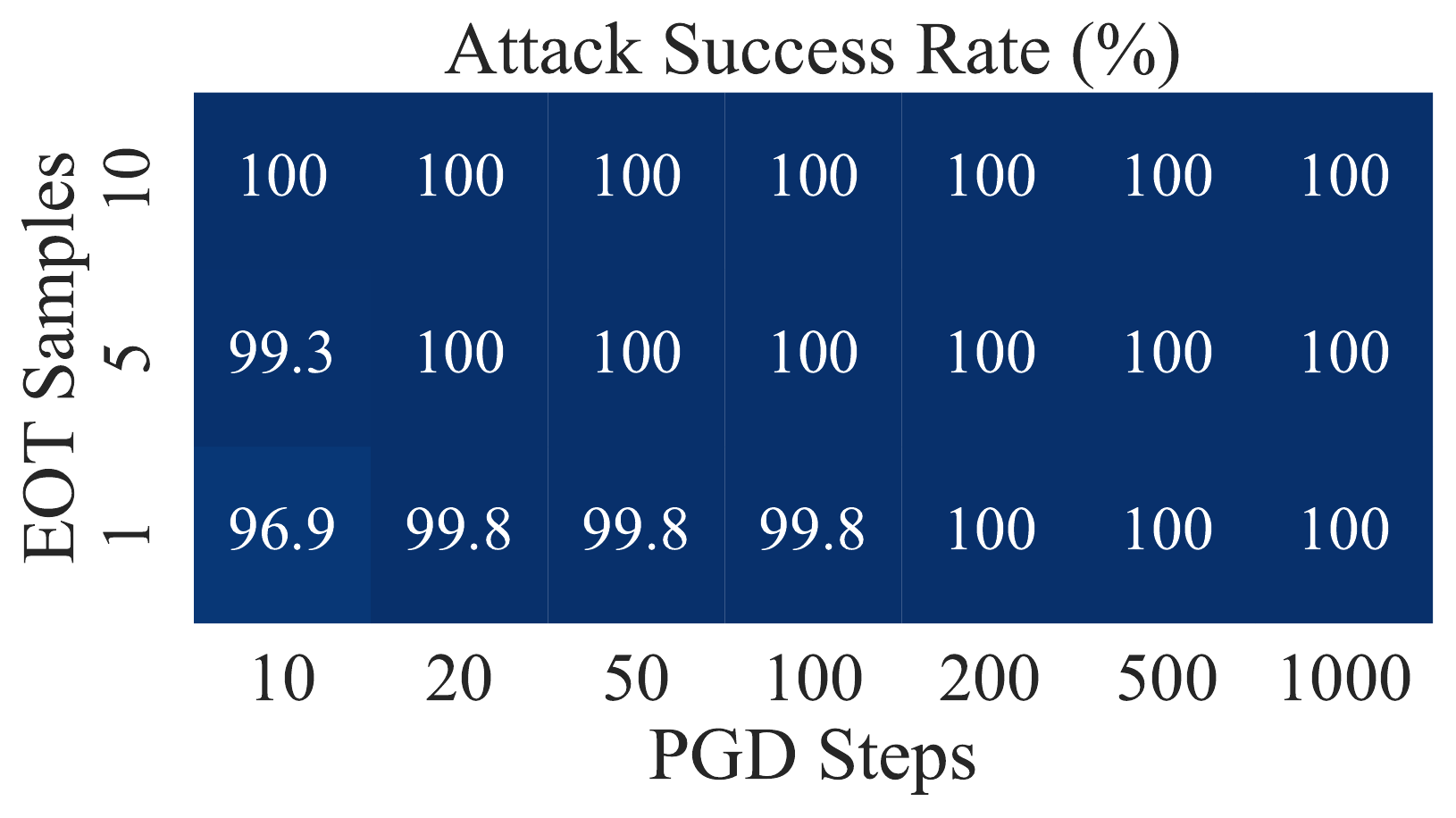}
		\caption{Untargeted (step size = 1/255)}
	\end{subfigure}
	\begin{subfigure}[t]{0.328\textwidth}
		\includegraphics[width=\linewidth]{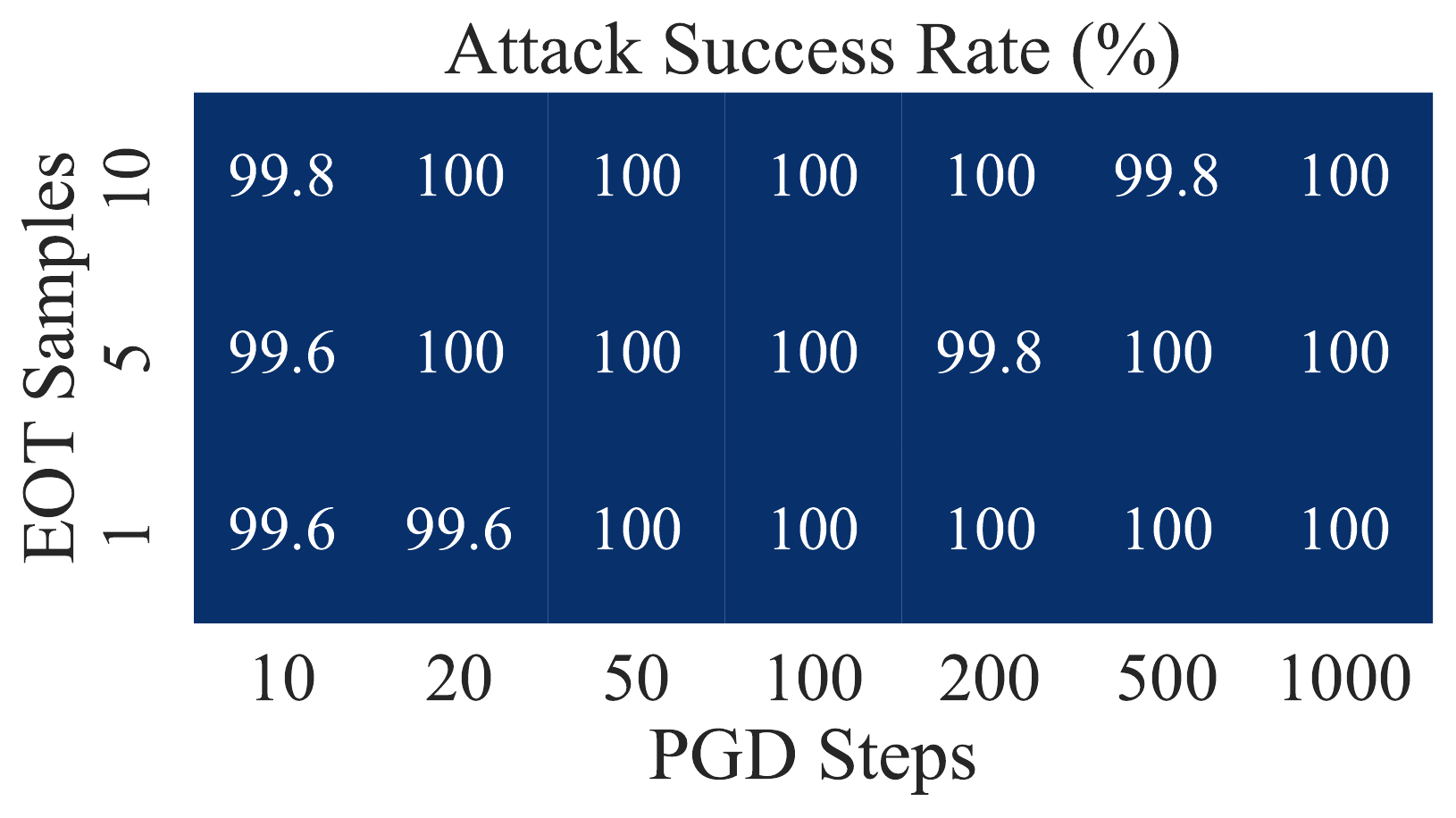}
		\caption{Untargeted (step size = 2/255)}
	\end{subfigure}
	\begin{subfigure}[t]{0.328\textwidth}
		\includegraphics[width=\linewidth]{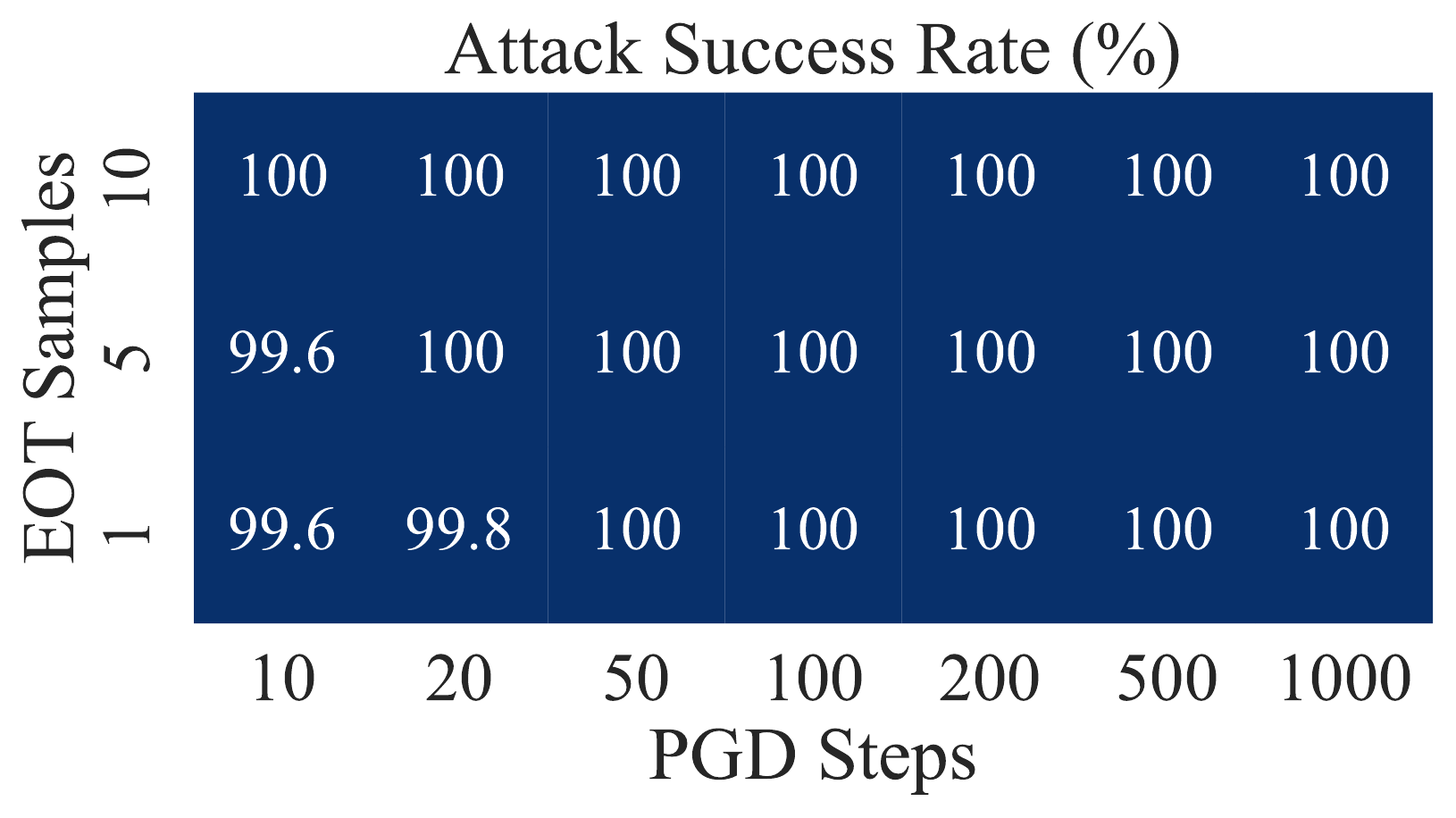}
		\caption{Untargeted (step size = \textbf{best})}
	\end{subfigure}
	\begin{subfigure}[t]{0.328\textwidth}
		\includegraphics[width=\linewidth]{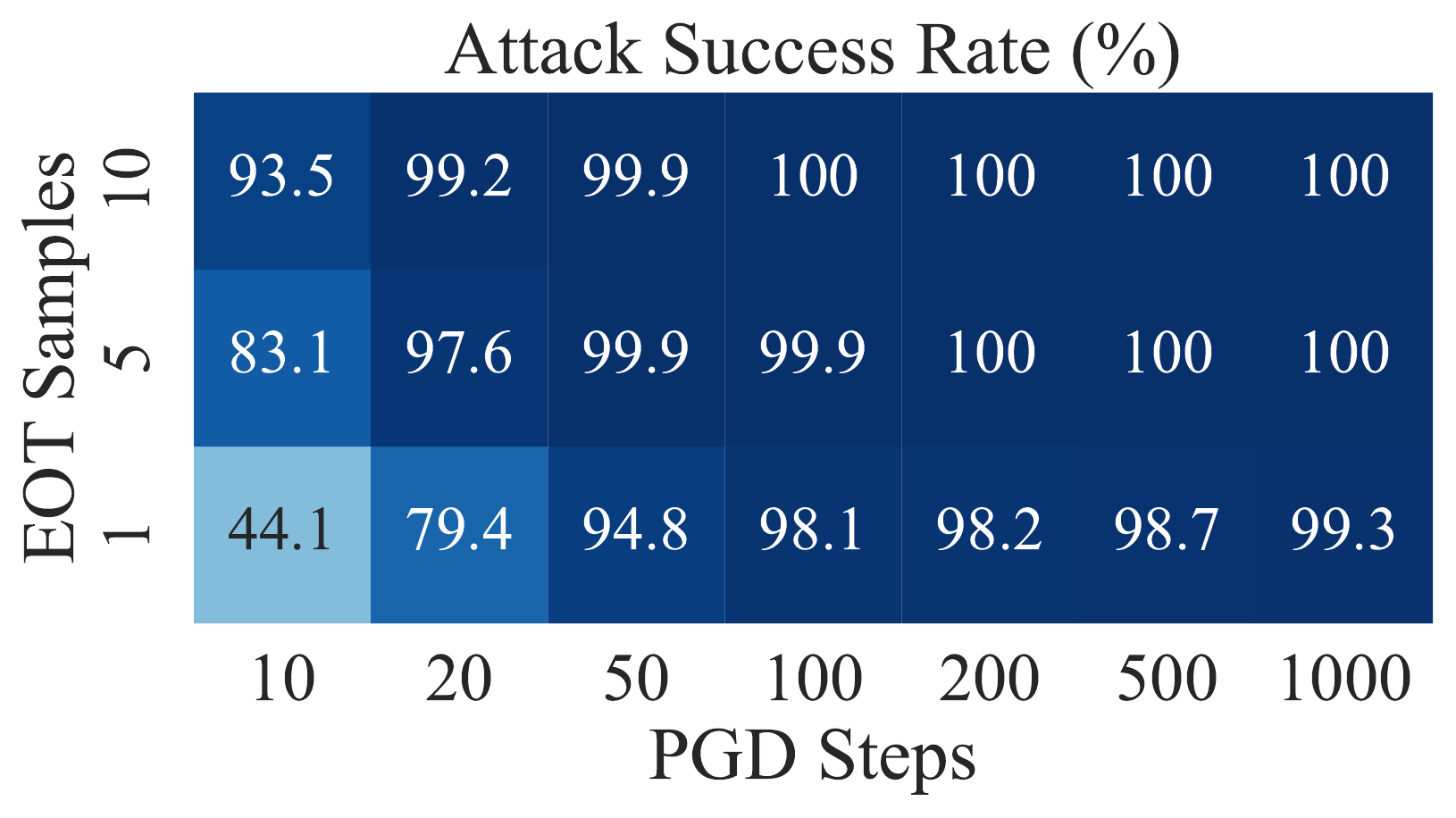}
		\caption{Targeted (step size = 1/255)}
	\end{subfigure}
	\begin{subfigure}[t]{0.328\textwidth}
		\includegraphics[width=\linewidth]{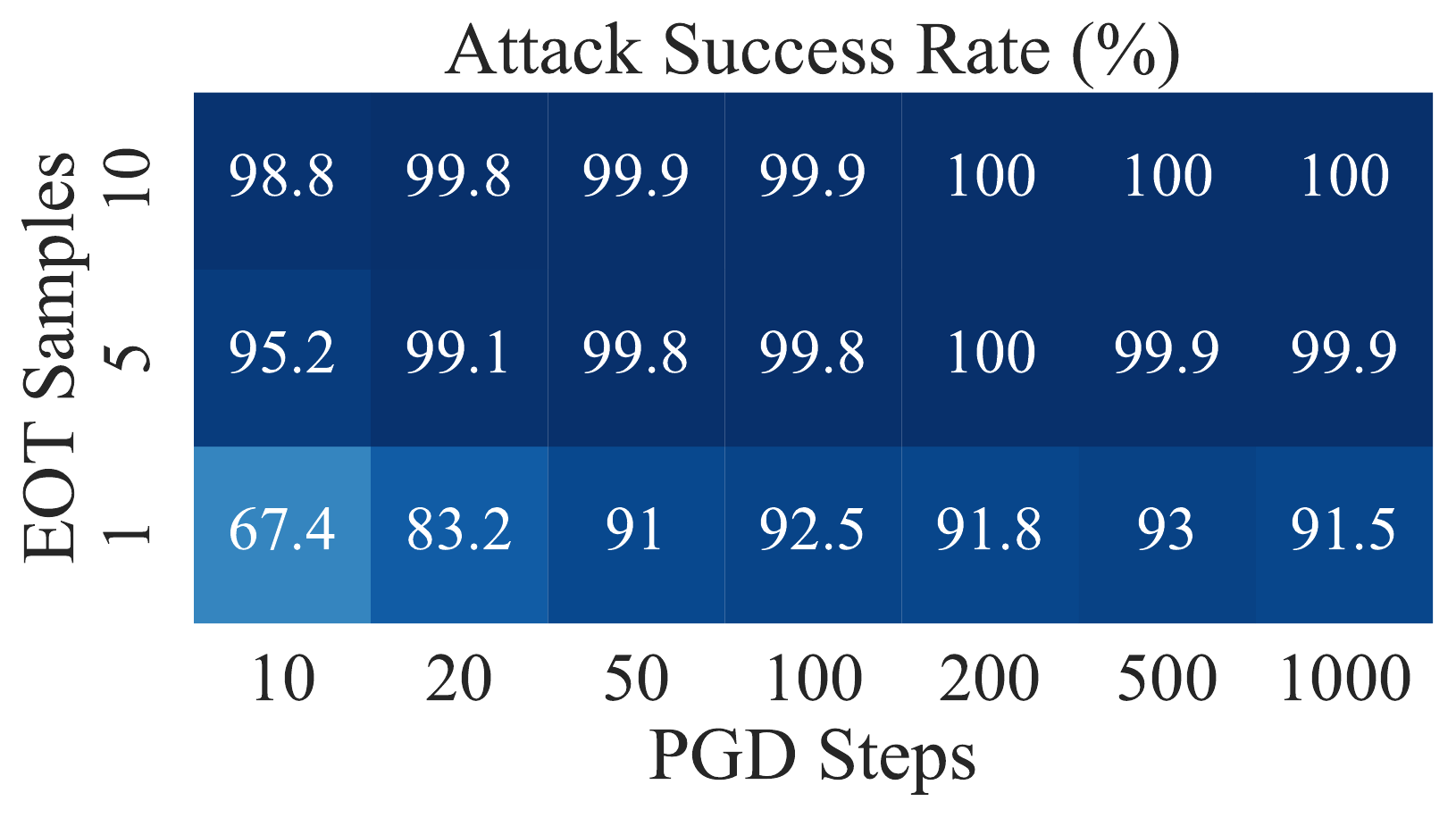}
		\caption{Targeted (step size = 2/255)}
	\end{subfigure}
	\begin{subfigure}[t]{0.328\textwidth}
		\includegraphics[width=\linewidth]{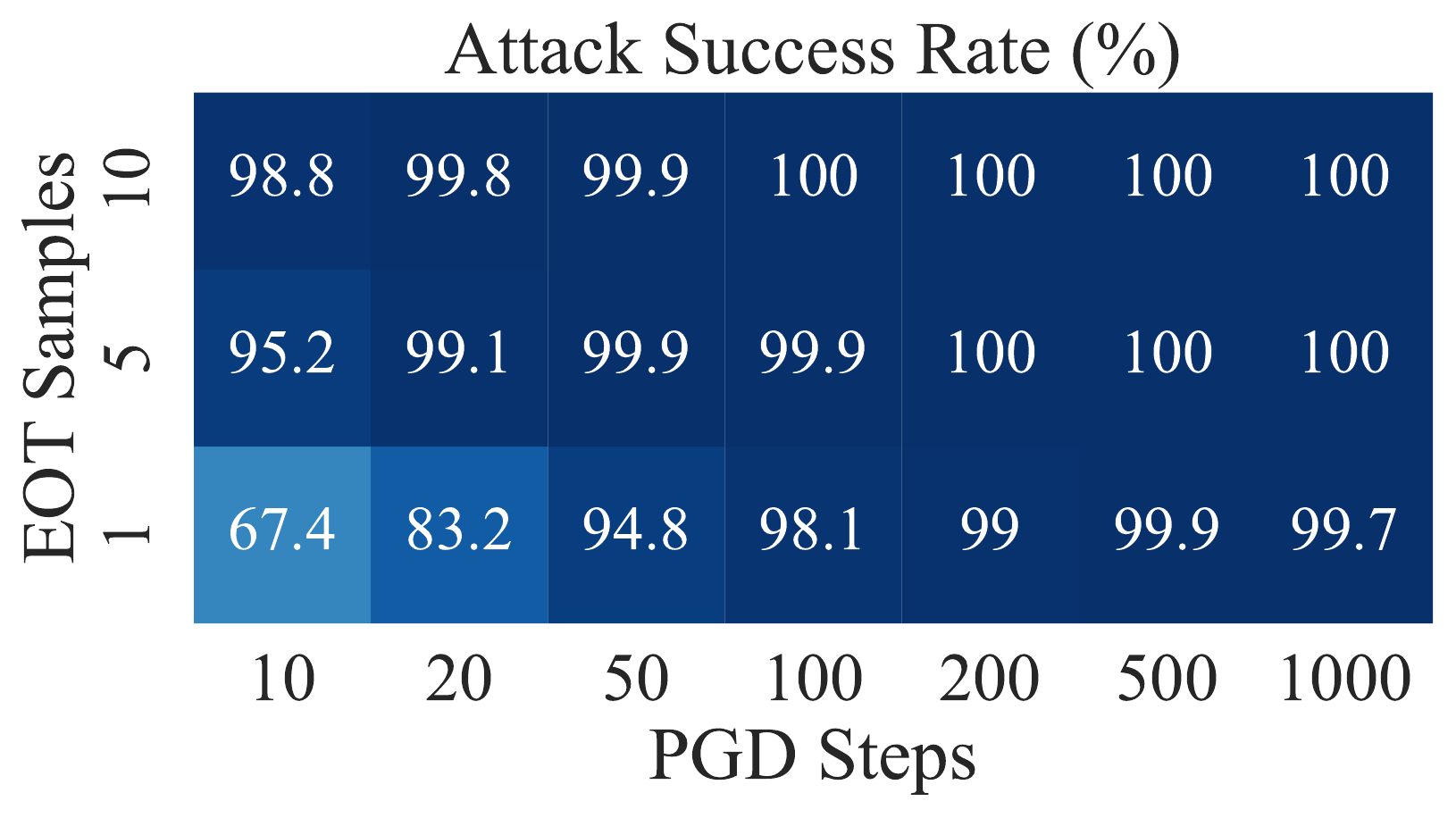}
		\caption{Targeted (step size = \textbf{best})}
		\label{fig:attack-noise-injection-targeted-best}
	\end{subfigure}
	\caption{Evaluation of BaRT's noise injection defense on ImageNet. Standard PGD without applying EOT (i.e., applying EOT-1) is already good enough, leaving limited space for EOT to improve.}
	\label{fig:attack-noise-injection}
\end{figure}

\textbf{Stochastic Defenses with Low Randomness.}
We start with BaRT's noise injection defense, which perturbs the input image with noise of distributions and parameters chosen at random. While this defense has low randomness, it yields meaningful results. We evaluate this defense with various combinations of PGD and EOT\footnotemark. The performance of untargeted and targeted attacks is shown in \Cref{fig:attack-noise-injection}. We test multiple step sizes and summarize their best results (discussed in \Cref{app:exp:results:lr}).

\footnotetext{Note that we do not intend to find a heuristic for the best combination of PGD-$k$ and EOT-$m$, as it is out of the scope of the question that we want to answer. However, it is still possible to correlate the choice of $k$ and $m$ with the convergence analysis of stochastic gradient descent, which we will briefly discuss in \Cref{app:prelim:strength}.}

In this case, standard PGD attacks are already good enough when the defense has insufficient randomness, leaving no space for improvements from EOT. In \Cref{fig:attack-noise-injection-targeted-best}, both (1) PGD-10 combined with EOT-10 and (2) PGD-100 combined with EOT-1 have near 100\% success rates. This result is consistent with our observations in \Cref{sec:revisit} in both untargeted and targeted settings\footnotemark.

\footnotetext{The only caveat is that targeted attacks are more likely to benefit from EOT, as their objectives are stricter and may have better performance with gradients of higher precision.}

\begin{figure}[tb]
	\centering
	\begin{subfigure}[t]{0.328\textwidth}
		\includegraphics[width=\linewidth]{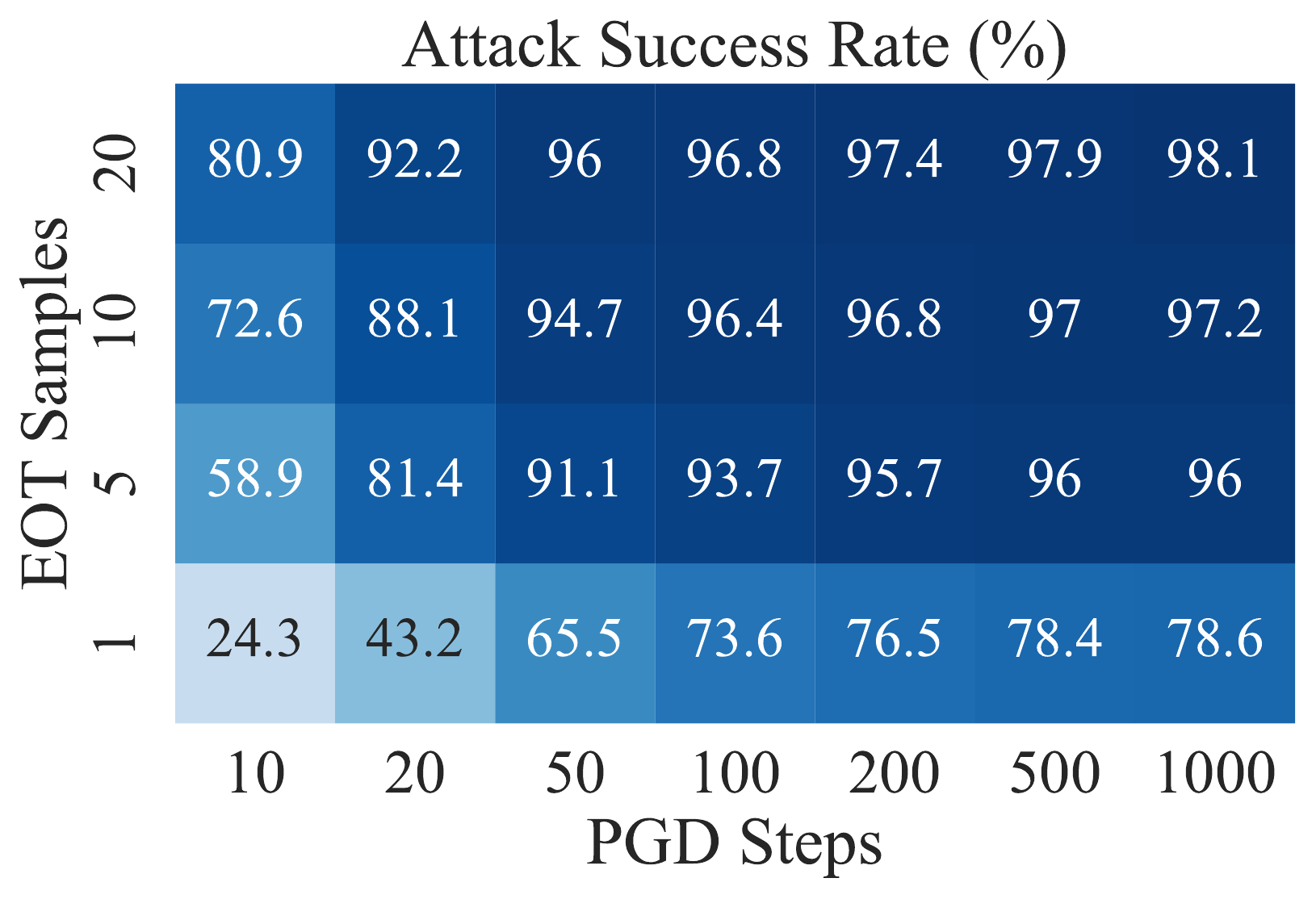}
		\caption{$\sigma = 0.25$ (step size = 2/255)}
	\end{subfigure}
	\begin{subfigure}[t]{0.328\textwidth}
		\includegraphics[width=\linewidth]{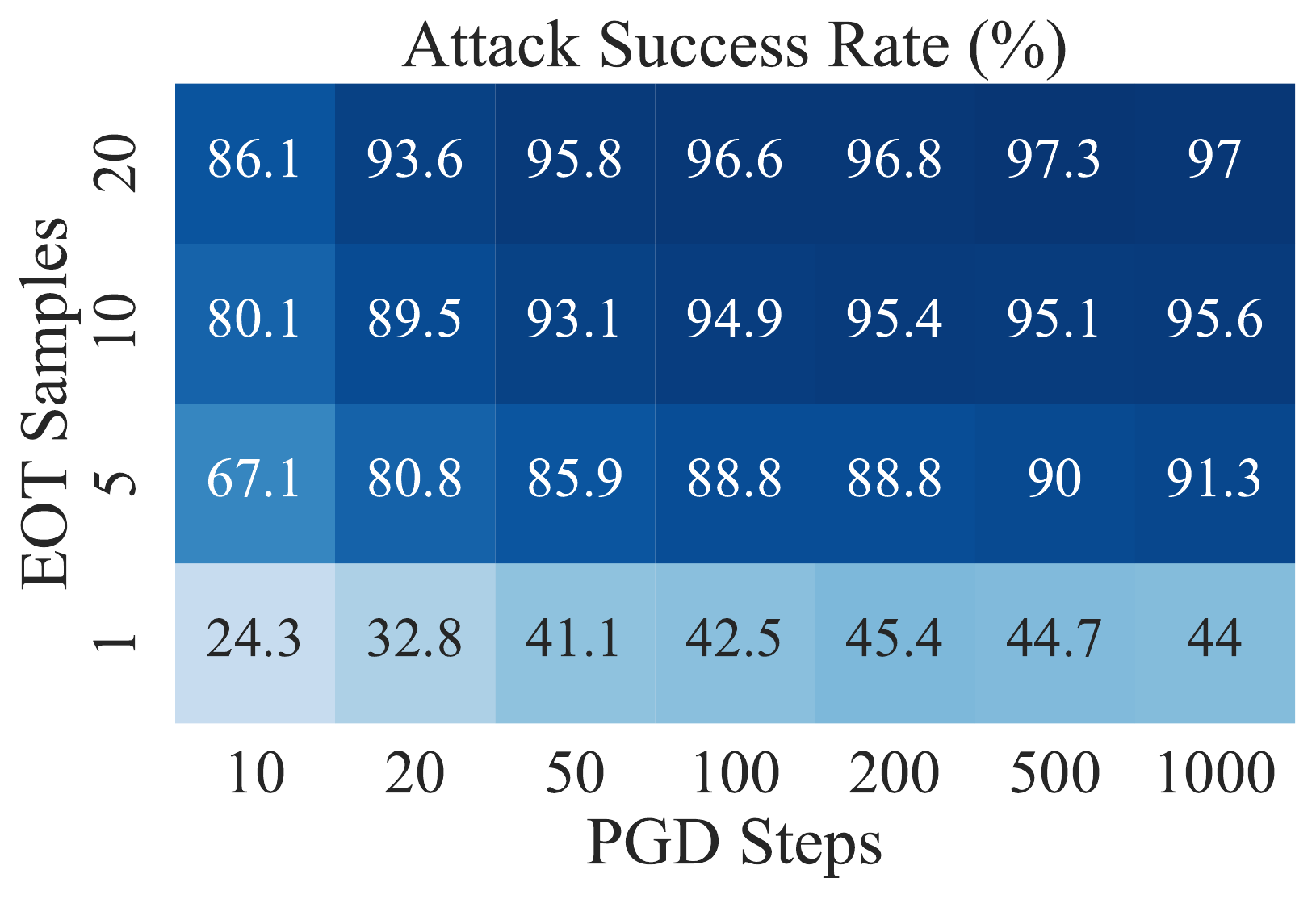}
		\caption{$\sigma = 0.25$ (step size = 4/255)}
	\end{subfigure}
	\begin{subfigure}[t]{0.328\textwidth}
		\includegraphics[width=\linewidth]{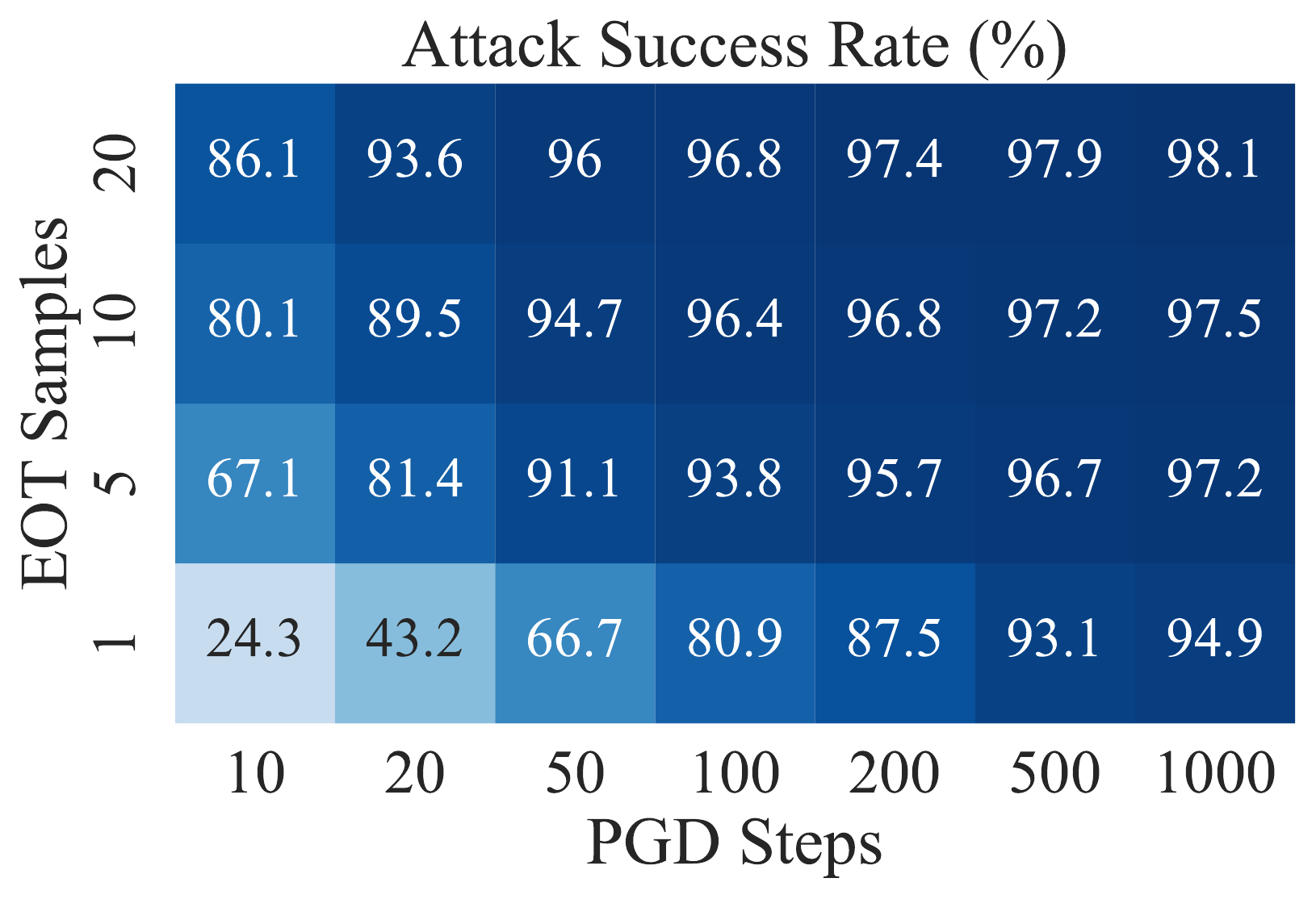}
		\caption{$\sigma = 0.25$ (step size = \textbf{best})}
		\label{fig:attack-smoothing-heatmap-0.25}
	\end{subfigure}
	\begin{subfigure}[t]{0.328\textwidth}
		\includegraphics[width=\linewidth]{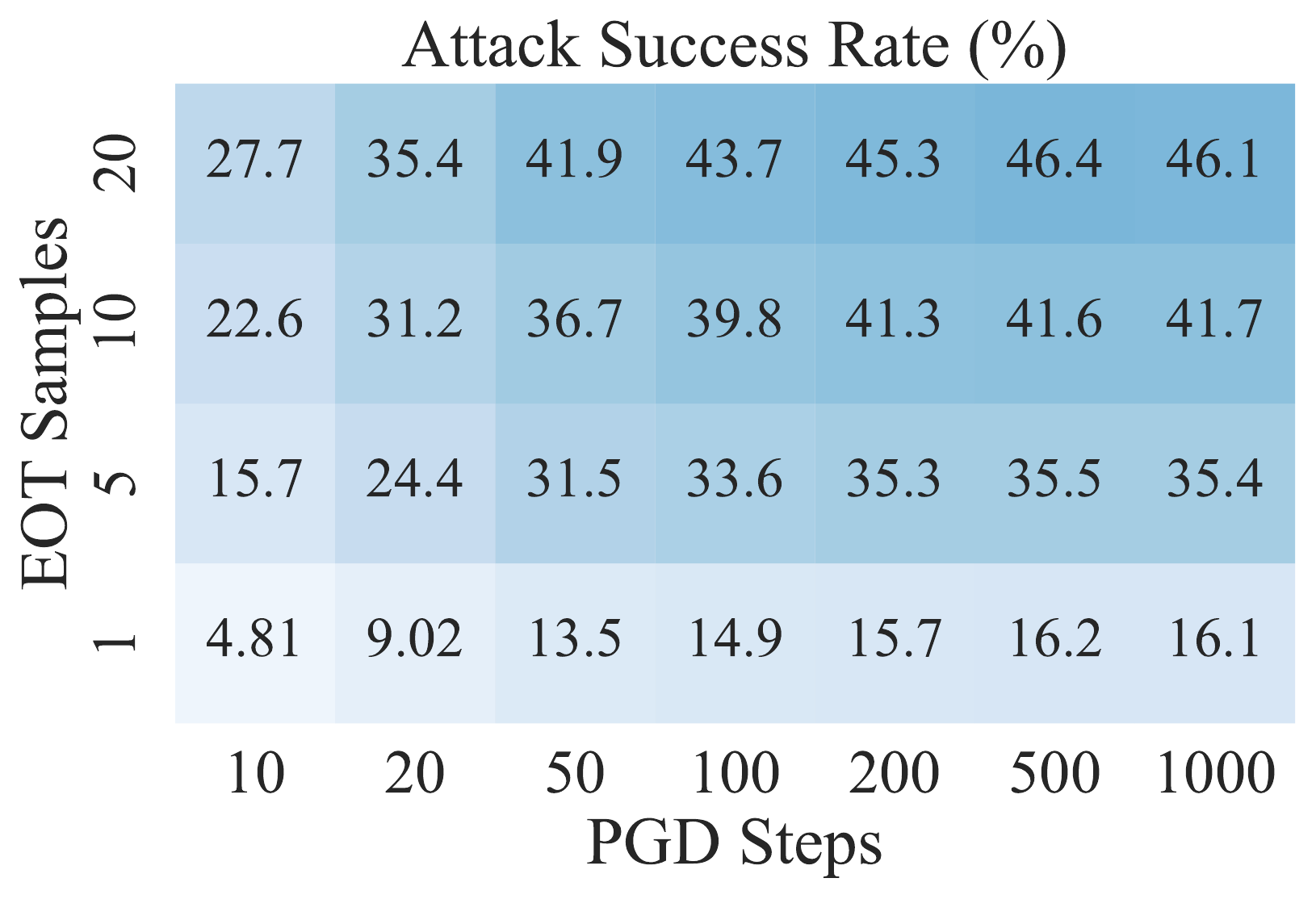}
		\caption{$\sigma = 0.50$ (step size = 2/255)}
	\end{subfigure}
	\begin{subfigure}[t]{0.328\textwidth}
		\includegraphics[width=\linewidth]{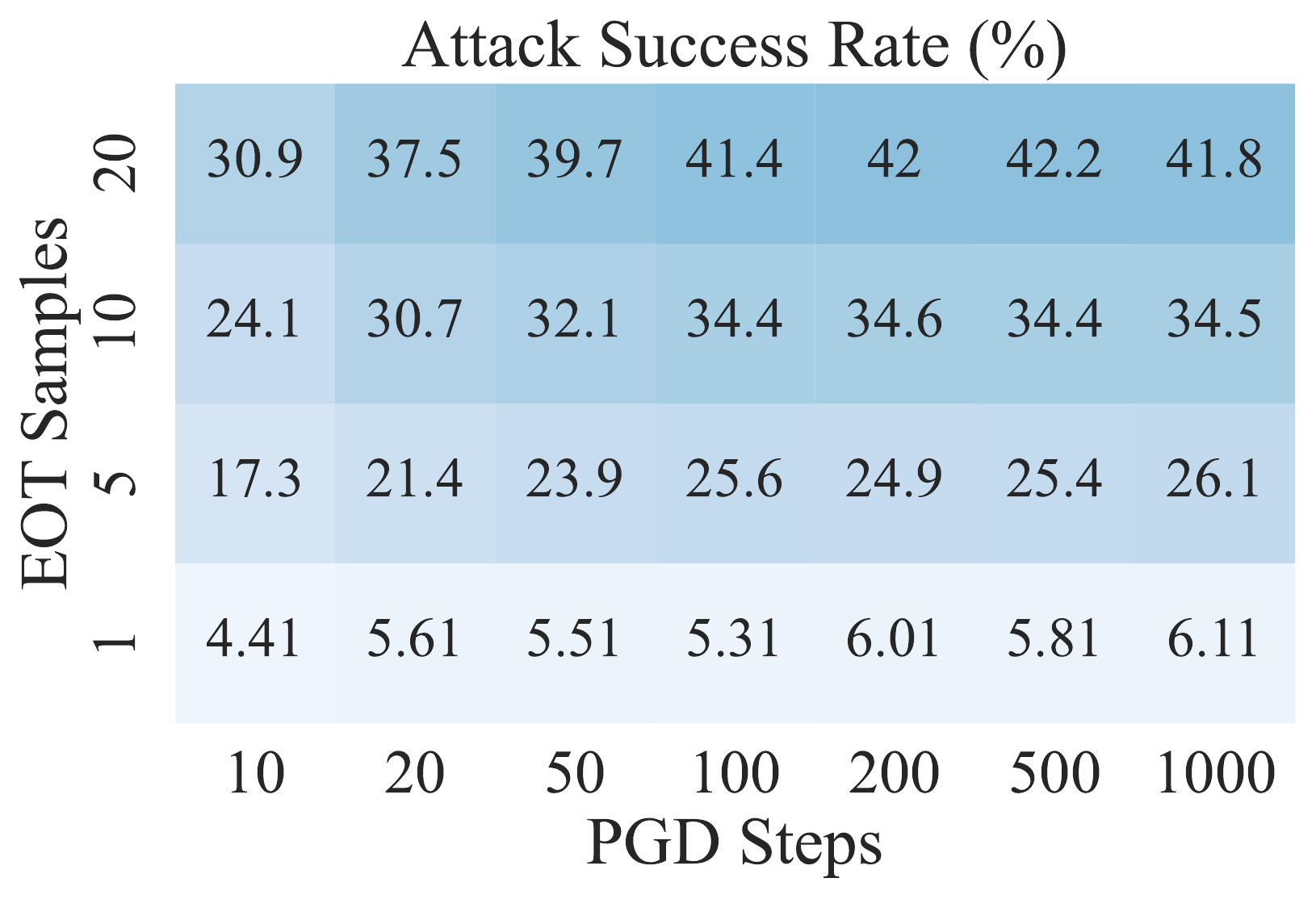}
		\caption{$\sigma = 0.50$ (step size = 4/255)}
	\end{subfigure}
	\begin{subfigure}[t]{0.328\textwidth}
		\includegraphics[width=\linewidth]{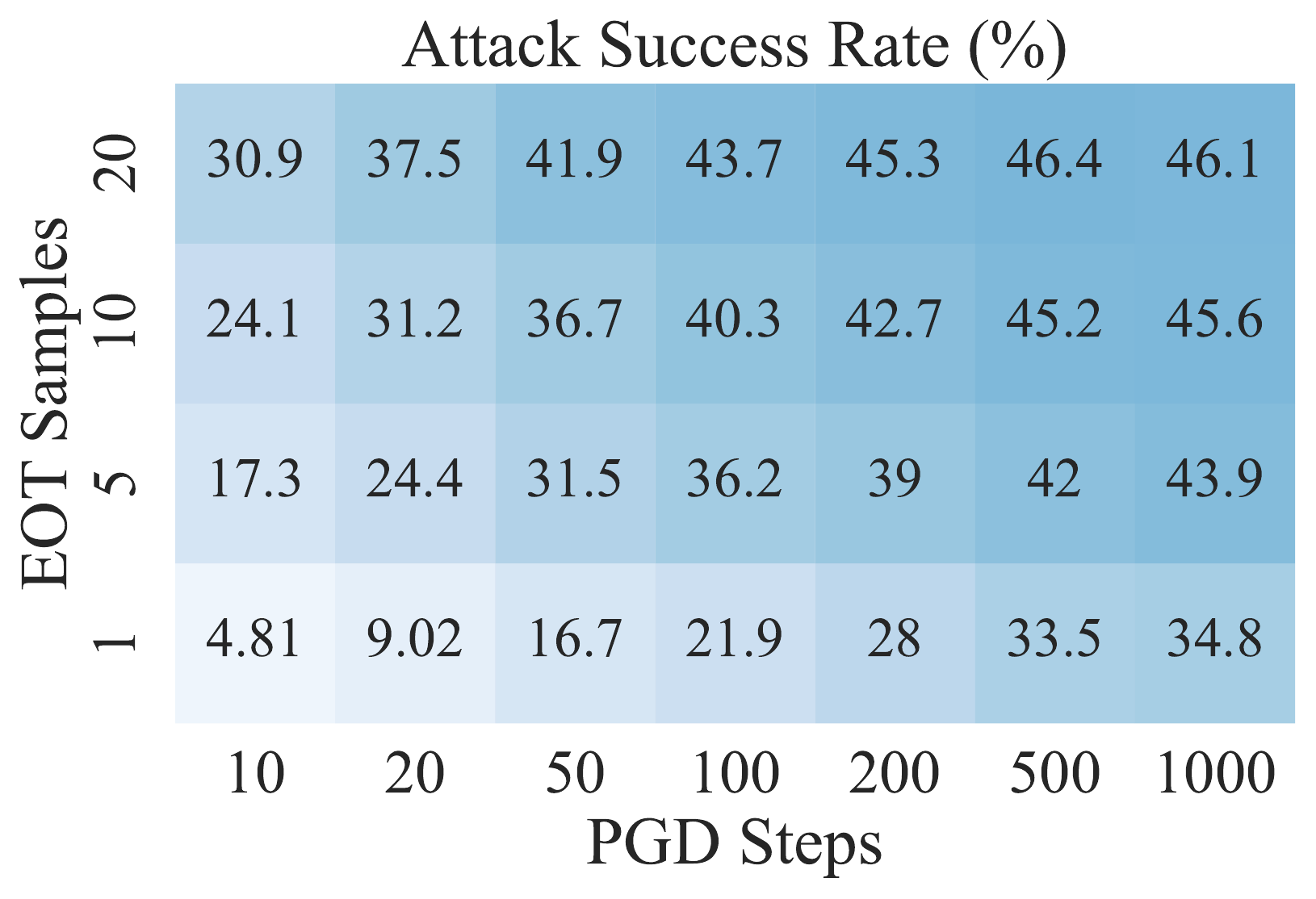}
		\caption{$\sigma = 0.50$ (step size = \textbf{best})}
		\label{fig:attack-smoothing-heatmap-0.50}
	\end{subfigure}
	\caption{Evaluation of randomized smoothing on ImageNet (targeted attacks). PGD performs well on lower variance ($\sigma=0.25$) if running for more steps. For a larger variance ($\sigma=0.50$), applying EOT starts to improve the attack significantly (for a fixed number of gradient computations).}
	\label{fig:attack-smoothing-heatmap}
\end{figure}

\textbf{Stochastic Defenses with Higher Randomness.}
We then examine the randomized smoothing defense that adds Gaussian noise to the input image. Although this defense was originally proposed for certifiable adversarial robustness, we adopt it to evaluate how randomness affects the benefits of applying EOT. Similarly, we evaluate this defense with PGD and EOT of different settings with a focus on the \emph{targeted} attack. The results are shown in \Cref{fig:attack-smoothing-heatmap}.

We observe that EOT starts to improve the attack when the defense has a higher level of randomness. For a fixed number of PGD steps, applying EOT significantly improves the attack in most of the settings. For a fixed attack strength (i.e., number of gradient computations), applying EOT always outperforms standalone PGD. In \Cref{fig:attack-smoothing-heatmap-0.50}, for example, PGD-100 combined with EOT-10 is 5.5\% higher than PGD-1,000 with EOT-1 (40.3\% vs.\ 34.8\%).

\textbf{Takeaways.}
Applying EOT is only beneficial when the defense has sufficient randomness, such as randomized smoothing with $\sigma=0.5$. This observation suggests that stochastic defenses only make standard attacks suboptimal when they have sufficient randomness. However, most existing stochastic defenses did not achieve this criterion, as we showed in \Cref{sec:revisit}. We also provide visualizations of adversarial examples under different settings and CIFAR10 results in \Cref{app:exp:vis,app:exp:cifar10}.

\subsection{Q2: Evaluate the Trade-off between Robustness and Invariance}
\label{sec:exp:tradeoff}

In \Cref{sec:tradeoff}, we present a theoretical setting where the trade-off between robustness and invariance provably exists; stochastic defenses become less robust when the defended model achieves higher invariance to their randomness. Here, we demonstrate this trade-off on realistic datasets, models, and defenses. In particular, we choose defenses with sufficient randomness (achieved in different ways) and compare their performance when being applied to models of different levels of invariance, where the invariance is achieved by applying the defense's randomness to the training data so as to guide the model in learning their transformations.

\begin{figure}[tb]
	\centering
	\begin{subfigure}[t]{0.24\textwidth}
		\includegraphics[width=\linewidth]{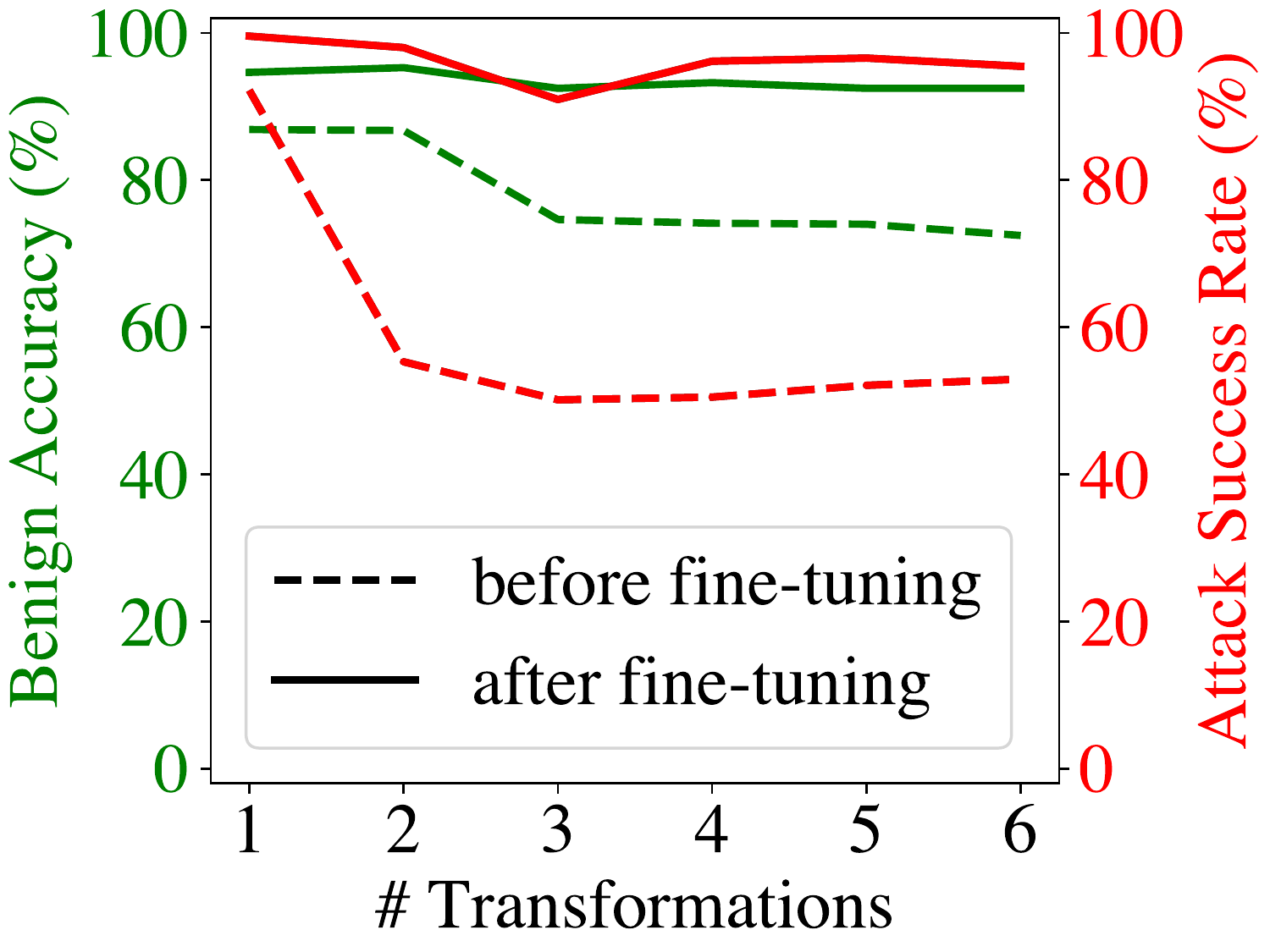}
		\caption{Targeted Attacks \newline \centering{(view by size)}}
		\label{fig:bart-invariance-targeted-var}
	\end{subfigure}
	\begin{subfigure}[t]{0.24\textwidth}
		\includegraphics[width=\linewidth]{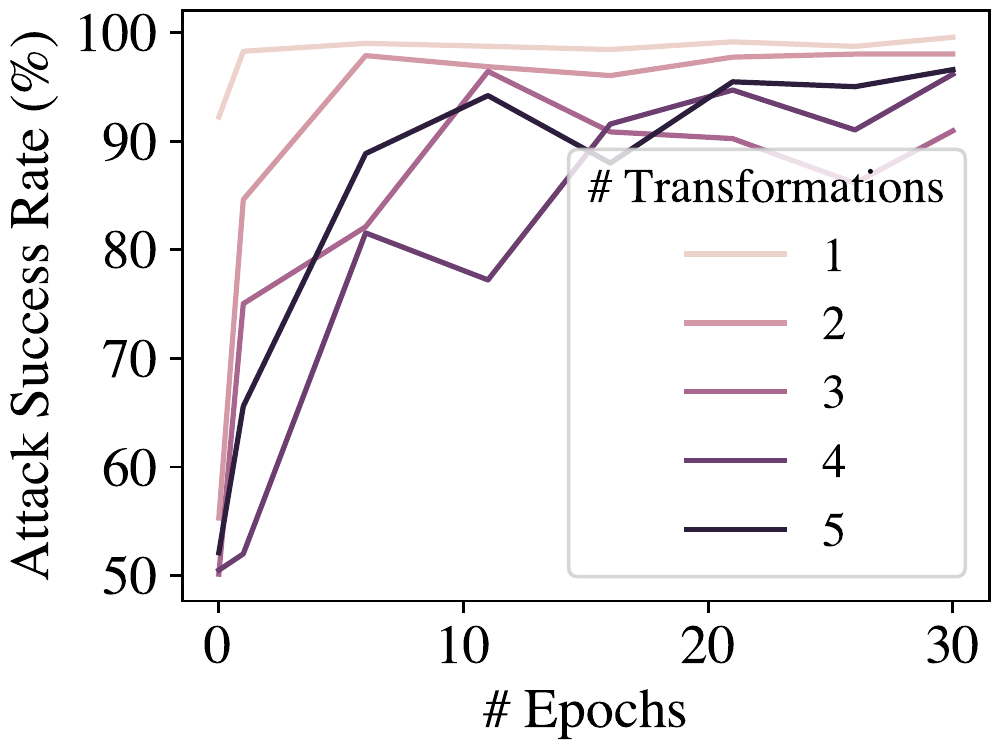}
		\caption{Targeted Attacks \newline \centering{(view by epochs)}}
		\label{fig:bart-invariance-targeted-epoch}
	\end{subfigure}
	\begin{subfigure}[t]{0.24\textwidth}
		\includegraphics[width=\linewidth]{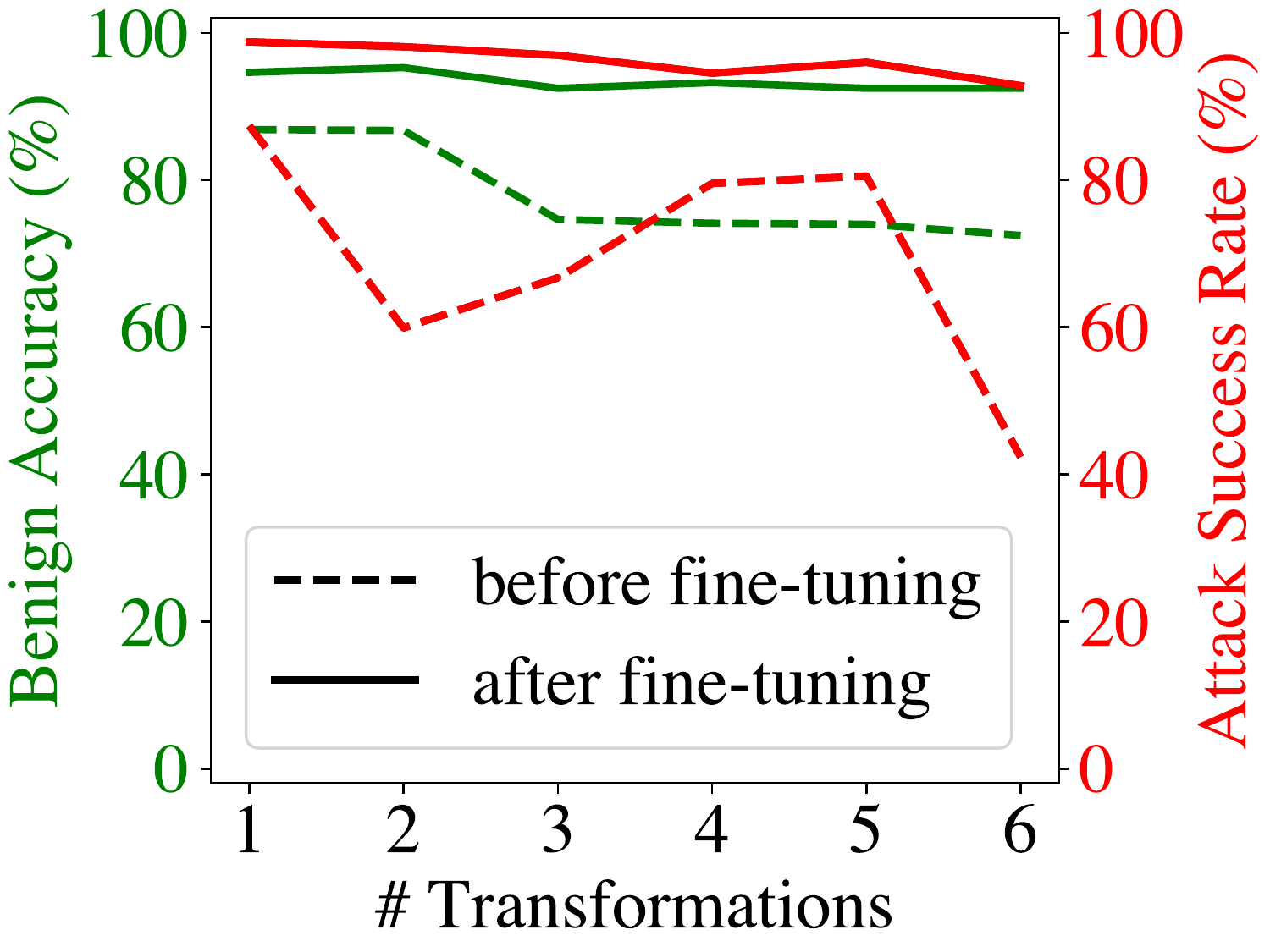}
		\caption{Untargeted Attacks \newline \centering{(view by size)}}
		\label{fig:bart-invariance-untargeted-var}
	\end{subfigure}
	\begin{subfigure}[t]{0.24\textwidth}
		\includegraphics[width=\linewidth]{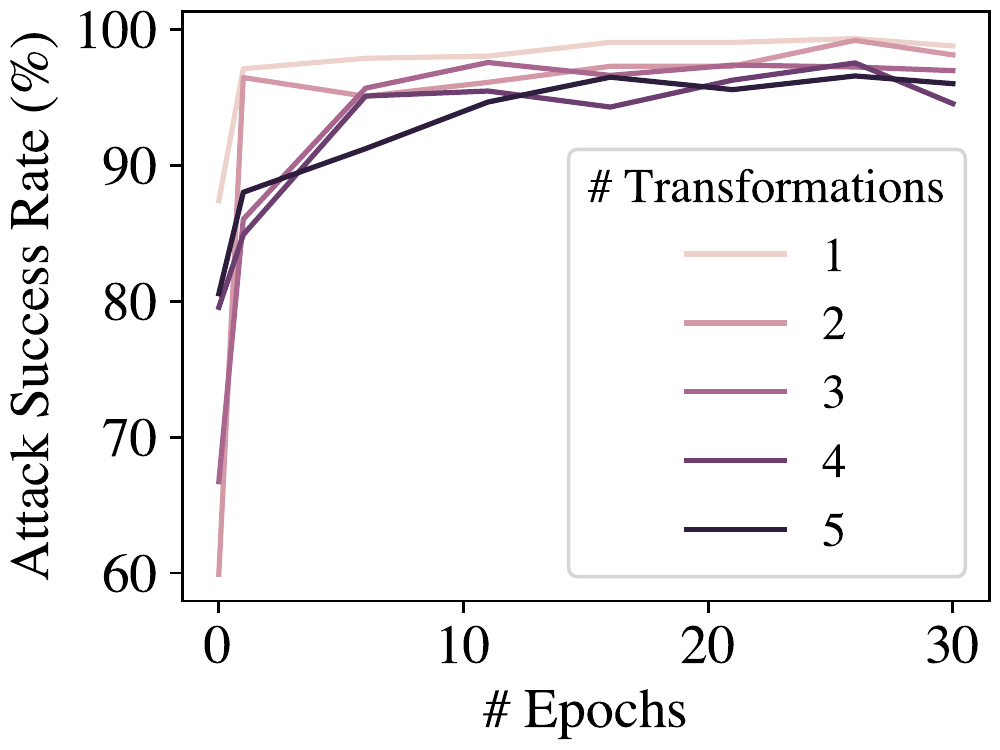}
		\caption{Untargeted Attacks \newline \centering{(view by epochs)}}
		\label{fig:bart-invariance-untargeted-epoch}
	\end{subfigure}
	\caption{Performance of the BaRT defense on ImageNette with different numbers of transformations before and after fine-tuning the model. While the model achieves higher invariance, the defense becomes nearly ineffective\protect\footnotemark, as evident from the top solid red curves in (a) and (c).}
	\label{fig:bart-invariance}
\end{figure}

\textbf{Randomness through Transformations.}
We first examine the BaRT defense, which pre-processes input images with $\kappa$ randomly composited stochastic transformations. It represents defenses aiming to increase randomness through diverse input transformations. 
Since our objective is to demonstrate the trade-off, it suffices to evaluate a subset of BaRT with $\kappa\leq6$ transformations; this also avoids the training cost of evaluating the original BaRT with $\kappa=25$.
\Cref{fig:bart-invariance} shows the performance of this defense with models before and after fine-tuning on its processed training data.

In \Cref{fig:bart-invariance-targeted-var,fig:bart-invariance-untargeted-var}, we first observe that fine-tuning indeed increases the model's invariance to the applied defense's randomness; the utility's dashed green curves are improved to the solid green curves beyond 90\%. However, as the model achieves higher invariance, the defense becomes nearly ineffective; the attack's dashed red curves boost to the solid red curves near 100\%. The same attack's effectiveness throughout the fine-tuning procedure further verifies this observation, as shown in \Cref{fig:bart-invariance-targeted-epoch,fig:bart-invariance-untargeted-epoch}. It shows a clear trade-off between the defense's robustness and the model's invariance. That is, stochastic defenses start to lose robustness when their defended models achieve higher invariance to their transformations.

\begin{figure}[tb]
	\centering
	\begin{subfigure}[t]{0.24\textwidth}
		\includegraphics[width=\linewidth]{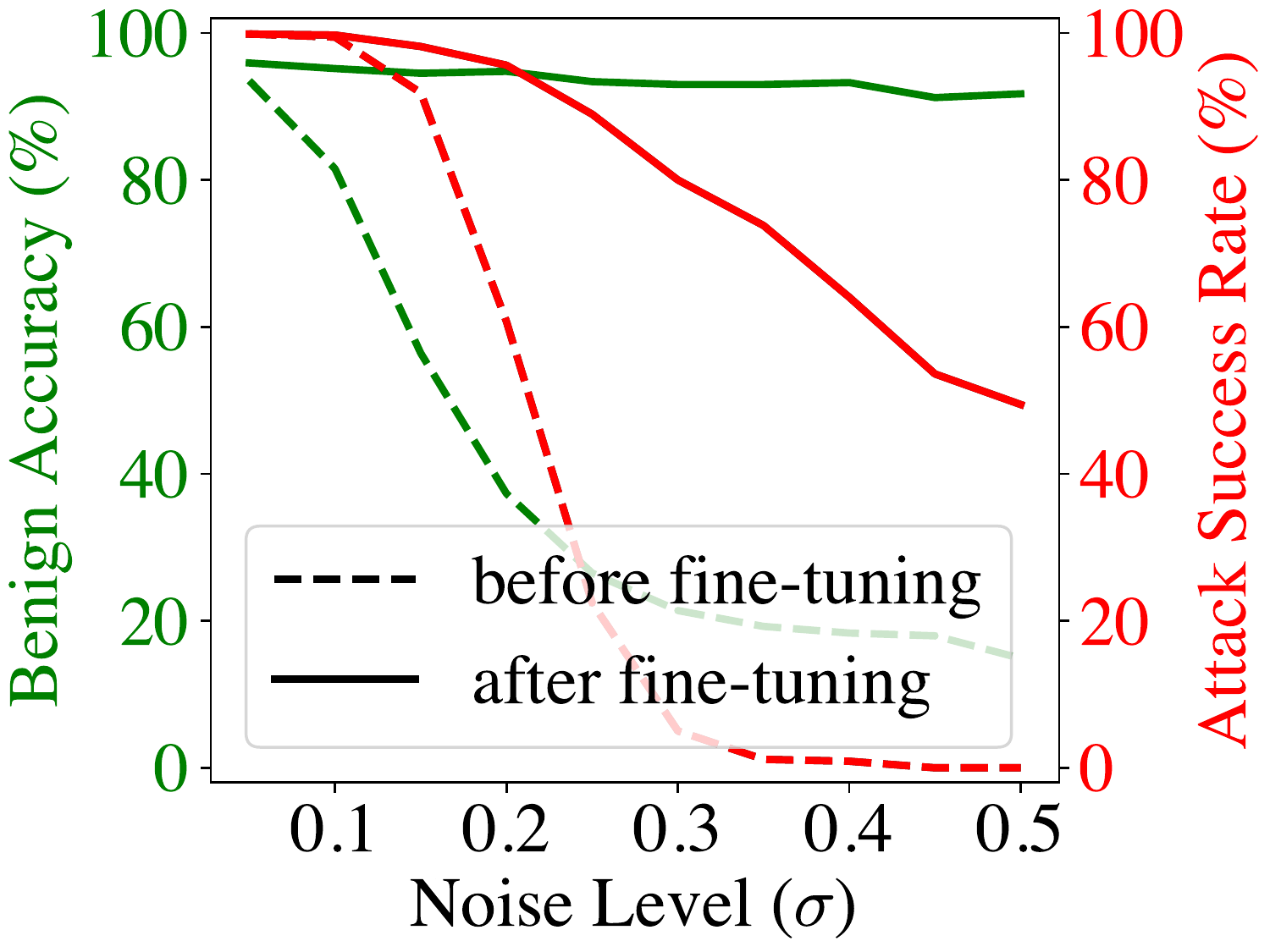}
		\caption{Targeted Attacks \newline \centering{(view by levels)}}
		\label{fig:smoothing-invariance-targeted-var}
	\end{subfigure}
	\begin{subfigure}[t]{0.24\textwidth}
		\includegraphics[width=\linewidth]{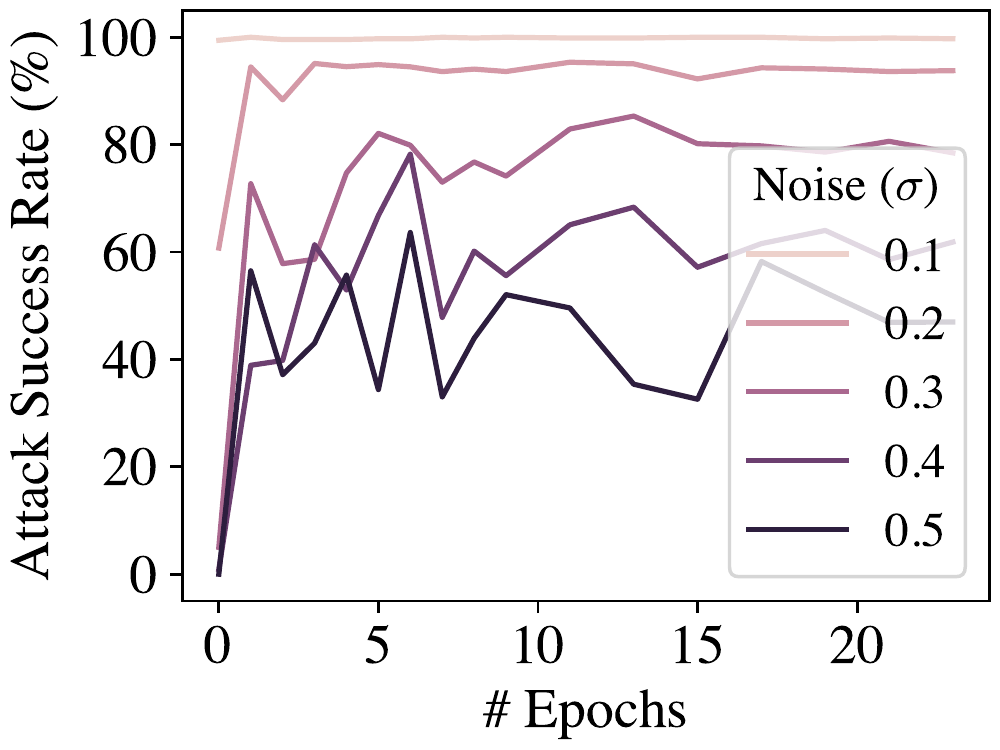}
		\caption{Targeted Attacks \newline \centering{(view by epochs)}}
		\label{fig:smoothing-invariance-targeted-epoch}
	\end{subfigure}
	\begin{subfigure}[t]{0.24\textwidth}
		\includegraphics[width=\linewidth]{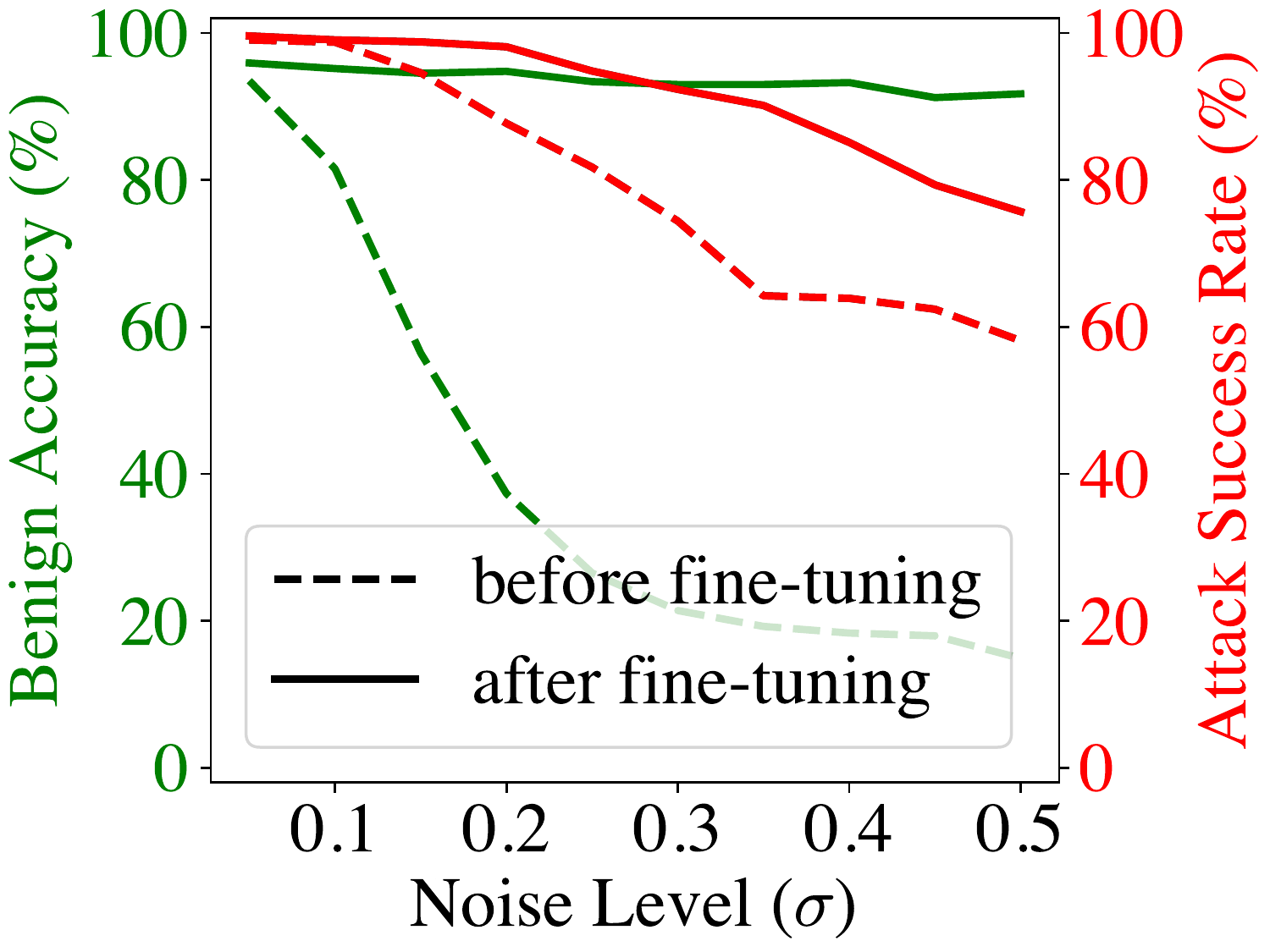}
		\caption{Untargeted Attacks \newline \centering{(view by levels)}}
		\label{fig:smoothing-invariance-untargeted-var}
	\end{subfigure}
	\begin{subfigure}[t]{0.24\textwidth}
		\includegraphics[width=\linewidth]{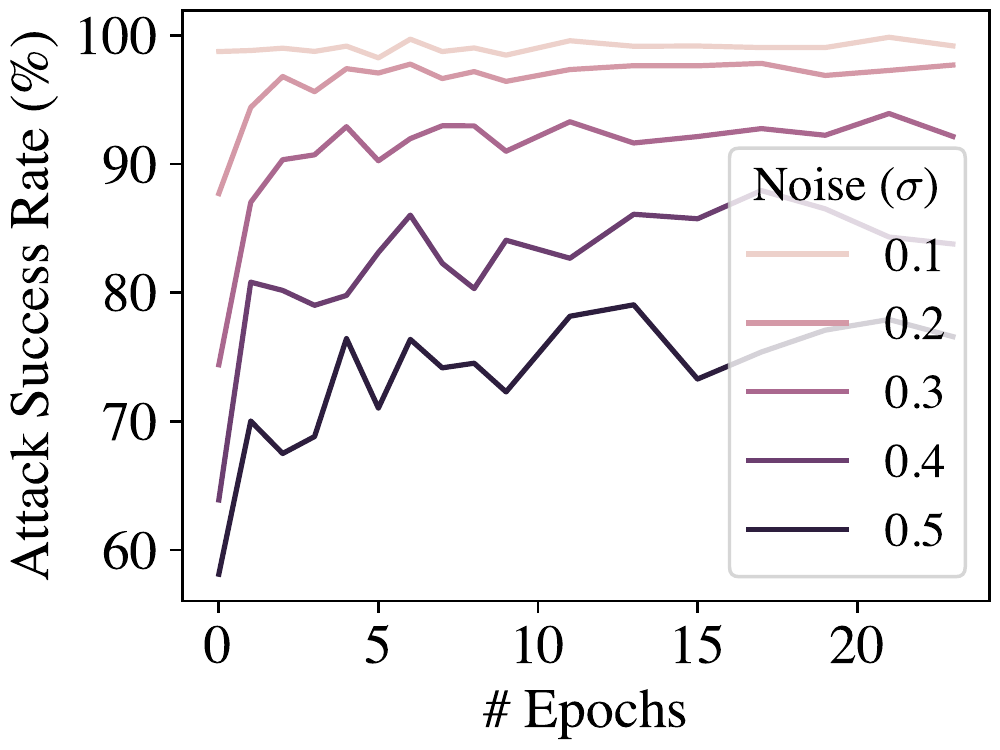}
		\caption{Untargeted Attacks \newline \centering{(view by epochs)}}
		\label{fig:smoothing-invariance-untargeted-epoch}
	\end{subfigure}
	\caption{Performance of the randomized smoothing defense on ImageNette with different noise levels before and after fine-tuning the model. While the model achieves higher invariance, the defense becomes less effective\protect\footnotemark, as evident from the gap between dashed and solid red curves in (a) and (c).}
	\label{fig:smoothing-invariance}
\end{figure}

\textbf{Randomness through Noise Levels.}
We then examine the randomized smoothing defense that adds Gaussian noise to the input image. Unlike BaRT's diverse transformations, randomized smoothing increases randomness directly through the added noise's variance $\sigma^2$. This allows us to rigorously increase the randomness without unexpected artifacts like non-differentiable components. We evaluate the performance of this defense ($\sigma\leq0.5$) with models before and after fine-tuning on training data perturbed with designated Gaussian noise. The results are shown in \Cref{fig:smoothing-invariance}.

In \Cref{fig:smoothing-invariance-targeted-var,fig:smoothing-invariance-untargeted-var}, fine-tuning improves the model's invariance, but the defense also becomes significantly weaker during this process. For example, the targeted attack is nearly infeasible when the model is variant to the large noise ($\sigma\geq0.3$), yet is significantly more effective when the model becomes invariant. The fine-tuning process in \Cref{fig:smoothing-invariance-targeted-epoch,fig:smoothing-invariance-untargeted-epoch} also verifies that stochastic defenses become weaker when their defended models become more invariant to their randomness.

\textbf{Takeaways.} For both the BaRT and the randomized smoothing defense, we observe a clear trade-off between the defense's robustness and the model's invariance to randomness, especially in the targeted setting. In particular, we find that stochastic defenses lose adversarial robustness when their defended models achieve higher invariance to their randomness. Our finding implies that such defenses would become ineffective when their defended models are perfectly invariant to their randomness.

\footnotetext{The defense may not grow stronger with more transformations, which is a drawback of BaRT that we will discuss in \Cref{app:exp:defenses}. Yet, our evaluations focus on the fact that solid curves are above the dashed curves.}

\footnotetext{One may also observe a trade-off between robustness and \emph{utility} by examining the curve's horizontal trend. However, we focus on the trade-off between robustness and \emph{invariance}, which manifests in the vertical gap.}% between dashed and solid curves of each metric.}

\section{Discussions}
\label{sec:discussions}

In this section, we discuss several questions that arose from our study of stochastic pre-processing defenses. Discussions about extensions, limitations, and broader topics can be found in \Cref{app:discuss}.

\textbf{What do stochastic pre-processing defenses really do?}
We show that stochastic pre-processing defenses do not introduce inherent robustness to the prediction task. Instead, they shift the input distribution through randomness and transformations, which results in variance and introduces errors during prediction. The observed ``robustness'', in an unusual meaning for this literature, is a result of these errors. This is fundamentally different from the inherent robustness provided by adversarial training~\cite{pgd}. Although defenses like adversarial training still cost accuracy~\cite{tradeoff-rob-acc-1,tradeoff-rob-acc-2}, they do not intentionally introduce errors like stochastic pre-processing defenses.

\textbf{What are the concrete settings that stochastic pre-processing defenses work?}
These defenses \emph{do make} the attack harder when the adversary has only limited knowledge of the defense's transformations, e.g., in a low-query setting. In such a case, the defense practically introduces noise to the attack's optimization procedure, making it difficult for a low-query adversary to find adversarial examples that consistently cross the probabilistic decision boundary. However, it is still possible for the adversary to infer pre-processors in a black-box model and compute their expectation locally~\cite{scaling,interplay}, unless the randomization space changes over time. Our theoretical analysis considers a powerful adversary with full knowledge of the defense's randomization space; hence it can optimize directly towards the defended model's decision boundary in expectation. The other setting is randomized smoothing, which remains effective in certifying the inherent robustness of a given decision.

\textbf{What are the implications for future research?}
Our work suggests that future defenses should try to decouple robustness and invariance; that is, either avoid providing robustness by introducing variance to the added randomness or the variance only applies to adversarial inputs. This implication is crucial as the research community continues improving defenses through more complicated transformations. For example, in parallel to our work, DiffPure~\cite{diffpure} adopts a complicated stochastic diffusion process to purify the inputs. However, fully understanding DiffPure's robustness requires substantial effort due to its complications and high computational costs, as we will discuss in \Cref{app:discuss:diffpure}

\textbf{How should we improve stochastic defenses?}
Stochastic defenses should rely on randomness that exploits the properties of the prediction task. One promising approach is dividing the problem into orthogonal subproblems. For example, some speech problems like keyword spotting are inherently divisible in the spectrum space~\cite{ahmed2022towardsmorerobustkws}, and vision tasks are divisible by introducing different modalities~\cite{singlesource}, independency~\cite{certifiedpatch}, or orthogonality~\cite{trs}. In such cases, randomization forces the attack to target all possible (independent) subproblems, where the model performs well on each (independent and) non-transferable subproblem. As a result, defenses can decouple robustness and invariance, hence reducing the effective attack budget and avoiding the pitfall of previous randomized defenses. While systematic guidance for designing defenses (and their attacks) remains an open question, we summarize some critical insights along this direction in \Cref{app:discuss:guidance}.

\textbf{What are the implications for adaptive attackers?}
Our findings suggest that an adaptive attacker needs to consider the spectrum of available standard attack algorithms, instead of just focusing on a given attack algorithm because of the defense's design. As we discover in this paper, EOT can be unnecessary for seemingly immune stochastic defenses, yet its application to break these said defenses gives a false impression about their security against weak attackers. When evaluating the robustness of a defense, the adaptive attack should start by tuning standard approaches, before resorting to more involved attack strategies. This approach helps us to identify the minimally capable attack that breaks the defense and develop a better understanding of the defense's fundamental weaknesses.

\section{Conclusion}
In this paper, we investigate stochastic pre-processing defenses and explain their limitations both empirically and theoretically. We show that most stochastic pre-processing defenses are weaker than previously thought, and recent defenses that indeed exhibit more randomness still face a trade-off between their robustness and the model's invariance to their transformations. While defending against adversarial examples remains an open problem and designing proper adaptive evaluations is arguably challenging, we demonstrate that stochastic pre-processing defenses are fundamentally flawed in their current form. Our findings suggest that future work will need to find new ways of using randomness that decouples robustness and invariance.

\section*{Acknowledgement}
We thank all anonymous reviewers for their insightful comments and feedback. We would like to acknowledge our sponsors, who support our research with financial and in-kind contributions: the DARPA GARD program under agreement number 885000, NSF through award CNS-2003129, CIFAR through the Canada CIFAR AI Chair program, and NSERC through the Discovery Grant and COHESA Strategic Alliance. Resources used in preparing this research were provided, in part, by the Province of Ontario, the Government of Canada through CIFAR, and companies sponsoring the Vector Institute. We would like to thank members of the CleverHans Lab for their feedback.

\clearpage
{\small\bibliography{main}}
\bibliographystyle{plainnat}

\clearpage
\section*{Checklist}

\begin{enumerate}

\item For all authors...
\begin{enumerate}
  \item Do the main claims made in the abstract and introduction accurately reflect the paper's contributions and scope?
    \answerYes{}
  \item Did you describe the limitations of your work?
    \answerYes{See \Cref{app:limitation}.}
  \item Did you discuss any potential negative societal impacts of your work?
    \answerYes{See \Cref{app:limitation}.}
  \item Have you read the ethics review guidelines and ensured that your paper conforms to them?
    \answerYes{}
\end{enumerate}

\item If you are including theoretical results...
\begin{enumerate}
  \item Did you state the full set of assumptions of all theoretical results?
    \answerYes{We outline all details for theoretical results in \Cref{app:tradeoff}.}
        \item Did you include complete proofs of all theoretical results?
    \answerYes{We include complete proofs in \Cref{app:tradeoff}.}
\end{enumerate}

\item If you ran experiments...
\begin{enumerate}
  \item Did you include the code, data, and instructions needed to reproduce the main experimental results (either in the supplemental material or as a URL)?
    \answerYes{Our code is available in the supplementary material.}
  \item Did you specify all the training details (e.g., data splits, hyperparameters, how they were chosen)?
    \answerYes{We specify brief settings in \Cref{sec:exp:settings} and provide complete details in \Cref{app:exp}.}
        \item Did you report error bars (e.g., with respect to the random seed after running experiments multiple times)?
    \answerNo{We study stochastic defenses, whose prediction procedure is already a majority vote over multiple times.}
        \item Did you include the total amount of compute and the type of resources used (e.g., type of GPUs, internal cluster, or cloud provider)?
    \answerYes{See \Cref{app:exp}.}
\end{enumerate}

\item If you are using existing assets (e.g., code, data, models) or curating/releasing new assets...
\begin{enumerate}
  \item If your work uses existing assets, did you cite the creators?
    \answerYes{}
  \item Did you mention the license of the assets?
    \answerYes{}
  \item Did you include any new assets either in the supplemental material or as a URL?
    \answerNA{We only included our own code for evaluation.}
  \item Did you discuss whether and how consent was obtained from people whose data you're using/curating?
    \answerNA{We only used public datasets as discussed in \Cref{sec:exp:settings}.}
  \item Did you discuss whether the data you are using/curating contains personally identifiable information or offensive content?
    \answerNA{We did not use such data.}
\end{enumerate}

\item If you used crowdsourcing or conducted research with human subjects...
\begin{enumerate}
  \item Did you include the full text of instructions given to participants and screenshots, if applicable?
    \answerNA{}
  \item Did you describe any potential participant risks, with links to Institutional Review Board (IRB) approvals, if applicable?
    \answerNA{}
  \item Did you include the estimated hourly wage paid to participants and the total amount spent on participant compensation?
    \answerNA{}
\end{enumerate}

\end{enumerate}

\clearpage
\appendix
\section*{Appendix: On the Limitations of Stochastic Pre-processing Defenses}
\renewcommand{\contentsname}{Table of Contents}
\definecolor{linkcolor}{RGB}{0,0,0}
\tableofcontents
% \startcontents[sections]
% \printcontents[sections]{ }{0}{\section*{Table of Contents}}
\definecolor{linkcolor}{HTML}{ED1C24}
\clearpage

\addtocontents{toc}{\protect\setcounter{tocdepth}{2}}  % enable toc for appendix
\section{More Preliminaries}
\label{app:prelim}

In this section, we expand on the preliminaries provided in \Cref{sec:prelim}.

\subsection{Aggregation Strategies for Stochastic Classifiers}
\label{app:prelim:aggregation}

As the stochastic classifier $\RandomizedClassifier$ returns varied outputs even for a fixed input, it needs to determine the final prediction by aggregating $\PredictionTime$ independent inferences with a particular strategy.

\textbf{Majority Vote.}
The most commonly used strategy is \emph{majority vote}, which can be formulated as
\begin{equation}
\label{eq:rule-vote}
\DecisionClassifier_{\RandomParameter}^\mathrm{vote}(\bx) \coloneqq
\argmax_{y\in\LabelSpace}
\sum_{i=1}^\PredictionTime \indicator\s[\bigg]{\RandomizedDecisionClassifier[\RandomParameter_i](\bx) = y},
\end{equation}
where $\s{\RandomParameter_i}_{i=1}^\PredictionTime\iid\RandomizationSpace$ are sampled parameters. We adopt this strategy when the defended classifier computes its prediction. But when attacking a defended classifier, we use the gradient obtained from the original stochastic classifier $\RandomizedClassifier$.

\textbf{Match All.}
A more restricted strategy is \emph{match all}, which requires all predictions to be identical:
\begin{equation}
\label{eq:rule-all}
\DecisionClassifier_{\RandomParameter}^\mathrm{all}(\bx) \coloneqq
y, \quad \mathrm{s.t.} \quad \sum_{i=1}^n \indicator\s[\bigg]{\RandomizedDecisionClassifier[\RandomParameter_i](\bx) = y} = n,
\end{equation}
where $\s{\RandomParameter_i}_{i=1}^n\iid\RandomizationSpace$ are sampled parameters. This strategy \emph{rejects} the input if the condition is not satisfied, which can be used as a strict setting for targeted attacks. We do not choose this strategy in our work because it is overly strict and is hard to satisfy, even for benign inputs.

\textbf{Averaged Logits.}
One may also determine the label from averaged logits over multiple inferences:
\begin{equation}
\DecisionClassifier_{\RandomParameter}^\mathrm{logits}(\bx) \coloneqq
\argmax_{y\in\LabelSpace}
\frac{1}{\PredictionTime}\sum_{i=1}^\PredictionTime \RandomizedClassifierOutputOf[\RandomParameter_i]{y}(\bx),
\end{equation}
where $\s{\RandomParameter_i}_{i=1}^n\iid\RandomizationSpace$ are sampled parameters. \citet{aggmopgd} leverage this strategy to design adaptive attacks against the BaRT~\cite{bart} defense. Still, we do not use this strategy because our main objective is not to break defenses but to analyze their fundamental weaknesses. We only apply the prediction strategy when using a stochastic classifier to evaluate a given set of inputs.

\textbf{The Choice of Prediction Times.}
The choice of $\PredictionTime$ typically depends on the stochastic classifier's variance to its randomness. In our setting, the randomness comes from the applied pre-processing defense $\InputTransformation_{\RandomParameter}$ with random variable $\RandomParameter$ drawn from the randomization space $\RandomizationSpace$. When the randomization space is small, it suffices to set $\PredictionTime=1$ for most such defenses~\cite{input-transformation,random-rescaling}. For defenses with a slightly larger randomization space, they can set $\PredictionTime=30$, for example for randomized cropping~\cite{input-transformation}. Finally, defenses with even larger randomization spaces set $\PredictionTime=500$ or more~\cite{smoothing,bart}. We fix $\PredictionTime=500$ in our main experiments in \Cref{sec:exp} for consistency.

\subsection{Formulation of Projected Gradient Descent}
\label{app:prelim:attack}

In this paper, we mainly use PGD~\cite{pgd} to evaluate the robustness of a stochastically defended model. Given a benign example $\bx^0$ and its ground-truth label $\y$, each iteration of the \emph{untargeted} PGD attack with \LL{\infty} norm budget $\epsilon$ can be formulated as
\begin{equation}
	\bx^{i+1} \gets \bx^{i} + \alpha \cdot \sgn\s[\big]{\grad{\mathcal{L}\p[\big]{\RandomizedClassifier(\bx^{i}), y}}},
\end{equation}
where $\alpha$ is the step size, $\mathcal{L}$ is the loss function, and each iteration is projected to the $\ell_\infty$ ball around $\bx^0$ of radius $\epsilon$. As for \emph{targeted} PGD attacks with a target label $\y^\prime$, the above iteration becomes
\begin{equation}
	\bx^{i+1} \gets \bx^{i} - \alpha \cdot \sgn\s[\big]{\grad{\mathcal{L}\p[\big]{\RandomizedClassifier(\bx^{i}), y^\prime}}},
\end{equation}
where we switch the optimizing direction and the label for computing the loss.

Similarly, the untargeted attack with \LL{2} norm budget $\epsilon$ is formulated as
\begin{equation}
	\bx^{i+1} \gets \bx^{i} + \alpha \cdot \norm[\big]{\grad{\mathcal{L}\p[\big]{\RandomizedClassifier(\bx^{i}), y}}}_2,
\end{equation}
where each iteration is projected to the \LL{2} norm ball around $\bx^0$ of radius $\epsilon$, and the targeted attack
\begin{equation}
	\bx^{i+1} \gets \bx^{i} - \alpha \cdot \norm[\big]{\grad{\mathcal{L}\p[\big]{\RandomizedClassifier(\bx^{i}), y^\prime}}}_2.
\end{equation}

\subsection{Quantifying the Strength of White-box Attacks}
\label{app:prelim:strength}

In this work, we consider attacks with different combinations of PGD steps and EOT samples, denoted by PGD-$k$ and EOT-$m$. For evaluations of deterministic defenses, quantifying the strength of PGD attacks by the number of steps $k$ is valid. However, this quantification is not informative enough when the evaluated defense is stochastic and involves EOT. For example, it is hard to tell whether PGD-1 with EOT-100 or PGD-100 with EOT-1 has more strength in terms of the number of steps. For a fair comparison between such attacks, we quantify their strength by the total \emph{number of gradient computations}, defined as
\begin{equation}
	\texttt{strength}(\text{PGD-}k, \text{EOT-}m) \coloneqq k \times m.
\end{equation}

This concept is similar to the \emph{query budget} in the black-box setting. Although we \emph{do not} constrain white-box attacks like this, it allows for a fair comparison between attacks with different settings. For example, we can now argue that the two attacks above have the same strength due to $k\times m = 100$.

Moreover, the above quantification has realistic implications for the white-box attack's computational cost under finite computing resources (w.r.t.\ the number of evaluated samples). In such a case, the computation of EOT is not parallelizable by batching the EOT samples. For example, when attacking $B$ samples with a maximally possible batch size of $B$, the attacker has to compute the gradients for $k\times m$ batches. Only when the maximally  possible batch size becomes $m\times B$, the attacker can parallelize the EOT samples and only needs to compute gradients for $k$ batches.

\paragraph{Potential Optimality Analysis.}
Although we evaluate various combinations of PGD-$k$ and EOT-$m$, we are not interested in finding a heuristic for the best combination for two reasons. Firstly, this discussion is beyond the scope of the question that we want to answer in \Cref{sec:exp:eot}. Secondly, white-box attackers in the real world have sufficient incentive to adopt a sufficiently large value of $k$ and $m$ (to make sure their attack converges), regardless of the potential optimal choice.

However, it is still possible to correlate the choice of $k$ and $m$ with the convergence rate of stochastic gradient descent (SGD). For example, it is well-known that the convergence rate of SGD can be affected by the estimated gradient's variance~\cite{ghadimi2013stochastic}, and this variance is again affected by the number of EOT samples $m$ we choose due to the central limit theorem. As a result, one can analyze the attack's convergence behavior with different choices of PGD-$k$ and EOT-$m$. Still, this discussion is beyond the scope of this work and is more beneficial in the context of black-box attacks.

\section{Experiment Setup: Most Stochastic Defenses Lack Sufficient Randomness}
\label{app:revisit}

In \Cref{sec:revisit}, we replicate the evaluation of five previous stochastic defenses from \citet{bpda}\footnote{\url{https://github.com/anishathalye/obfuscated-gradients}} and \citet{adaptive}\footnote{\url{https://github.com/wielandbrendel/adaptive_attacks_paper}} without applying EOT. Here, we provide more details of these defenses and their evaluation settings.

\textbf{Case Study: Random Rotation.} In this case study, we evaluate this defense on 1,000 randomly chosen ImageNet images and a pre-trained ResNet-50 model. The settings are consistent with our main evaluation described later in \Cref{app:exp}.

\subsection{Randomized Image Cropping~\cite{input-transformation}}

\textbf{Defense Details.}
This defense randomly crops $m=30$ patches of size $90\times90$ from each input image of size $299\times299$. These patches are sent to the classifier, and the final prediction is a majority vote over the predictions of these patches.

\textbf{Original Evaluation.}
\citet{bpda} evaluate this defense with an \LL{2}-bounded adversary under the root-mean-square perturbation budget of 0.05. Their attack decreases the classification loss (averaged over $m$ patches) using gradient descent with 1,000 iterations and a learning rate 0.1. They decrease the accuracy of an Inception-v3~\cite{inception} target model to 0\% (i.e., 100\% attack success rate) on 1,000 randomly sampled ImageNet~\cite{imagenet} images.

\textbf{Our Ablation Study.}
We replicate this evaluation by setting $m=1$ when running the attack (the final defense still uses $m=30$ patches). This means that we only attack a randomly cropped small patch from the entire image at each iteration. We then change the learning rate from 0.1 to 0.001 and are able to achieve 99.0\% attack success rate.

\subsection{Randomized Image Rescaling~\cite{random-rescaling}}

\textbf{Defense Details.}
This defense randomly rescales the input image of size $299\times299$ to $r\times r$, where $r\in[200,331)$ is chosen uniformly at random, and then randomly pads the image with zeros to size $331\times 331$. The resulting padded image is sent to the classifier for one-time prediction.

\textbf{Original Evaluation.}
\citet{bpda} evaluate this defense with an \LL{\infty}-bounded adversary under the perturbation budget of 8/255. They generate adversarial examples using PGD-1000 with a step size of 0.1, where each step applies EOT-30 to compute the gradients averaged over 30 samples processed from the evaluated input image. They decrease the accuracy of an Inception-v3~\cite{inception} target model to 0\% (i.e., 100\% attack success rate) on 1,000 randomly sampled ImageNet~\cite{imagenet} images.

\textbf{Our Ablation Study.}
We replicate this evaluation with PGD-200 and EOT-1, with all the other parameters unchanged. We are still able to achieve a 100\% attack success rate in this case.

\subsection{Randomized Activation Pruning~\cite{activation-pruning}}

\textbf{Defense Details.}
This defense randomly drops (zeros out) some neurons of each layer with probability proportional to their absolute value. The defense considers several levels of probability, and we use the setting used by \citet{bpda}.

\textbf{Original Evaluation.}
\citet{bpda} evaluate this defense with an \LL{\infty}-bounded adversary under the perturbation budget of 8/255. They decrease the margin between the correct label's logit and the wrong label's logit with gradient descent using the Adam~\cite{adam} optimizer. The attack runs for 500 steps with a learning rate 0.1, where each iteration averages the gradient over 10 samples. The attack achieves 100\% success rate on an Inception-v3~\cite{inception} target model and the CIFAR-10~\cite{cifar10} dataset. 

\textbf{Our Ablation Study.}
We replicate this evaluation by simply setting the number of EOT samples to 1 and are still able to obtain 100\% success rate.

\subsection{Discontinuous Activation~\cite{kwinners}}

\textbf{Defense Details.}
This defense replaces the standard ReLU activation function inside the neural network with a discontinuous function, so that only the $k$ largest elements are preserved. Although this defense is not stochastic by itself, we evaluate it because the existing evaluation relies heavily on the application of EOT.

\textbf{Original Evaluation.}
\citet{adaptive} evaluate this defense with several techniques that approximate the correct gradient. For each input, they estimate the average local gradient with $m=20,000$ random perturbations drawn from a standard normal distribution with standard deviation $\epsilon=8/255$. Given this estimated gradient, they consider an \LL{\infty}-bounded adversary with perturbation budget 8/255 and run the PGD attack with 100 steps with step size 0.01. Their evaluation code uses a fine-grained choice of $m$, which is set to 100, 1K, and 20K at the 1st, 20th, and 40th iterations, respectively. We report 1K in the main paper.

As a result, their attack achieves 100\% attack success rate on a ResNet-18~\cite{kwinners} model from the original defense and the CIFAR-10~\cite{cifar10} dataset.

\textbf{Our Ablation Study.}
We replicate this evaluation by moving all gradient computations from the estimation side $m$ to the attack's iteration side $k$. That is, instead of running PGD-100 with EOT-1K (100K gradient computations), we run PGD-40K and EOT-1 (40K gradient computations). This setting achieves 98.4\% success rate on the same model and dataset. In \Cref{app:exp:results:lr}, we discuss an interesting observation when evaluating this defense; it shows that PGD may capture randomness as well as EOT with a carefully fine-tuned learning rate.

\subsection{Statistical Detection~\cite{odds}}

\textbf{Defense Details.}
This defense is a statistical test for detecting adversarial examples. It checks if a given input image is overly robust under Gaussian noise, which is a property of adversarial examples generated by PGD~\cite{pgd} and C\&W~\cite{cw}; benign images are sensitive to such noise.
 
\textbf{Original Evaluation.}
\citet{adaptive} evaluate this defense with logit matching. Specifically, they generate adversarial examples with a logit that matches a given target image's logit in terms of (1) low mean squared error (MSE) distance and (2) similar robustness under the Gaussian noise. Their attack combines the above two objectives and runs for 100 steps with a learning rate 0.2/255, where the robustness under the Gaussian noise is measured under $m=40$ samples. The resulting PGD-100 and EOT-40 attack achieves 100\% success rate on a target ResNet~\cite{resnet} model and 1,000 randomly sampled ImageNet~\cite{imagenet} images.

\textbf{Our Ablation Study.}
We replicate the evaluation by moving all EOT samples to PGD steps. That is, we run PGD-4K and EOT-1 with a learning rate 0.1/255. As a result, our attack achieves 96.1\% success rate, only 3.9\% lower than the attack using EOT. We did not tune the step size further.

%\clearpage
\section{Theoretical Analysis: Trade-off between Robustness and Invariance}
\label{app:tradeoff}

\begin{figure}[t]
\centering
\begin{subfigure}{0.325\linewidth}
	\includegraphics[width=\linewidth]{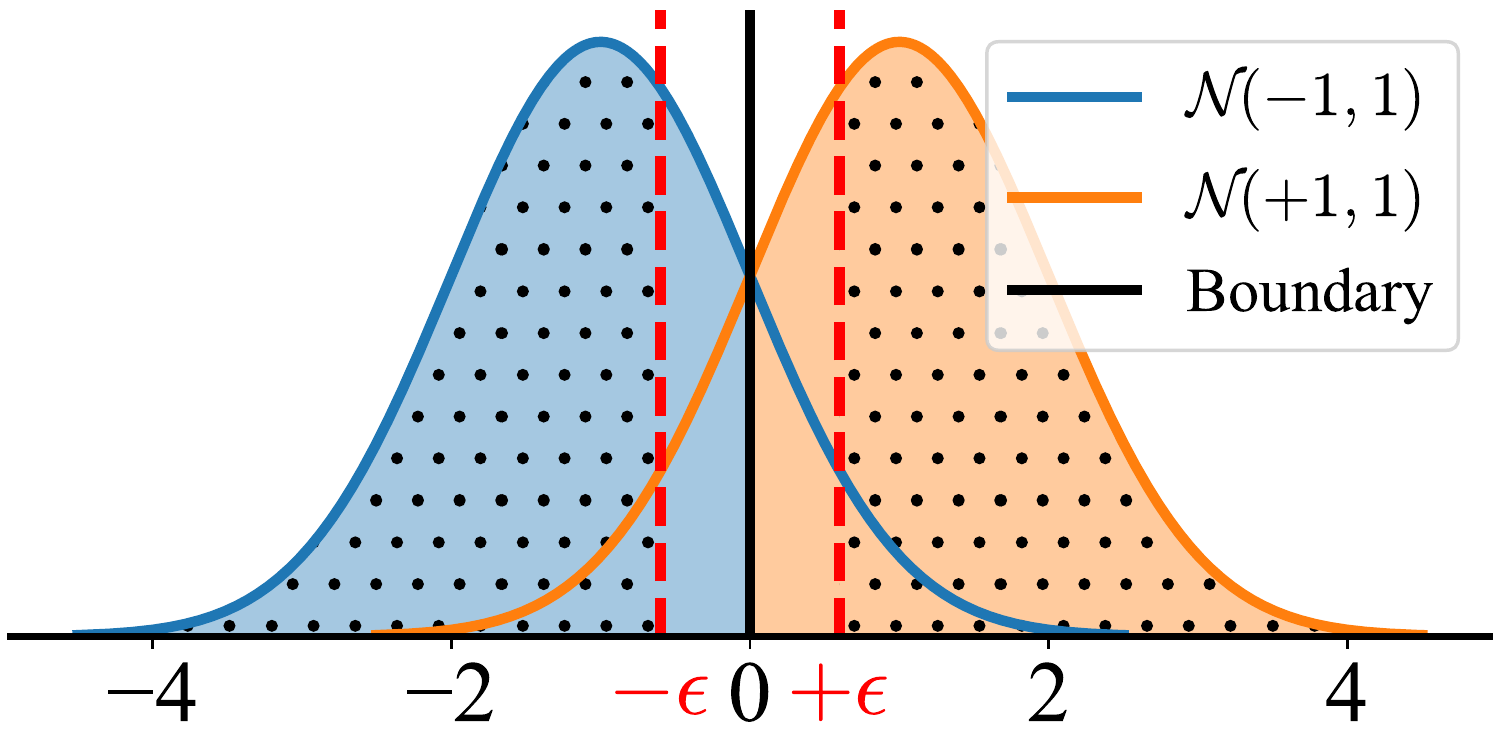}
	\caption{Undefended}
	\label{fig:tradeoff:1}
\end{subfigure}
\begin{subfigure}{0.325\linewidth}
	\includegraphics[width=\linewidth]{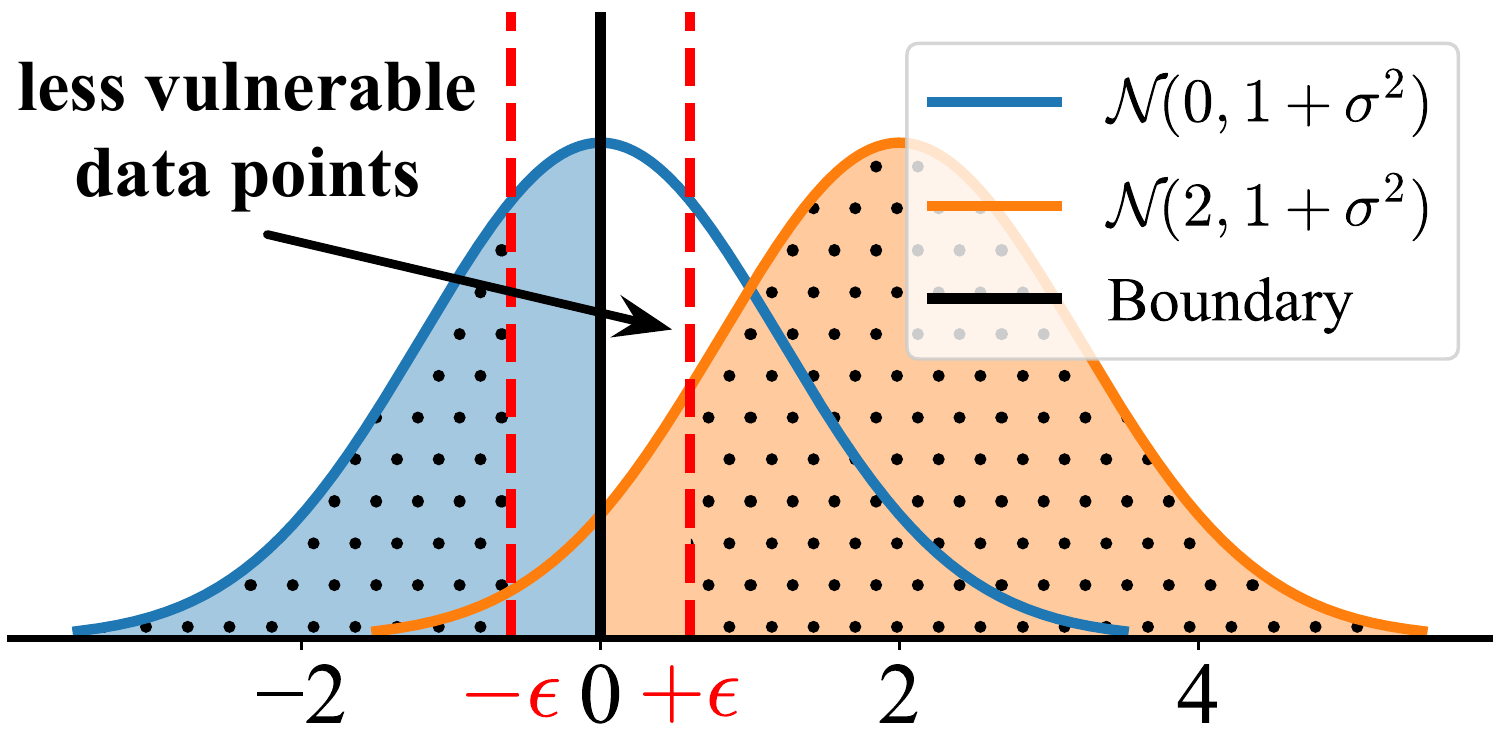}
	\caption{Defended (Lack of Invariance)}
	\label{fig:tradeoff:2}
\end{subfigure}
\begin{subfigure}{0.325\linewidth}
	\includegraphics[width=\linewidth]{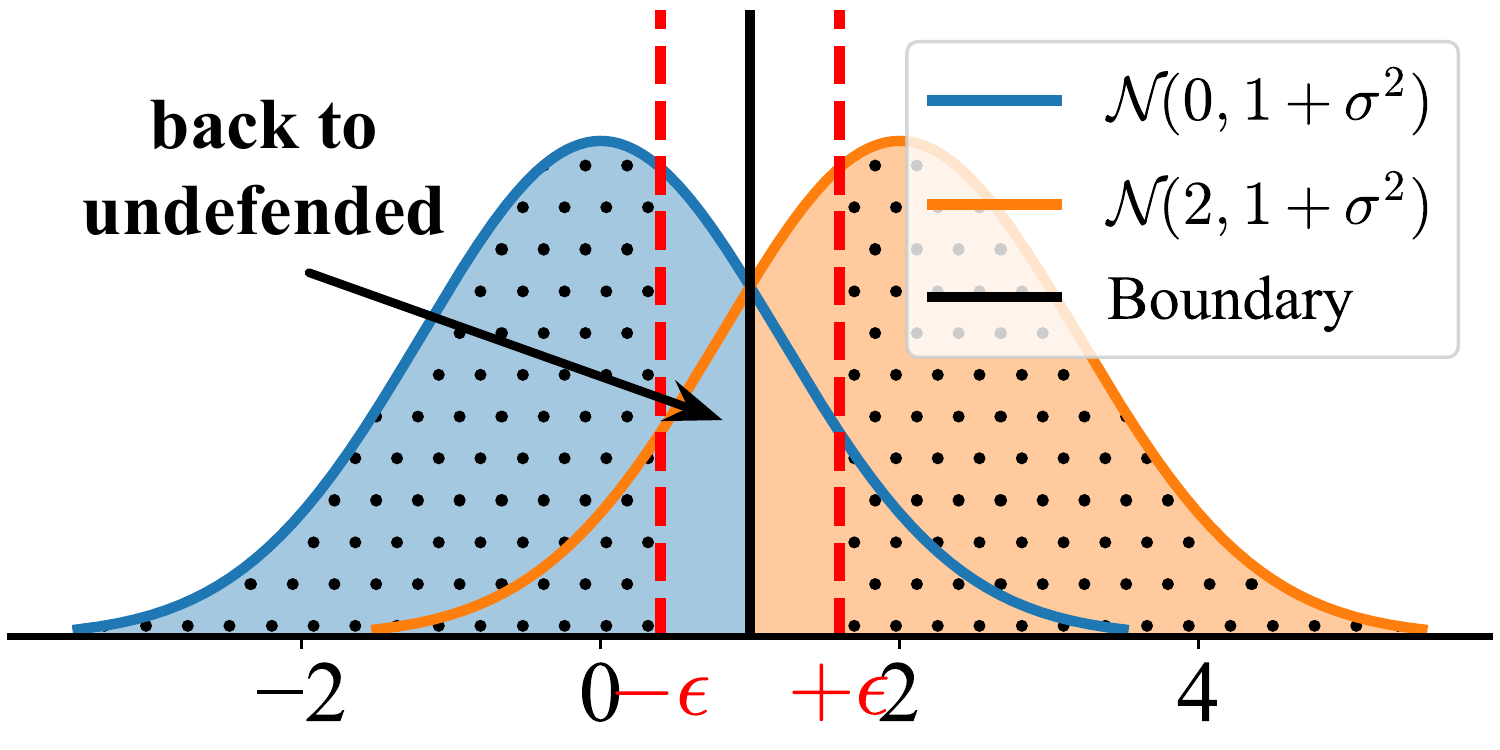}
	\caption{Defended (Trained Invariance)}
	\label{fig:tradeoff:3}
\end{subfigure}
\caption{Illustration of the binary classification task we consider. The curves are the probability density function of two classes of data. Shadowed area denotes correct classification. Dotted area denotes robustly correct classification under the \LL{\infty}-bounded adversary with perturbation budget $\epsilon$.}
\end{figure}

%This section provides details of the theoretical setting discussed in \Cref{sec:tradeoff}. 

We consider a class-balanced dataset $\mathcal{D}$ consisting of input-label pairs $(x, y)$ with $y\in\s{-1,+1}$ and $x|y\sim\mathcal{N}(y, 1)$, where $\mathcal{N}(\mu, \sigma^2)$ is a normal distribution with mean $\mu$ and variance $\sigma^2$. An \LL{\infty}-bounded adversary perturbs the input with a small $\delta$ to fool the classifier for $\norm{\delta}_{\infty}\leq\epsilon$. We quantify the classifier's robustness by its robust accuracy, i.e., the ratio of correctly classified samples that remain correct after being perturbed by the adversary. 
We also consider a stochastic pre-processing defense $\InputTransformation_{\theta}(x) \coloneqq x + \theta$, where $\theta\sim\mathcal{N}(1, \sigma^2)$ is the random variable parameterizing the defense. %Here, we have $\InputTransformation_{\theta}(x)\sim\mathcal{N}(y+1, 1+\sigma^2)$.

We formalize our assumptions as follows. Assumptions \ref{assumption:defense} to \ref{assumption:vote} characterize the standard behavior of classifiers that employ the pre-processing defense, and Assumption \ref{assumption:simplicity} specifies a set of hyper-parameters to simplify the analysis without loss of generality.

\begin{assumption}[pre-processing defense]\label{assumption:defense}
	The classifier only employs a pre-processing defense of the form $t_\theta(x)\coloneqq x+\theta$. As such, the defended classifier is defined as $F_\theta(x)\coloneqq\sgn(x+\theta-k)$, where $k$ is the decision boundary it wants to optimize.
\end{assumption}

\begin{assumption}[trained invariance]\label{assumption:invariance}
	The defended classifier controls its invariance to the defense's transformation through trained invariance, i.e., shifting the decision boundary $k$.
\end{assumption}

\begin{assumption}[majority vote]\label{assumption:vote}
	The defended classifier employs majority vote (for higher invariance) only after it improves the trained invariance. We only consider a sufficiently large number of votes.
\end{assumption}

\begin{assumption}[hyper-parameters]\label{assumption:simplicity}
	For simplicity, we assume that the defender applies $\theta\sim\mathcal{N}(1,1)$ and the adversary is reasonably strong with a perturbation budget $\epsilon=1$. Note that $\epsilon=1$ allows the adversary to shift half of the data across the decision boundary in the undefended scenario.
\end{assumption}

\textbf{Disambiguation of Notations.} We use $x + \delta$ to denote the perturbed input passed to a classifier, such as $F(x+\delta)$, but its actual value can be any value chosen from $[x-\epsilon, x+\epsilon]$. We use $\varphi$ and $\Phi$ to denote the PDF and CDF of the standard normal distribution $\mathcal{N}(0, 1)$, respectively. We use $\varphi^\prime$ and $\Phi^\prime$ to denote the PDF and CDF for non-standard normal distributions, whose parameters will be specified in the context.

\subsection{Detailed Analysis of the Binary Classification Task}
\label{app:tradeoff:analysis}

We outline the detailed computations of \Cref{sec:tradeoff} below.

\subsubsection{Undefended Classification}
\label{app:tradeoff:binary:undefended}
The Bayes optimal linear classifier $\DecisionClassifier(x)=\sgn(x)$ without any defense is illustrated in \Cref{fig:tradeoff:1}. This classifier  has benign accuracy (i.e., shadowed area):
\begin{equation}\label{eq:undefended:acc}
\begin{aligned}[b]
  \Prob{ \DecisionClassifier(x) = y }
  & = \frac{1}{2}\p[\Big]{ \Prob{ \DecisionClassifier(x) = y \given y = -1} + \Prob{ \DecisionClassifier(x) = y \given y = +1} } \\
  & = \frac{1}{2}\p[\Big]{ \Prob{ x < 0 \given y = -1} + \Prob{ x > 0 \given y = +1} } \\
  & = \frac{1}{2}\p[\Big]{ \Prob{ \mathcal{N}(-1, 1) < 0 } + \Prob{ \mathcal{N}(+1, 1) > 0 } } \\
  & = \frac{1}{2}\p[\Big]{ \Phi(1) + 1 - \Phi(-1) } \\
  & = \Phi(1).
\end{aligned}
\end{equation}

We then compute its probability of making robustly correct predictions (i.e., dotted area)
\begin{equation}
\begin{aligned}[b]
  & \Prob{ \DecisionClassifier(x + \delta) = y \land \DecisionClassifier(x) = y } \\
  & = \frac{1}{2}\p[\Big]{
          \Prob{ \DecisionClassifier(x + \delta) = y \land \DecisionClassifier(x) = y \given y = -1 }
        + \Prob{ \DecisionClassifier(x + \delta) = y \land \DecisionClassifier(x) = y \given y = +1 }
      } \\
  & = \frac{1}{2}\p[\Big]{
          \Prob{ x + \epsilon < 0 \land x < 0 \given y = -1 }
        + \Prob{ x - \epsilon > 0 \land x > 0 \given y = +1 }
      } \\
  & \quad \text{(where we use $x-\epsilon$ when $y=+1$ because the correctly classified sample must lie on the right)} \\
  & = \frac{1}{2}\p[\Big]{
          \Prob{ x < -\epsilon \given y = -1 }
        + \Prob{ x > \epsilon \given y = +1 }
      } \\
  & = \frac{1}{2}\p[\Big]{
          \Prob{ \mathcal{N}(-1, 1) < -\epsilon }
        + \Prob{ \mathcal{N}(+1, 1) > \epsilon }
      } \\
  & = \frac{1}{2}\p[\Big]{
          \Phi(1-\epsilon)
        + 1 - \Phi(\epsilon - 1)
      } \\
  & = \Phi(1-\epsilon),
\end{aligned}
\end{equation}
which shows that this classifier has robust accuracy (i.e., dotted area over shadowed area)
\begin{equation}\label{eq:undefended:rob}
\begin{aligned}[b]
  \Prob{ \DecisionClassifier(x + \delta) = y \given \DecisionClassifier(x) = y}
   = \frac{\Prob{ \DecisionClassifier(x + \delta) = y \land \DecisionClassifier(x) = y}}{\DecisionClassifier(x) = y} 
   = \frac{\Phi(1-\epsilon)}{\Phi(1)},
\end{aligned}
\end{equation}
which verifies the computation in \Cref{eq:linear-natural}.

\subsubsection{Defended Classification}
\label{app:tradeoff:binary:defended}

The defended classifier $\DecisionClassifier_{\theta}(x)=\sgn(x+\theta)$ is illustrated in \Cref{fig:tradeoff:2}, with benign accuracy
\begin{equation}\label{eq:defended:acc}
\begin{aligned}[b]
  \Prob{ F_\theta(x) = y }
  & = \frac{1}{2}\p[\Big]{ \Prob{ F_\theta(x) = y \given y = -1 } + \Prob{ F_\theta(x) = y \given y = +1} } \\
  & = \frac{1}{2}\p[\Big]{ \Prob{ x + \theta < 0 \given y = -1 } + \Prob{ x + \theta > 0 \given y = +1} } \\
  & = \frac{1}{2}\p[\Big]{ \Prob{ \mathcal{N}(0, 1+\sigma^2) < 0 } + \Prob{ \mathcal{N}(2, 1+\sigma^2) > 0 } } \\
  & = \frac{1}{2}\p[\Big]{ \Phi^\prime(0) + \Phi^\prime(2) } \\
\end{aligned}
\end{equation}
where $\Phi^\prime(x)\coloneqq\Phi(x/\sqrt{1+\sigma^2})$ is the cumulative distribution function of $\mathcal{N}(0, 1+\sigma^2)$.

We then compute its probability of making robustly correct predictions (i.e., dotted area)
\begin{equation}
\begin{aligned}[b]
  & \Prob{ F_\theta(x + \delta) = y \land F_\theta(x) = y } \\
  & = \frac{1}{2}\p[\Big]{
          \Prob{ F_\theta(x + \delta) = y \land F_\theta(x) = y \given y = -1 }
        + \Prob{ F_\theta(x + \delta) = y \land F_\theta(x) = y \given y = +1 }
      } \\
  & = \frac{1}{2}\p[\Big]{
          \Prob{ x + \theta + \epsilon < 0 \land x + \theta < 0 \given y = -1 }
        + \Prob{ x + \theta - \epsilon > 0 \land x + \theta > 0 \given y = +1 }
      } \\
  & \quad \text{(where we use $x+\theta-\epsilon$ when $y=+1$ because the correctly classified sample must lie on the right)} \\
  & = \frac{1}{2}\p[\Big]{
          \Prob{ x + \theta < -\epsilon \given y = -1 }
        + \Prob{ x + \theta > \epsilon \given y = +1 }
      } \\
  & = \frac{1}{2}\p[\Big]{
          \Prob{ \mathcal{N}(0, 1+\sigma^2) < -\epsilon }
        + \Prob{ \mathcal{N}(2, 1+\sigma^2) > \epsilon }
      } \\
  & = \frac{1}{2}\p[\Big]{
          \Phi^\prime(-\epsilon)
        + \Phi^\prime(2-\epsilon)
      },
\end{aligned}
\end{equation}
where $\Phi^\prime(x)\coloneqq\Phi(x/\sqrt{1+\sigma^2})$ is the cumulative distribution function of $\mathcal{N}(0, 1+\sigma^2)$.

It shows that this classifier has robust accuracy (i.e., dotted area over shadowed area)
\begin{equation}\label{eq:defended:rob}
\begin{aligned}[b]
  \Prob{ \DecisionClassifier_{\theta}(x + \delta) = y \given \DecisionClassifier_{\theta}(x) = y}
   = \frac{\Prob{ \DecisionClassifier_{\theta}(x + \delta) = y \land \DecisionClassifier_{\theta}(x) = y}}{\DecisionClassifier_{\theta}(x) = y} 
   = \frac{\Phi^\prime(-\epsilon) + \Phi^\prime(2-\epsilon)}{\Phi^\prime(0) + \Phi^\prime(2)},
\end{aligned}
\end{equation}
where $\Phi^\prime(x)\coloneqq\Phi(x/\sqrt{1+\sigma^2})$ is the cumulative distribution function of $\mathcal{N}(0, 1+\sigma^2)$. This verifies the computation in \Cref{eq:linear-defended}.

Here, we can make a quick observation under \Cref{assumption:simplicity}, where we assume $\sigma=1$ and $\epsilon=1$ for simplicity. It shows that the stochastic pre-processing defense in our setting explicitly reduces invariance and utility to gain robustness. The general case is proven in \Cref{theorem}. 
\begin{observation}
	The defended classifier $\DecisionClassifier_{\theta}(x)=\sgn(x+\theta)$ has higher robust accuracy (67.7\% vs.\ 59.4\%) yet lower benign accuracy (71.1\% vs.\ 84.1\%) than the undefended classifier $\DecisionClassifier(x)=\sgn(x)$.
\end{observation}

\subsubsection{Defended Classification (Trained Invariance)}
\label{app:tradeoff:binary:trained}

One critical step of stochastic pre-processing defenses is to preserve the defended model's utility by minimizing the risk over processed data $\InputTransformation_\theta(x)$, which leads to a new defended classifier $\DecisionClassifier^+_{\theta}(x)=\sgn(x+\theta-1)$ that is optimal on transformed data, as illustrated in \Cref{fig:tradeoff:3}. It has benign accuracy
\begin{equation}\label{eq:trained:acc}
\begin{aligned}[b]
  \Prob{ F_{\theta}^+(x) = y }
  & = \frac{1}{2}\p[\Big]{ \Prob{ F_{\theta}^+(x) = y \given y = -1 } + \Prob{ F_{\theta}^+(x) = y \given y = +1 } } \\
  & = \frac{1}{2}\p[\Big]{ \Prob{ x + \theta - 1 < 0 \given y = -1 } + \Prob{ x + \theta - 1 > 0 \given y = +1 } } \\
  & = \frac{1}{2}\p[\Big]{ \Prob{ \mathcal{N}(0, 1+\sigma^2) < 1 } + \Prob{ \mathcal{N}(2, 1+\sigma^2) > 1 } } \\
  & = \frac{1}{2}\p[\Big]{ \Phi^\prime(1) + 1 - \Phi^\prime(-1) } \\
  & = \Phi^\prime(1),
\end{aligned}
\end{equation}
where $\Phi^\prime(x)\coloneqq\Phi(x/\sqrt{1+\sigma^2})$ is the cumulative distribution function of $\mathcal{N}(0, 1+\sigma^2)$.

We then compute its probability of making robustly correct predictions (i.e., dotted area)
%\begin{equation}
%\begin{aligned}[b]
%  \Prob{ \DecisionClassifier_{\theta}^+(x + \epsilon) = y \land \DecisionClassifier_{\theta}^+(x) = y}
%  & = \Prob{ \DecisionClassifier_{\theta}^+(x + \epsilon) = y \land \DecisionClassifier_{\theta}^+(x) = y \given y = -1} \cdot \Prob{ y = -1 } \\
%      & \qquad + \Prob{ \DecisionClassifier_{\theta}^+(x + \epsilon) = y \land \DecisionClassifier_{\theta}^+(x) = y \given y = +1} \cdot \Prob{ y = +1 } \\
%  & = \Prob{ x + \theta - 1 + \epsilon < 0 \land x + \theta - 1 < 0 \given y = -1} \cdot \Prob{ y = -1 } \\
%      & \qquad + \Prob{ x + \theta - 1 - \epsilon > 0 \land x + \theta - 1 > 0 \given y = +1} \cdot \Prob{ y = +1 } \\
%  & = \Prob{ x + \theta < 1-\epsilon \given y = -1} \cdot \Prob{ y = -1 } \\
%      & \qquad + \Prob{ x + \theta  > 1 + \epsilon \given y = +1} \cdot \Prob{ y = +1 } \\
%  & = \Prob{ \mathcal{N}(0, 1+\sigma^2) < 1-\epsilon } \cdot \Prob{ y = -1 } \\
%      & \qquad + \Prob{ \mathcal{N}(2, 1+\sigma^2) > 1+\epsilon } \cdot \Prob{ y = +1 } \\ 
%  & = \Phi^\prime(1-\epsilon) \times \frac{1}{2} + \Phi^\prime(1-\epsilon) \times \frac{1}{2} \\
%  & = \Phi^\prime(1-\epsilon),
%\end{aligned}
%\end{equation}
\begin{equation}
\begin{aligned}[b]
  & \Prob{ F_\theta^+(x + \delta) = y \land F_\theta^+(x) = y } \\
  & = \frac{1}{2}\p[\Big]{
          \Prob{ F_\theta^+(x + \delta) = y \land F_\theta^+(x) = y \given y = -1 }
        + \Prob{ F_\theta^+(x + \delta) = y \land F_\theta^+(x) = y \given y = +1 }
      } \\
  & = \frac{1}{2}\p[\Big]{
          \Prob{ x + \theta - 1 + \epsilon < 0 \land x + \theta - 1 < 0 \given y = -1 }
        + \Prob{ x + \theta - 1 - \epsilon > 0 \land x + \theta - 1 > 0 \given y = +1 }
      } \\
  & \quad \text{(where we use $x+\theta-1-\epsilon$ when $y=+1$ because the correctly classified sample must lie on the right)} \\
  & = \frac{1}{2}\p[\Big]{
          \Prob{ x + \theta < 1 - \epsilon \given y = -1 }
        + \Prob{ x + \theta > 1 + \epsilon \given y = +1 }
      } \\
  & = \frac{1}{2}\p[\Big]{
          \Prob{ \mathcal{N}(0, 1+\sigma^2) < 1 - \epsilon }
        + \Prob{ \mathcal{N}(2, 1+\sigma^2) > 1 + \epsilon }
      } \\
  & = \frac{1}{2}\p[\Big]{
          \Phi^\prime(1 - \epsilon)
        + 1 - \Phi^\prime(\epsilon - 1)
      } \\
  & = \Phi^\prime(1 - \epsilon),
\end{aligned}
\end{equation}
where $\Phi^\prime(x)\coloneqq\Phi(x/\sqrt{1+\sigma^2})$ is the cumulative distribution function of $\mathcal{N}(0, 1+\sigma^2)$.

It shows that this classifier has robust accuracy (i.e., dotted area over shadowed area)
\begin{equation}\label{eq:trained:rob}
\begin{aligned}[b]
  \Prob{ \DecisionClassifier_{\theta}^+(x + \delta) = y \given \DecisionClassifier_{\theta}^+(x) = y}
   = \frac{\Prob{ \DecisionClassifier_{\theta}^+(x + \delta) = y \land \DecisionClassifier_{\theta}^+(x) = y}}{\DecisionClassifier_{\theta}^+(x) = y} 
   = \frac{\Phi^\prime(1-\epsilon)}{\Phi^\prime(1)},
\end{aligned}
\end{equation}
where $\Phi^\prime(x)\coloneqq\Phi(x/\sqrt{1+\sigma^2})$ is the cumulative distribution function of $\mathcal{N}(0, 1+\sigma^2)$. This verifies the computation in \Cref{eq:linear-trained}.

Here, we can make the following two observations under \Cref{assumption:simplicity}, where we assume $\sigma=1$ and $\epsilon=1$ for simplicity. They show that the defense has to increase the invariance, which was previously reduced to gain robustness, to recover utility. The general case is proven in \Cref{theorem}.
\begin{observation}
The defended classifier with trained invariance	$\DecisionClassifier^+_{\theta}(x)=\sgn(x+\theta-1)$ is less robust (65.8\% vs.\ 70.4\%) than the defended classifier $\DecisionClassifier_{\theta}(x)=\sgn(x+\theta)$ without trained invariance.
\end{observation}

\begin{observation}
The defended classifier with trained invariance	$\DecisionClassifier^+_{\theta}(x)=\sgn(x+\theta-1)$ is more robust (65.8\% vs.\ 59.4\%) than the original undefended classifier $\DecisionClassifier(x)=\sgn(x)$ at the cost of utility (76.0\% vs.\ 84.1\%).
\end{observation}

\subsubsection{Defended Classification (Perfect Invariance)}
\label{app:tradeoff:binary:invariant}

Furthermore, these defenses usually leverage majority vote to obtain stable predictions, which finally produces a perfectly invariant defended classifier:
\begin{equation}
\begin{aligned}[b]
  \DecisionClassifier_{\theta}^*(x)
  & = \argmax_{y\in\s{-1, +1}} \sum_{i=1}^n \indicator\s[\big]{\DecisionClassifier^+_{\theta_i}(x) = y} \\
  & = \sgn\p*{\frac{1}{n}\sum_{i=1}^n \DecisionClassifier^+_{\theta_i}(x)} \\
  & = \sgn\p*{\frac{1}{n}\sum_{i=1}^n \sgn\p*{x+\theta_i-1}} \\
  & \to \sgn\p*{\mathop{\EE}_{\theta\sim\mathcal{N}(1, \sigma^2)}\ProbBr[\Big]{\sgn\p*{x+\theta-1} \given x}} \\
  & = \sgn\p*{\mathop{\EE}_{z|x\sim\mathcal{N}(x, \sigma^2)}\ProbBr[\Big]{\sgn(z) \given x}} \\
  & = \sgn\p*{\Prob[\Big]{\mathcal{N}(x, \sigma^2) > 0 \given x} - \Prob[\Big]{\mathcal{N}(x, \sigma^2) < 0 \given x}} \\
  & = \sgn(x),
\end{aligned}
\end{equation}
where the last equality holds because $\mathcal{N}(x, \sigma^2)$ has more probability on the positive side if and only if $x>0$ and has more probability on the negative side if and only if $x<0$. As we can observe, the defended classifier with trained invariance and majority vote reduces to the original undefended classifier $\DecisionClassifier(x)=\sgn(x)$, which verifies \Cref{eq:linear-reduction}.

\subsection{Theorem: Trade-off between Robustness and Invariance}
\label{app:tradeoff:theorem}

In this section, we extend the above coupling between robustness and invariance to a general trade-off, where we can control the invariance through shifting decision boundary and employing majority vote.

Recall that $x|y\sim\mathcal{N}(y,1)$ and $\theta\sim\mathcal{N}(1,\sigma^2)$, we denote their density functions by
\begin{equation}
	\varphi_x = \begin{cases}
		\varphi(x+1), & y=-1 \\
		\varphi(x-1), & y=+1
	\end{cases},
	\Phi_x = \begin{cases}
		\Phi(x+1), & y=-1 \\
		\Phi(x-1), & y=+1
	\end{cases},
	\varphi_\theta = \varphi\p*{\frac{\theta-1}{\sigma}},
	\Phi_\theta = \Phi\p*{\frac{\theta-1}{\sigma}},
\end{equation}
where $\varphi$ and $\Phi$ are the probability and cumulative density functions of $\mathcal{N}(0,1)$, respectively.

\textbf{Rate of Invariance.} To facilitate our analysis, given the theoretical setting and assumptions specified in \Cref{app:tradeoff:analysis}, we define the \emph{rate of invariance} for a defended classifier $F_{\theta}(x)$ as
\begin{equation}
R(k)\coloneqq\Prob{\DecisionClassifier_{\theta}(x) = \DecisionClassifier(x)},
\end{equation}
where $F_{\theta}(x)=\sgn(x+\theta-k)$, and $F(x)=\sgn(x)$ is the undefended classifier.

%, where $k\in[0, 1]$ is the decision boundary. From the above discussion, the defended classifier approximate trained invariance by shifting the decision boundary from $k=0$ to $k=1$.

We formalize the trade-off between robustness and invariance in the following theorem proven in \Cref{app:proofs:tradeoff}. It shows that stochastic pre-processing defenses provide robustness by intentionally reducing the model's invariance to added randomized transformations.

%\begin{restatable}[Trade-off between Robustness and Invariance]{theorem}{theoremTradeoff}
%\label{theorem}
%	Given the above theoretical setting and assumptions, when the defended classifier $F_\theta(x)$ achieves higher invariance $R(k)$ under the defense's randomization space to preserve utility, the adversarial robustness provided by the defense strictly decreases.
%\end{restatable}
\theoremTradeoff*

We prove this theorem by characterizing the (strictly opposite) monotonic behavior of invariance and robustness as the defended classifier shifts its decision boundary towards the optimal decision boundary $k=1$ on transformed data (see \Cref{app:tradeoff:binary:trained}) and applies majority vote at the end. We formalize such characterizations in the following lemmas and corollaries.

First, we show in \Cref{lemma:invariance} that the defended classifier's rate of invariance strictly increases as the decision boundary shifts towards the optima; applying majority vote further yields perfect invariance, as we show in \Cref{{corollary:invariance}}. We prove them in \Cref{app:proofs:invariance}. 

\begin{restatable}[Strictly Increasing Invariance]{lemma}{lemmaInvariance}
\label{lemma:invariance}
	The defended classifier's invariance $R(k)$ strictly increases as the decision boundary approaches $k=1$ without applying majority vote.
\end{restatable}

\begin{restatable}[Perfect Invariance by Majority Vote]{corollary}{corollaryInvariance}
\label{corollary:invariance}
	When the defended classifier maximizes trained invariance at $k=1$, employing majority vote further improves the rate of invariance $R(k)$ to one.
\end{restatable}

Second, we show in \Cref{lemma:robustness} that the defended classifier's robust accuracy strictly decreases as the decision boundary shifts towards the optima. When the trained invariance is approximated, we show in \Cref{corollary:robustness} that applying majority vote strictly decreases the robust accuracy further. We prove them in \Cref{app:proofs:robustness}. 

\begin{restatable}[Strictly Decreasing Robustness]{lemma}{lemmaRobustness}
\label{lemma:robustness}
	The defended classifier's robust accuracy strictly decreases as the decision boundary approaches $k=1$ without applying majority vote.
\end{restatable}

\begin{restatable}[Strictly Decreasing Robustness by Majority Vote]{corollary}{corollaryRobustness}
\label{corollary:robustness}
	 When the defended classifier approximates the trained invariance by shifting its decision boundary to $k\in[0,2]$, applying majority vote strictly decreases its robust accuracy.
\end{restatable}

Finally, we show in \Cref{lemma:accuracy} that the defended classifier indeed preserves its utility by shifting the decision boundary towards the optima, and applying majority vote recovers the full utility as we show in \Cref{corollary:accuracy}. We prove them in \Cref{app:proofs:accuracy}.

\begin{restatable}[Strictly Increasing Accuracy]{lemma}{lemmaAccuracy}
\label{lemma:accuracy}
	The defended classifier's benign accuracy strictly increases as the decision boundary approaches $k=1$ without applying majority vote.
\end{restatable}

\begin{restatable}[Strictly Increasing Accuracy by Majority Vote]{corollary}{corollaryAccuracy}
\label{corollary:accuracy}
	When the defended classifier approximates the trained invariance by shifting its decision boundary to $k\in[0,2]$, applying majority vote strictly increases its accuracy.
\end{restatable}

\subsection{Proofs}
\label{app:proofs}

We provide complete proofs for the theorems, lemmas, and corollaries that we present above.

\subsubsection{Strictly Increasing Invariance}
\label{app:proofs:invariance}

\lemmaInvariance*
\begin{proof}

We directly compute the rate of invariance $R(k)$ as
\begin{equation}
\begin{aligned}[b]
\label{eq:temp1}
	R(k)
	& = \Prob{\DecisionClassifier_{\theta}(x) = \DecisionClassifier(x)} \\
	& = \Prob{\sgn(x+\theta-k) = \sgn(x)} \\
	& = \Prob{\sgn(x+\theta-k) = \sgn(x) \land x < 0} + \Prob{\sgn(x+\theta-k) = \sgn(x) \land x > 0} \\
	& = \Prob{\theta < k - x \land x < 0} + \Prob{\theta > k - x \land x > 0} \\
	& = \int_{-\infty}^0\int_{-\infty}^{k-x}\varphi_x(x)\cdot\varphi_\theta(\theta) \dd \theta \dd x
	  + \int_0^{\infty}\int_{k-x}^{\infty}\varphi_x(x)\cdot\varphi_\theta(\theta) \dd \theta \dd x \\
	& = \int_{-\infty}^0 \varphi_x(x)\cdot\Phi_\theta(k-x) \dd x
	  - \int_0^{\infty} \varphi_x(x)\cdot\Phi_\theta(k-x) \dd x
	  + \int_0^{\infty} \varphi_x(x) \dd x,
\end{aligned}
\end{equation}
whose gradient with respect to $k$ is
\begin{equation}
\label{eq:rob_grad}
\begin{aligned}[b]
\frac{\partial}{\partial k} R(k) 
& = \int_{-\infty}^0 \varphi_x(x)\cdot\varphi_\theta(k-x) \dd x - \int_0^{\infty} \varphi_x(x)\cdot\varphi_\theta(k-x) \dd x \\
& = \frac{1}{2} \p*{\int_{-\infty}^0 \varphi(x+1)\cdot\varphi_\theta(k-x) \dd x - \int_0^{\infty} \varphi(x+1)\cdot\varphi_\theta(k-x) \dd x} \\
& \quad + \frac{1}{2} \p*{\int_{-\infty}^0 \varphi(x-1)\cdot\varphi_\theta(k-x) \dd x - \int_0^{\infty} \varphi(x-1)\cdot\varphi_\theta(k-x) \dd x}.
\end{aligned}
\end{equation}

From calculus and the error function $\erf$ we have
\begin{equation}
\begin{aligned}[b]
	p_1(x) & \coloneqq \int \varphi(x+1) \cdot \varphi_\theta(k-x) \dd x = \frac{1}{4\sqrt\pi} \exp\p*{-\frac{k^2}{4}} \cdot \erf\p*{1-\frac{k}{2}+x} \\
	p_2(x) & \coloneqq \int \varphi(x-1) \cdot \varphi_\theta(k-x) \dd x = \frac{-1}{4\sqrt\pi} \exp\p*{-\frac{(k-2)^2}{4}} \cdot \erf\p*{\frac{k}{2}-x}
%	    -
%	    \int_0^{\infty} \varphi(x+1)\cdot\varphi_\theta(k-x) \dd x \\
%	   &= \frac{1}{\sqrt{2\pi}}\exp\p*{-\frac{k^2}{4}} \cdot \erf\p*{1-\frac{k}{2}} \\
%	    -
%	    \int_0^{\infty} \varphi(x-1)\cdot\varphi_\theta(k-x) \dd x \\
%	    &= -\frac{1}{\sqrt{2\pi}}\exp\p*{-\frac{(k-2)^2}{4}} \cdot \erf\p*{\frac{k}{2}},
\end{aligned},
\end{equation}
where we have assumed $\theta\sim\mathcal{N}(1,1)$ to simplify the analysis by \Cref{assumption:simplicity}.

It shows that the gradient in \Cref{eq:rob_grad} is
\begin{equation}
\begin{aligned}
\frac{\partial}{\partial k} R(k)
& = \frac{1}{2}\p[\Big]{p_1(0) - p_1(-\infty) - p_1(\infty) + p_1(0)} + \frac{1}{2}\p[\Big]{p_2(0) - p_2(-\infty) - p_2(\infty) + p_2(0)} \\
& = p_1(0) + p_2(0) \\
%& = \frac{1}{\sqrt{2\pi}}\exp\p*{-\frac{k^2}{4}} \cdot \erf\p*{1-\frac{k}{2}} -\frac{1}{\sqrt{2\pi}}\exp\p*{-\frac{(k-2)^2}{4}} \cdot \erf\p*{\frac{k}{2}} \\
& \propto \exp\p*{-\frac{k^2}{4}} \cdot \erf\p*{1-\frac{k}{2}} - \exp\p*{-\frac{(k-2)^2}{4}} \cdot \erf\p*{\frac{k}{2}}.
\end{aligned}
\end{equation}

Notice that $G(k)\coloneqq\frac{\partial}{\partial k} R(k)$ is a symmetric function with respect to the point $(1, 0)$:
\begin{equation}
	G(1+z) + G(1-z) = 0, \quad \forall z\in\RR,
\end{equation}
which shows that $G(k)$ attains zero at $k=1$.

Since both $\exp$ and $\erf$ are strictly increasing functions, for $k<1$, we have
\begin{equation}
\begin{aligned}
	-\frac{k^2}{4} & > -\frac{(k-2)^2}{4} && \implies \exp\p*{-\frac{k^2}{4}}  > \exp\p*{-\frac{(k-2)^2}{4}}, \\
	1-\frac{k}{2} & > \frac{k}{2} && \implies \erf\p*{1-\frac{k}{2}} > \erf\p*{\frac{k}{2}},
\end{aligned}
\end{equation}
which shows that $G(k) > 0$ when $k<1$, and $G(k) < 0$ when $k>1$ by symmetry.

Therefore, the rate of invariance $R(k)$ strictly increases for $k<1$ and strictly decreases for $k>1$.
\end{proof}

\corollaryInvariance*

\begin{proof}
	We showed in \Cref{app:tradeoff:binary:invariant} that the defended classifier $F_\theta(x)=\sgn(x+\theta-1)$ converges to the optimal classifier $F_\theta(x)=\sgn(x)$ if given a sufficiently large number of votes.

In such a case, it is straightforward to show that the rate of invariance converges to one:
	\begin{equation}
		R(k=1) = \Prob{F_{\theta}(x) = F(x)} \rightarrow \Prob{\sgn(x) = \sgn(x)} = 1.
	\end{equation}
\end{proof}

\subsubsection{Strictly Decreasing Robustness}
\label{app:proofs:robustness}

\lemmaRobustness*

\begin{proof}
We directly compute the robust accuracy of the defended classifier $F_\theta(x)=\sgn(x+\theta-k)$ and characterize its monotonic behavior. Recall that $x+\theta\sim\mathcal{N}(y+1, 2)$.

We first compute the defended classifier's benign accuracy:
\begin{equation}
\label{eq:benign_acc}
\begin{aligned}[b]
	& \Prob{F_\theta(x) = y} \\
	& = \Prob{\sgn(x+\theta-k) = y} \\
	& = \Prob{x+\theta < k \given y = -1}\cdot\Prob{y=-1} + \Prob{x+\theta > k \given y = +1}\cdot\Prob{y=+1} \\
	& = \Prob{\mathcal{N}(0, 2) < k}\cdot\Prob{y=-1} + \Prob{\mathcal{N}(2,2) > k}\cdot\Prob{y=+1} \\
	& = \frac{1}{2}\p[\Big]{\Phi^\prime(k) + \Phi^\prime(2-k)},
\end{aligned}
\end{equation}
where $\Phi^\prime$ denotes the cumulative density function of $\mathcal{N}(0,2)$.

We then compute the probability of robustly correct predictions, where we use $x+\delta$ to denote the adversarial example that actually can take any value from $[x-\epsilon,x+\epsilon]$ to change the prediction:
\begin{equation}
\begin{aligned}[b]
	& \Prob{F_\theta(x+\delta) = y \land F_\theta(x) = y} \\
	& = \frac{1}{2}\p[\Big]{\Prob{F_\theta(x+\delta) = y \land F_\theta(x) = y \given y = -1} + \Prob{F_\theta(x+\delta) = y \land F_\theta(x) = y \given y = +1}} \\
	& = \frac{1}{2}\p[\Big]{\Prob{x+\theta-k+\epsilon<0 \land x+\theta-k<0 \given y = -1} + \Prob{x+\theta-k-\epsilon>0 \land x+\theta-k>0 \given y = +1}} \\
	& \quad \text{(where we use $-\epsilon$ when $y=+1$ because the correctly classified sample must lie on the right)} \\
	& = \frac{1}{2}\p[\Big]{\Prob{x+\theta<k-\epsilon \land x+\theta<k \given y = -1} + \Prob{x+\theta>k+\epsilon \land x+\theta>k \given y = +1}} \\
	& = \frac{1}{2}\p[\Big]{\Prob{x+\theta<k-\epsilon \given y = -1} + \Prob{x+\theta>k+\epsilon \given y = +1}} \\
	& = \frac{1}{2}\p[\Big]{\Prob{\mathcal{N}(0,2)<k-\epsilon} + \Prob{\mathcal{N}(2,2)>k+\epsilon}} \\
	& = \frac{1}{2}\p[\Big]{\Phi^\prime(k-\epsilon) + \Phi^\prime(2-k-\epsilon)},
\end{aligned}
\end{equation}
where $\Phi^\prime$ denotes the cumulative density function of $\mathcal{N}(0,2)$.

Now we can compute the robust accuracy at decision boundary $k$ as
\begin{equation}
  \Prob{F_\theta(x+\delta) = y \given F_\theta(x) = y}
= \frac{\Prob{F_\theta(x+\delta) = y \land F_\theta(x) = y}}{\Prob{F_\theta(x) = y}}
= \frac{\Phi^\prime(k-\epsilon) + \Phi^\prime(2-k-\epsilon)}{\Phi^\prime(k) + \Phi^\prime(2-k)}.
\end{equation}

While the argument holds for any fixed $\epsilon$, we will show a simple example and assume a reasonably strong adversary with $\epsilon=1$ (\Cref{assumption:simplicity}), which initializes the robust accuracy to:
\begin{equation}
\label{eq:rob_acc_e1}
	\mathtt{Rob}(k)\coloneqq \frac{\Phi^\prime(k-1) + \Phi^\prime(1-k)}{\Phi^\prime(k) + \Phi^\prime(2-k)},
\end{equation}
where $\Phi^\prime$ denotes the cumulative density function of $\mathcal{N}(0,2)$. Its gradient with respect to $k$ is
\begin{equation}
G(k)\coloneqq\frac{\partial}{\partial k}\mathtt{Rob}(k) =\frac{2\exp\p*{-1-\frac{k^2}{4}}(e^k-e)}{\sqrt{\pi}\p*{2-\erf\p*{\frac{k-2}{2}}-\erf\p*{-\frac{k}{2}}}^2}\propto e^k-e,
\end{equation}
which shows that $G(k)<0$ when $k<1$, $G(k)>0$ when $k>1$, and $G(k)=0$ when $k=1$.

Therefore, the robust accuracy $\mathtt{Rob}(k)$ strictly decreases as $k$ approaches $k=1$ from either side.
\end{proof}

\corollaryRobustness*

\begin{proof}
	For this proof, we assume the applied defense adopts $\theta\sim\mathcal{N}(1,\sigma^2)$, which reformulates the robust accuracy in \Cref{eq:rob_acc_e1} as
\begin{equation}
\mathtt{Rob}(k)\coloneqq \frac{\Phi^\prime(k-1) + \Phi^\prime(1-k)}{\Phi^\prime(k) + \Phi^\prime(2-k)}=
2\p*{2+\erf\p*{\frac{2-k}{\sqrt{2}\sqrt{1+\sigma^2}}}+\erf\p*{\frac{k}{\sqrt{2}\sqrt{1+\sigma^2}}}}^{-1},
\end{equation}
where $\Phi^\prime$ is the cumulative density function of $\mathcal{N}(0,1+\sigma^2)$.

For $k\in[0,2]$, where the argument for $\erf$ is non-negative, decreasing $\sigma$ will also decrease the robust accuracy (the $\erf$ function is strictly increasing). Given that majority vote effectively reduces the noise's variance, having a larger number of votes will strictly decrease the robust accuracy.
\end{proof}

\subsubsection{Strictly Increasing Accuracy}
\label{app:proofs:accuracy}

\lemmaAccuracy*

\begin{proof}
In \Cref{eq:benign_acc}, we showed that the defended classifier $F_\theta(x)=\sgn(x+\theta-k)$ has benign accuracy
\begin{equation}
\label{eq:acc_tt}
\mathtt{Acc}(k)\coloneqq\Prob{F_\theta(x) = y} = \frac{1}{2}\p[\Big]{\Phi^\prime(k) + \Phi^\prime(2-k)},
\end{equation}
where $\Phi^\prime$ denotes the cumulative density function of $\mathcal{N}(0,2)$.

Its gradient with respect to $k$ is
\begin{equation}
G(k)\coloneqq\frac{\partial}{\partial k}\mathtt{Acc}(k) = \frac{\exp\p*{-1-\frac{k^2}{4}}\p*{e-e^k}}{2\sqrt\pi}\propto e-e^k,
\end{equation}
which shows that $G(k)>0$ when $k<1$, $G(k)<0$ when $k>1$, and $G(k)=0$ when $k=1$.

Therefore, the benign accuracy $\mathtt{Acc}(k)$ strictly increases as $k$ approaches $k=1$ from either side.
\end{proof}

\corollaryAccuracy*

\begin{proof}
For this proof, we assume the applied defense adopts $\theta\sim\mathcal{N}(1,\sigma^2)$, which reformulates the benign accuracy in \Cref{eq:acc_tt} as
	\begin{equation}
\mathtt{Acc}(k)\coloneqq \frac{1}{2}\p[\Big]{\Phi^\prime(k) + \Phi^\prime(2-k)} = 2+\erf\p*{\frac{2-k}{\sqrt{2}\sqrt{1+\sigma^2}}}+\erf\p*{\frac{k}{\sqrt{2}\sqrt{1+\sigma^2}}},
\end{equation}
where $\Phi^\prime$ denotes the cumulative density function of $\mathcal{N}(0,1+\sigma^2)$.

For $k\in[0,2]$, where the argument for $\erf$ is non-negative, decreasing $\sigma$ will also decrease the robust accuracy (the $\erf$ function is strictly increasing). Given that majority vote effectively reduces the noise's variance, having a larger number of votes will strictly increase the robust accuracy.

As a special case, when $k=1$ and $\sigma\to0$, we have
\begin{equation}
	\mathtt{Acc}(k) = \frac{1}{2}\p[\Big]{\Phi^\prime(k) + \Phi^\prime(2-k)} = \frac{1}{2}\p*{\Phi\p*{\frac{k}{\sqrt{1+\sigma^2}}} + \Phi\p*{\frac{2-k}{\sqrt{1+\sigma^2}}}} \to \Phi(1),
\end{equation}
which recovers the full utility of the undefended classifier in \Cref{eq:undefended:acc}.
\end{proof}

\subsubsection{Trade-off between Robustness and Invariance}
\label{app:proofs:tradeoff}

\theoremTradeoff*

\begin{proof}
The proof follows by directly combining the lemmas and corollaries proven above.

By \Cref{lemma:accuracy} and \Cref{corollary:accuracy}, when the defended classifier $F_\theta(x)=x+\theta-k$ shifts its decision boundary towards $k=1$, its benign accuracy strictly increases and is maximized at $k=1$ with the application of majority vote. This verifies that the defended classifier in our setting indeed preserves utility by shifting the decision boundary towards $k=1$.

By \Cref{lemma:invariance} and \Cref{corollary:invariance}, when the defended classifier shifts the decision boundary towards $k=1$ to preserve utility, its rate of invariance strictly increases and is maximized at $k=1$ with the application of majority vote. This verifies that the defended classifier in our setting strictly controls its invariance by shifting the decision boundary.

By \Cref{lemma:robustness} and \Cref{corollary:robustness}, when the defended classifier shifts the decision boundary towards $k=1$ to acquire more invariance, the adversarial robustness strictly decreases and is minimized at $k=1$ with the application of majority vote. 

The above arguments show that the defended classifier strictly improves its invariance by approaching $k=1$, yet the adversarial robustness strictly decreases during this process.
When perfect invariance is achieved, the utility and robustness go back to those of the undefended classifier, nullifying the initially applied stochastic pre-processing defense.
\end{proof}

\section{Experiment Setup: Main Evaluation}
\label{app:exp}

In this section, we provide more details of our main evaluation.

\subsection{Datasets}
\label{app:exp:datasets}

We conduct all experiments on the public ImageNet~\cite{imagenet} and ImageNette~\cite{imagenette} datasets.

For ImageNet, our test data consists of 1,000 images randomly sampled from the validation set. These images are only sampled once and are fixed for all experiments. We did not train models on the ImageNet training data in our experiments.

ImageNette is a ten-class subset of ImageNet. Its original training set and validation set have 9,469 and 3,925 images, respectively. We randomly split its original training set into our 90\% and 10\% training and validation data, and adopt 1,000 images randomly sampled from its original validation set as our test data. The data split and test images are only sampled once and fixed for all experiments. We use the high-resolution version of ImageNette, where all images are larger than $320\times320$.

Because some of our experiments require fine-tuning models on processed training data, we switch to ImageNette to reduce the training cost. We evaluate on ImageNet only when (1) model fine-tuning is not needed, or (2) model fine-tuning is needed but a pre-trained model is publicly available.

\subsection{Models}
\label{app:exp:models}

We adopt various ResNet~\cite{resnet} models mainly depending on the examined defense. All models make the prediction with majority vote over $n=500$ samples if a stochastic defense is applied.

For defenses with low randomness, which require no model fine-tuning, we evaluate them on ImageNet with a ResNet-50 model pre-trained by TorchVision\footnote{\url{https://pytorch.org/vision/stable/models.html}}, which attains 76.13\% Top-1 accuracy and 92.86\% Top-5 accuracy on ImageNet.

For defenses with higher randomness, which require model fine-tuning, we evaluate them on ImageNette with our own ResNet-34 models, detailed as follows.

To first obtain a baseline model for ImageNette, we adopt a ResNet-34 model pre-trained by TorchVision, which attains 73.31\% Top-1 accuracy and 91.42\% Top-5 accuracy on ImageNet. We fine-tune this model on ImageNette's training set with gradient descent for 70 epochs using the AdamW~\cite{adamw} optimizer and the Cosine Annealing~\cite{cosine} learning rate scheduler, where we use batch size 256, initial learning rate 0.001, and weight decay 0.01. We choose the model that performs best on the validation set, which attains 96.9\% Top-1 accuracy on the test set.

We then fine-tune the above baseline ResNet-34 model on training data pre-processed by the defense we examine in each experiment. We adopt the same  training configs as those used to train the baseline model but reduce the number of epochs to 30.

As a special case, when we evaluate randomized smoothing in \Cref{sec:exp:eot}, which requires model fine-tuning with data perturbed by Gaussian noise, we adopt the ResNet-50 models pre-trained on such perturbed ImageNet from \citet{smoothing}. These models attain 67\% and 57\% Top-1 accuracy when the input is perturbed with Gaussian noise of standard deviation 0.25 and 0.50, respectively.

\subsection{Defenses}
\label{app:exp:defenses}

Our main evaluation focuses on two stochastic defenses, detailed as follows.

\textbf{BaRT~\cite{bart}.} The original BaRT defense considers a randomization space of 25 diverse input transformations, and the parameters of each transformation are further randomized. At each inference, it randomly samples $\kappa$ randomized transformations, composites them together in a random order, and applies the composited transformation to the input image.

Since our evaluation only aims to examine the limitations of BaRT but not to break it, it suffices to analyze a \emph{subset} of transformations. Specifically, we consider a randomization space of $\kappa\leq 6$ input transformations and composite all $\kappa$ transformations in a random order to pre-process the input image before feeding it to the classifier. We outline the chosen randomized transformations below and refer to \citet{bart} for more details. Our implementation is available in the code.
\begin{itemize}[leftmargin=*]
	\item \emph{Noise Injection.} This transformation perturbs the input image with noise of distributions and parameters chosen uniformly at random. The set of candidate noise distributions includes Gaussian, Poisson, Salt, Pepper, Salt and Pepper, and Speckle.
	\item \emph{Gaussian Blur.} This transformation blurs the input image using a Gaussian filter with the kernel size randomly chosen from $[2, 14]$ and the standard deviation randomly chosen from $[0.1, 3.1]$.
	\item \emph{Median Blur.} This transformation blurs the input image using a median filter with the kernel size randomly chosen from $[2, 14]$.
	\item \emph{Swirl Transformation.} This transformation applies the swirl transformation\footnote{\url{https://scikit-image.org/docs/stable/auto_examples/transform/plot_swirl.html}}, a non-linear image deformation that creates a whirlpool effect. Its strength, radius, and location are chosen uniformly at random from $[0.1, 2.0]$, $[10, 200]$, and $[1, 200]$, respectively. We adopt BPDA~\cite{bpda} with the identity function to handle the non-differentiable problem.
	\item \emph{Quantization.} This transformation quantizes the input image's pixel values within $[0, 1]$ to a limited number of bins, where the number of bins is chosen uniformly at random from $[8, 200]$. For example, if the number of bins is set to 4, all pixels will be quantized to $\s{0.00, 0.25, 0.50, 0.75, 1.00}$.
	\item \emph{FFT Perturbation.} This transformation perturbs the 2D FFT of each channel of the input image. For each channel in the frequency domain, it randomly zeros out a fraction of coefficients. The fraction is chosen uniformly at random from $[0.00, 0.95]$.
\end{itemize}

In our setting, we form the randomization space by compositing the first $\kappa$ transformations in random order. While increasing the space of transformations typically leads to a more effective defense, randomly compositing transformations may not always lead to stronger defenses. For example, the quantization may decrease the effectiveness of other transformations. However, this drawback \emph{does not} affect our evaluation, as our main objective is to compare the defense's performance before and after the defended model achieves higher invariance. Rigorous comparisons between the defense's performance before and after increasing the randomness are largely orthogonal to our work.

\textbf{Randomized Smoothing~\cite{smoothing}.}
Randomized smoothing adds Gaussian noise to the input image and makes predictions with majority vote over a large number of samples. This defense was initially proposed for certifiable adversarial robustness. In our evaluation, we adopt this defense to examine (1) how randomness affects the effectiveness of applying EOT and (2) how invariance affects the robustness provided by the defense. Specifically, we control the level of randomness by varying the added Gaussian noise's standard deviation $\sigma$. For evaluation on ImageNet, we choose $\sigma\in\s{0.25, 0.50}$ as models pre-trained on data perturbed by such noise are available from \citet{smoothing}. For evaluation on ImageNette, we are able to scale the evaluation for $\sigma$ from 0.10 to 0.50 with a step size of 0.05. We ignore the \emph{abstain} output in the original defense, as we do not study the certification.

\subsection{Attacks}
\label{app:exp:attacks}

We evaluate defenses with standard PGD combined with EOT and focus on the \LL{\infty}-bounded adversary with a perturbation budget $\epsilon=8/255$ in both untargeted and targeted settings. We do not introduce any techniques other than EOT to explicitly handle the randomness, such as random restarts~\cite{autopgd} and momentum-based optimizers~\cite{aggmopgd}. We also utilize AutoPGD~\cite{autopgd} to avoid selecting the best step size when it is computationally expensive to repeat some experiments. More importantly, we only conduct adaptive evaluations, where the defense is always included in the attack loop with non-differentiable components approximated by the identity function~\cite{bpda}. For targeted attacks, we choose the last class of each dataset as the target label: ImageNet (999) and ImageNette (9).

We adopt various attack settings depending on the experiment, as detailed below.

In \Cref{sec:exp:eot}, we aim to evaluate the benefits of applying EOT under different settings. For this end, we apply the standard PGD attack of $k\in\s{10, 20, 50, 100, 200, 500, 1000}$ steps and combine them with EOT of $m\in\s{1,5,10,20}$ samples. For each combination, we further test several step sizes chosen from $\alpha\in\s{0.5/255, 1/255, 2/255, 4/255}$ and report their best performance.

In \Cref{sec:exp:tradeoff}, we aim to evaluate the trade-off between the defense's robustness and the model's invariance to the added randomness. For this end, we evaluate the defense's performance when it applies to models of different levels of invariance. Since it is computationally expensive to repeat the attack for multiple step sizes, we utilize AutoPGD~\cite{autopgd} to tune the step size automatically. In this experiment, we evaluate all defenses with AutoPGD of 200 steps and disable all techniques designed to capture the randomness, including EOT.

We make this choice due to three considerations. First, disabling EOT effectively reduces the computational cost. Second, we already showed in \Cref{sec:exp:eot} that PGD attacks can already assess the robustness of stochastic defenses without applying EOT. Last but not least, our objective is to evaluate the defense's performance when it applies to different models \emph{under the same attack}. Under this setting, we can observe that the same attack (regardless of its strength) that hardly works for the defense (before fine-tuning \& low invariance) now becomes more effective (after fine-tuning \& high invariance). We can surely run each attack for more iterations and samples, but the current setting suffices to show that the defense provides robustness by explicitly reducing invariance.

\textbf{Computing Resources.}
All experiments are conducted on two Linux workstations, each with 48 Intel Xeon CPUs and 8 GeForce RTX 2080 Ti GPUs. We only train ResNet-34 models on ImageNette without the distributed setting. Standard training (70 epochs) takes 35 minutes. Training with data processed by Gaussian noise (30 epochs) takes 15 minutes. Training with data processed by BaRT (30 epochs) takes 18 minutes for $\kappa\in\s{1,2}$, 70 minutes for $\kappa\in\s{3,4,5}$, and 3 hours for $\kappa\in\s{6}$. The training time for $\kappa\geq3$ is higher due to CPU-bounded transformations; implementing such transformations using native PyTorch operations on GPUs should decrease the training cost.

\FloatBarrier

\section{More Experiment Results}
\label{app:exp:results}

In this section, we provide more experiment results that strengthen discussions in the main paper.

\subsection{PGD Captures Randomness with Fine-grained Learning Rates}
\label{app:exp:results:lr}

\begin{figure}[tb]
    \centering
    \begin{subfigure}[t]{0.35\textwidth}
        \includegraphics[width=\linewidth]{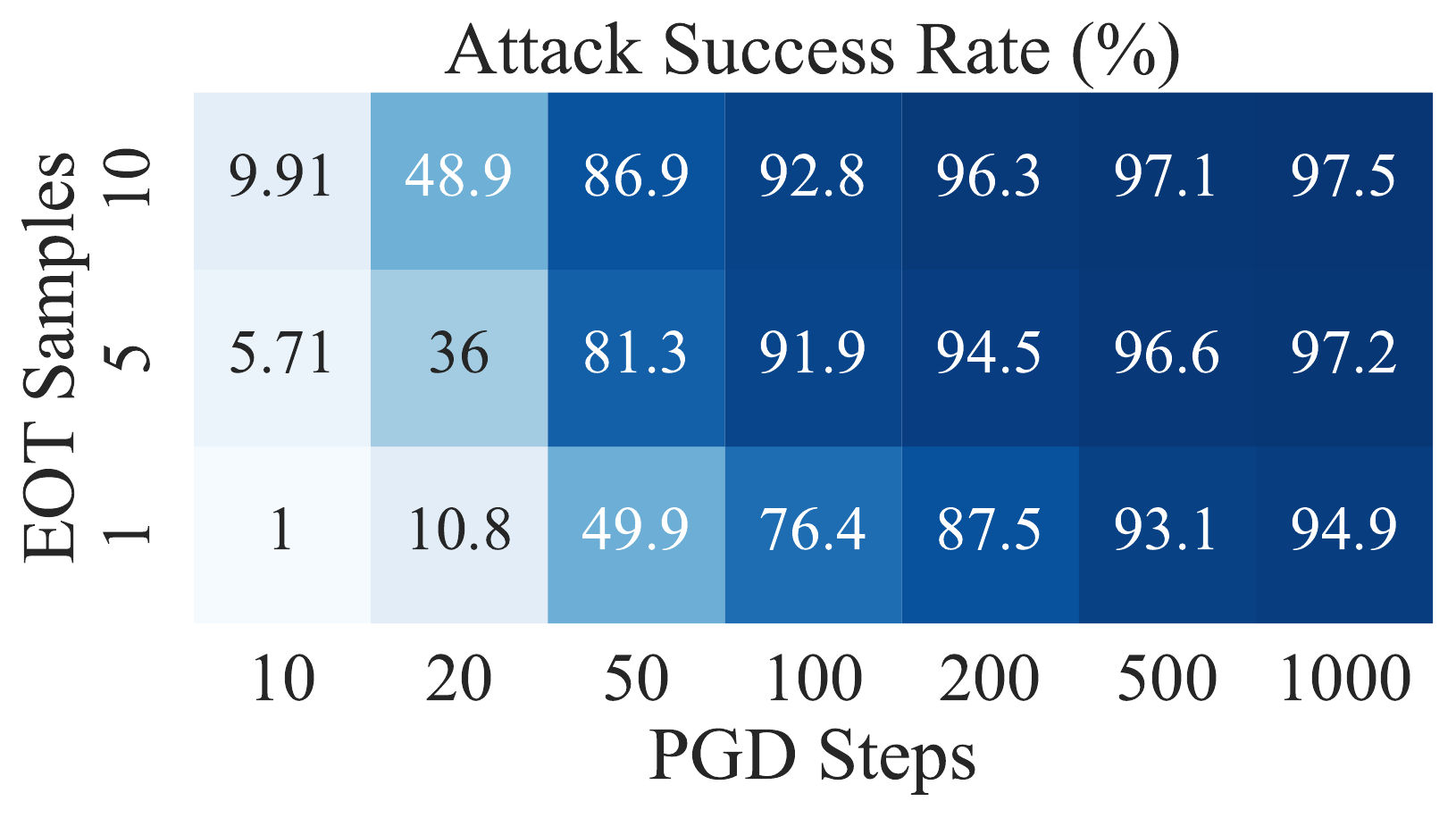}
        \caption{step size = 0.5/255}
    \end{subfigure}
    \hspace{3em}
    \begin{subfigure}[t]{0.35\textwidth}
        \includegraphics[width=\linewidth]{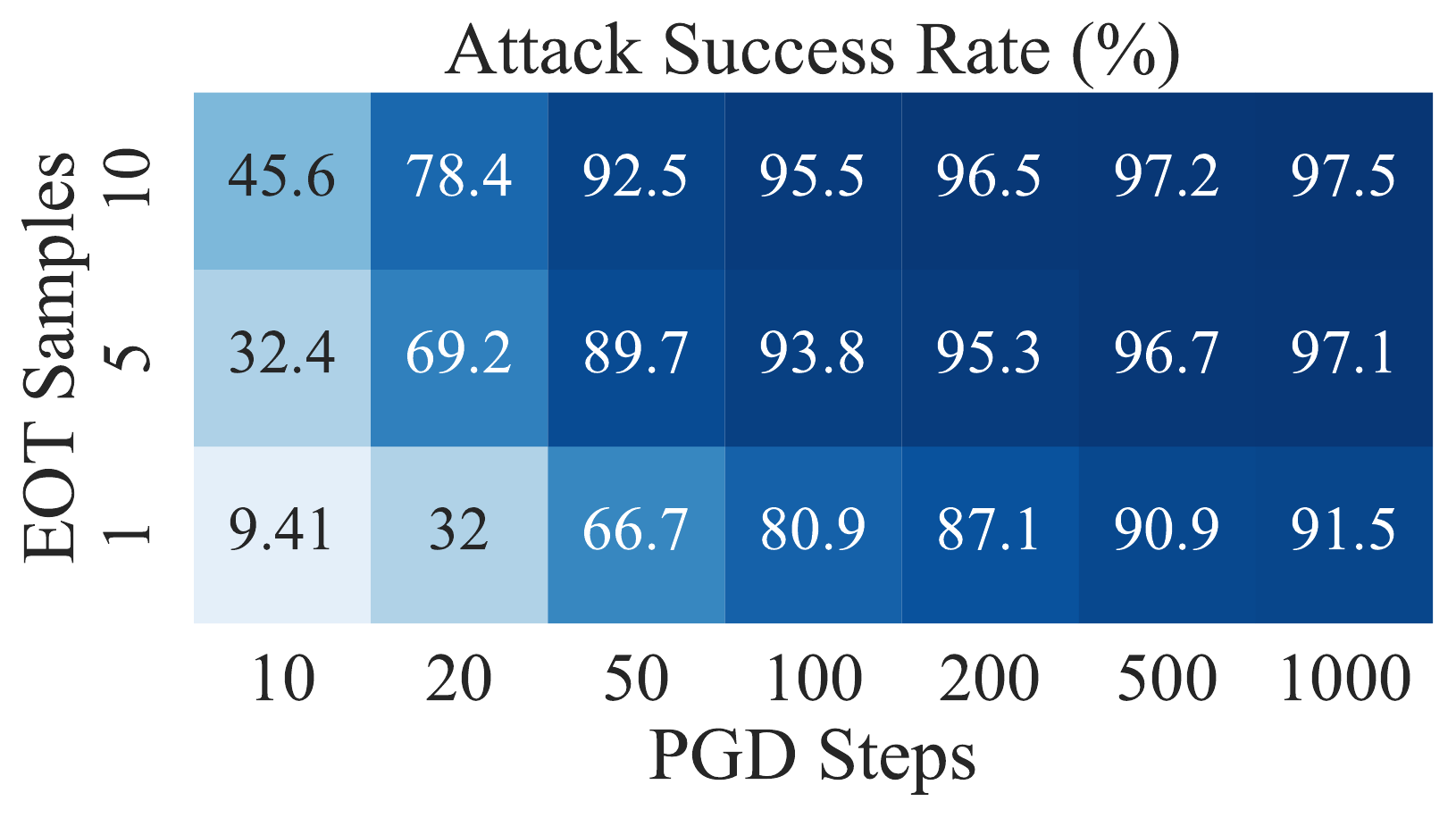}
        \caption{step size = 1.0/255}
    \end{subfigure}

    \begin{subfigure}[t]{0.35\textwidth}
        \includegraphics[width=\linewidth]{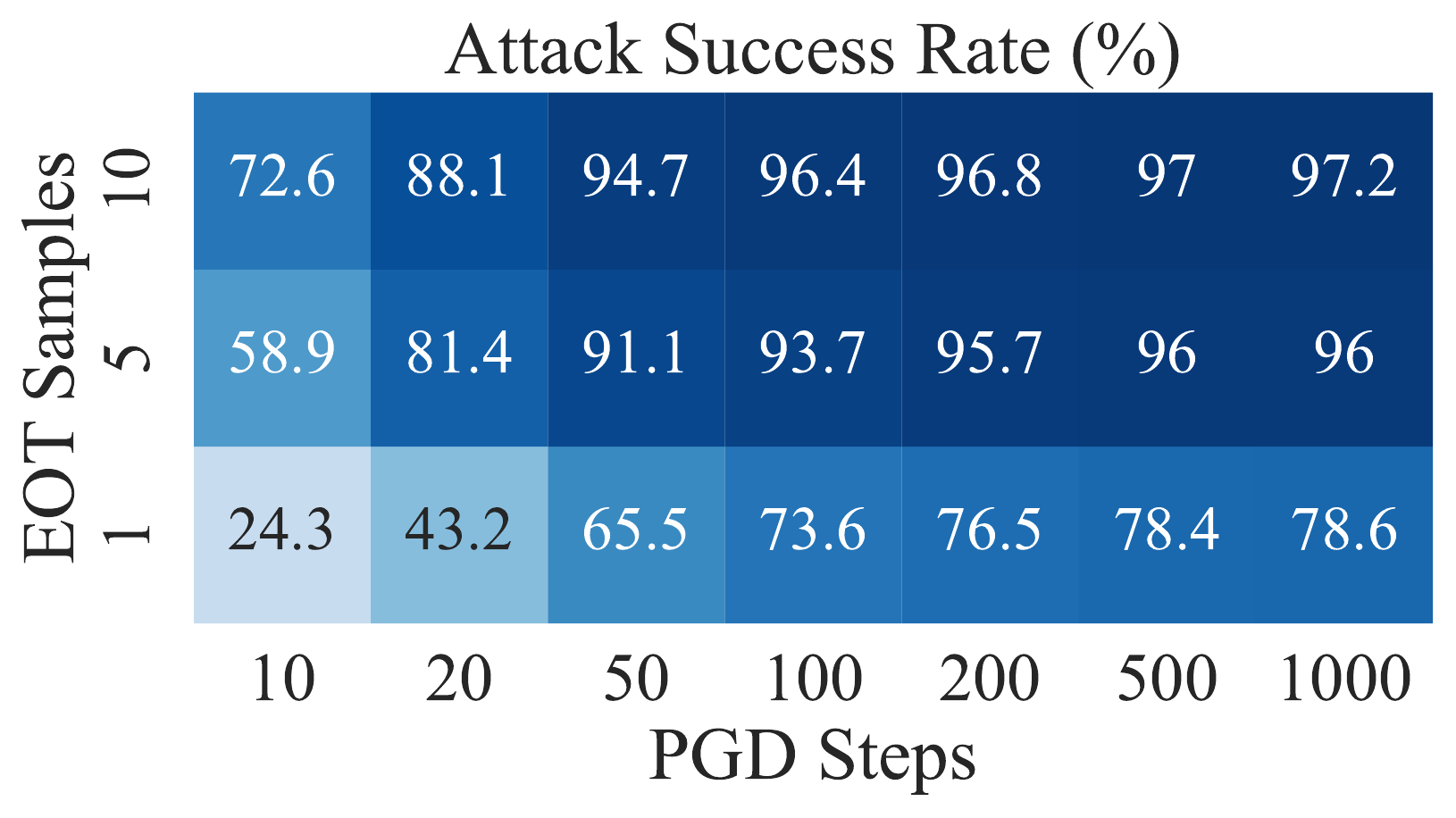}
        \caption{step size = 2.0/255}
    \end{subfigure}
    \hspace{3em}
    \begin{subfigure}[t]{0.35\textwidth}
        \includegraphics[width=\linewidth]{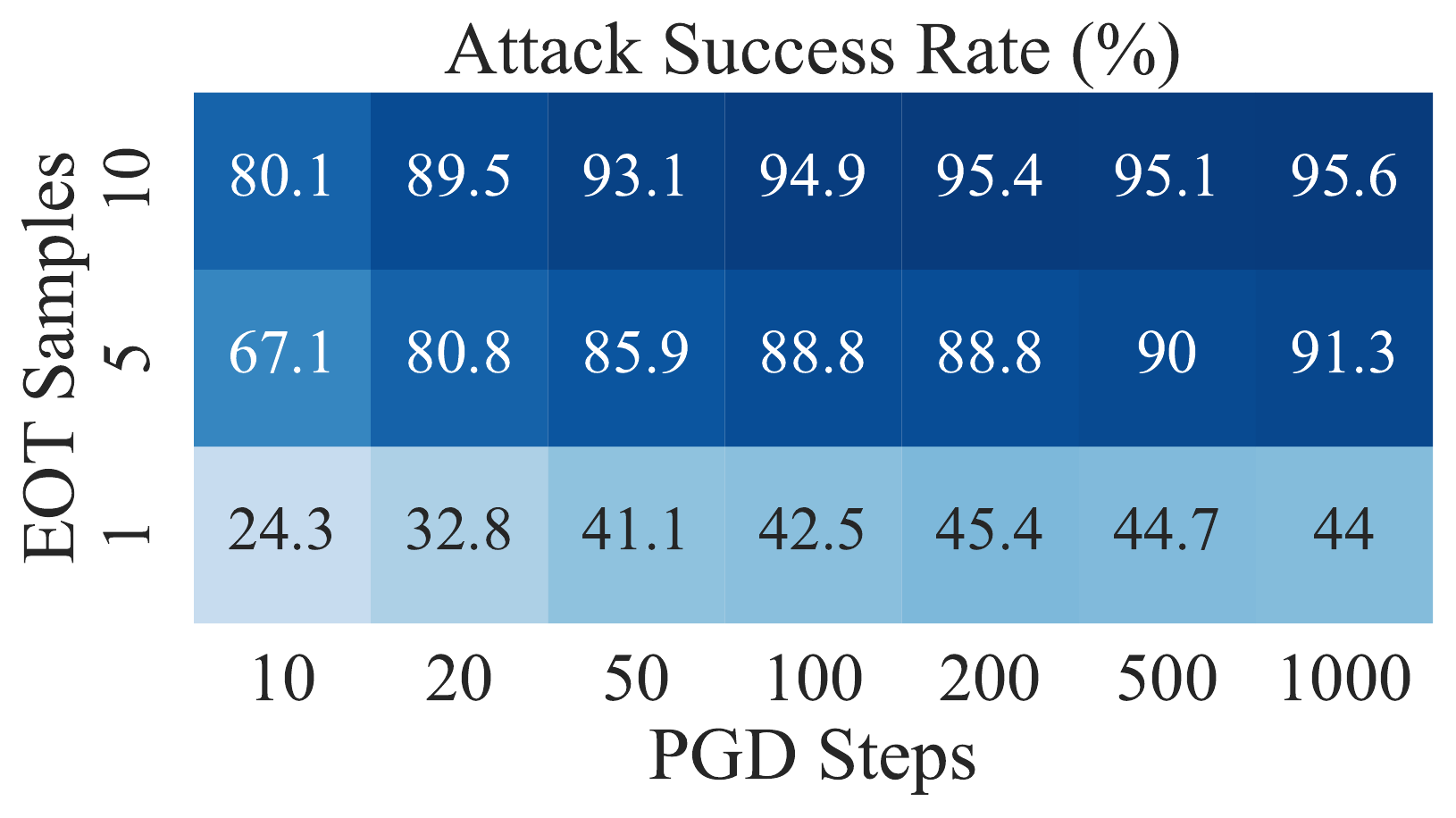}
        \caption{step size = 4.0/255}
    \end{subfigure}
    \caption{Evaluation of randomized smoothing on ImageNet (targeted attacks, $\sigma=0.25$).}
    \label{fig:app:lr1}
\end{figure}

During our evaluation, we recognize that the effectiveness of PGD attacks, when combined with EOT, is sensitive to the choice of step size (i.e., learning rate). Here, we provide the full results of four different choices of step size when evaluating the randomized smoothing defense. The results with step size chosen from $\alpha\in\s{0.5/255, 1.0/255, 2.0/255, 4.0/255}$ are shown in \Cref{fig:app:lr1}.

As we can observe, standard PGD without EOT achieves better performance when the step size is small, yet the application of EOT requires larger step sizes to perform better. We conjecture that this is because EOT reduces the variance of gradients so the attack algorithm can take a larger step, yet standalone PGD only gets noisy gradients and is only ``confident'' to take a small step.

\begin{figure}[tb]
    \centering
    \begin{subfigure}[t]{0.49\textwidth}
        \includegraphics[width=\linewidth]{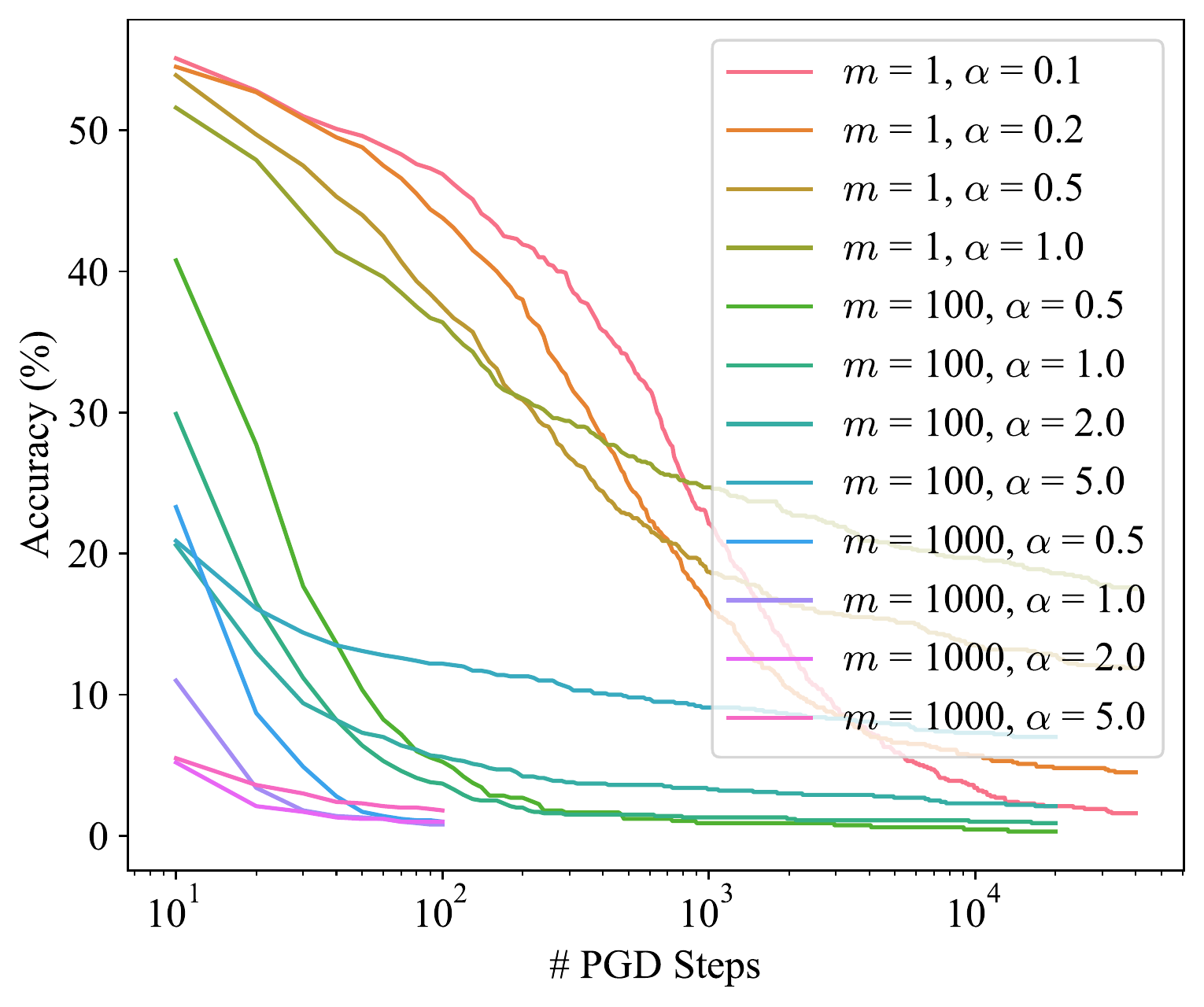}
        \caption{Accuracy under Attack (view by PGD Steps)}
        \label{fig:app:lr3-1}
    \end{subfigure}
    \begin{subfigure}[t]{0.49\textwidth}
        \includegraphics[width=\linewidth]{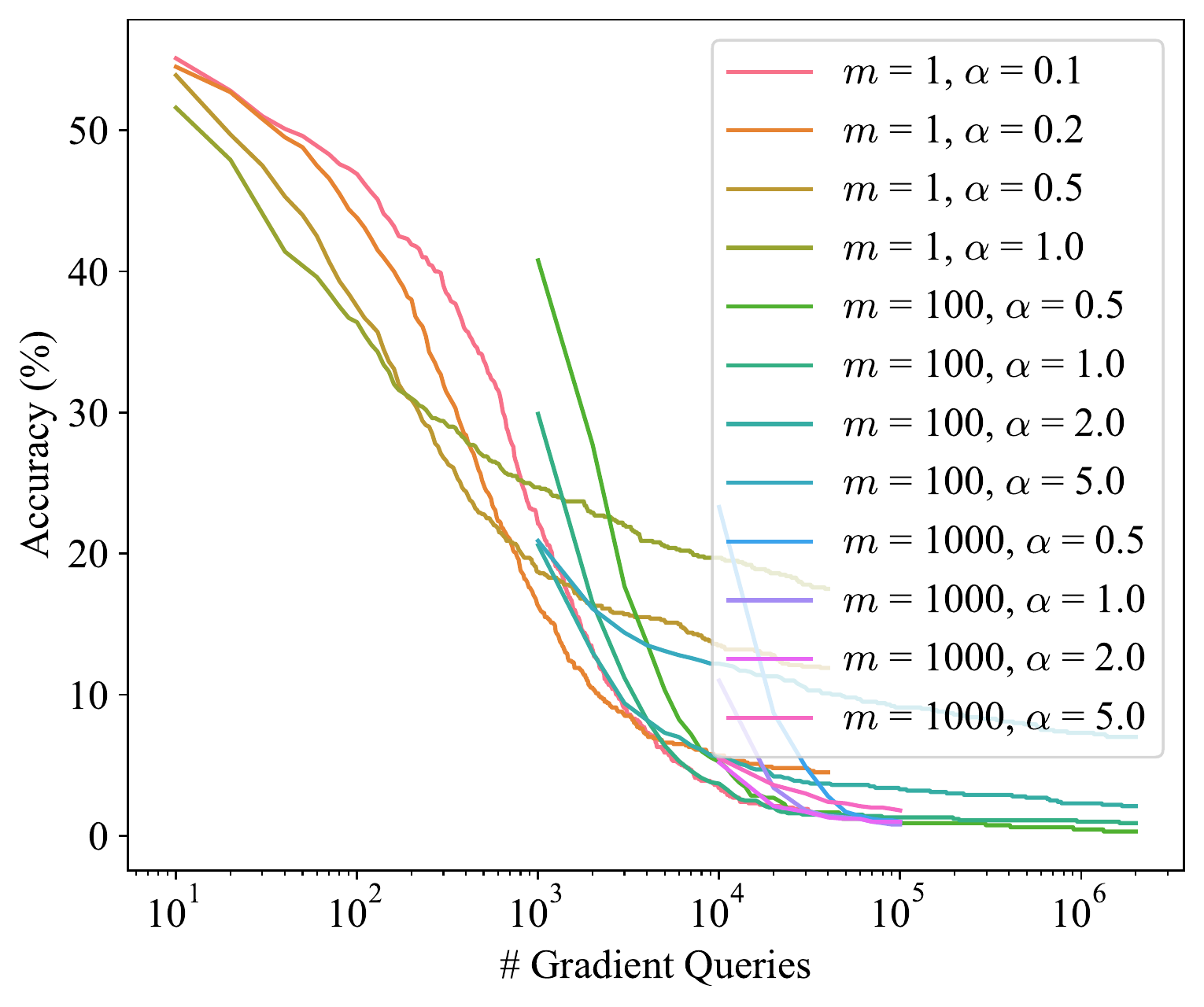}
        \caption{Accuracy under Attack (view by Gradient Queries)}
        \label{fig:app:lr3-2}
    \end{subfigure}
    \caption{Evaluation of the discontinuous activation~\cite{kwinners} defense with EOT-$m$ and step size $\alpha$.}
    \label{fig:app:lr3}
\end{figure}

However, this may not always prevent PGD from converging to a competitive solution. For example, we evaluate the discontinuous activation~\cite{kwinners} defense with different attack settings, where the attack adds Gaussian noise around the input to estimate the correct gradient. The convergence curves of different settings are demonstrated in \Cref{fig:app:lr3}.

When we examine the convergence in terms of PGD steps in \Cref{fig:app:lr3-1}, applying EOT obtains better gradients and quickly decreases the defended model's accuracy to zero. However, when we examine the convergence in terms of the total number of gradient queries in \Cref{fig:app:lr3-2}, we observe that (1) PGD without EOT given a smaller learning rate and (2) PGD with EOT given a larger learning rate have almost the same convergence behavior. This interesting observation suggests that standard PGD attacks may be sufficient in some cases if using a carefully fine-tuned learning rate. For example, \citet{aggmopgd} showed that PGD attacks could be significantly improved by applying the AggMo~\cite{aggmo-optimizer} optimizer, which leverages multiple momentum terms.

%\begin{figure}[tb]
%    \centering
%    \begin{subfigure}[t]{0.35\textwidth}
%        \includegraphics[width=\linewidth]{imgs/gaussian_pgd_targeted_smooth_var0.50_lr0.5_app.pdf}
%        \caption{step size = 0.5/255}
%    \end{subfigure}
%    \hspace{3em}
%    \begin{subfigure}[t]{0.35\textwidth}
%        \includegraphics[width=\linewidth]{imgs/gaussian_pgd_targeted_smooth_var0.50_lr1.0_app.pdf}
%        \caption{step size = 1.0/255}
%    \end{subfigure}
%
%    \begin{subfigure}[t]{0.35\textwidth}
%        \includegraphics[width=\linewidth]{imgs/gaussian_pgd_targeted_smooth_var0.50_lr2.0_app.pdf}
%        \caption{step size = 2.0/255}
%    \end{subfigure}
%    \hspace{3em}
%    \begin{subfigure}[t]{0.35\textwidth}
%        \includegraphics[width=\linewidth]{imgs/gaussian_pgd_targeted_smooth_var0.50_lr4.0_app.pdf}
%        \caption{step size = 4.0/255}
%    \end{subfigure}
%    \caption{Evaluation of randomized smoothing on ImageNet (targeted attacks, $\sigma=0.50$).}
%    \label{fig:app:lr2}
%\end{figure}

\subsection{Inability to Remove Invariance that Does Not Hurt the Utility}
\label{app:exp:results:utility}
\begin{figure}[tb]
    \centering
    \begin{subfigure}[t]{0.32\textwidth}
        \includegraphics[width=\linewidth]{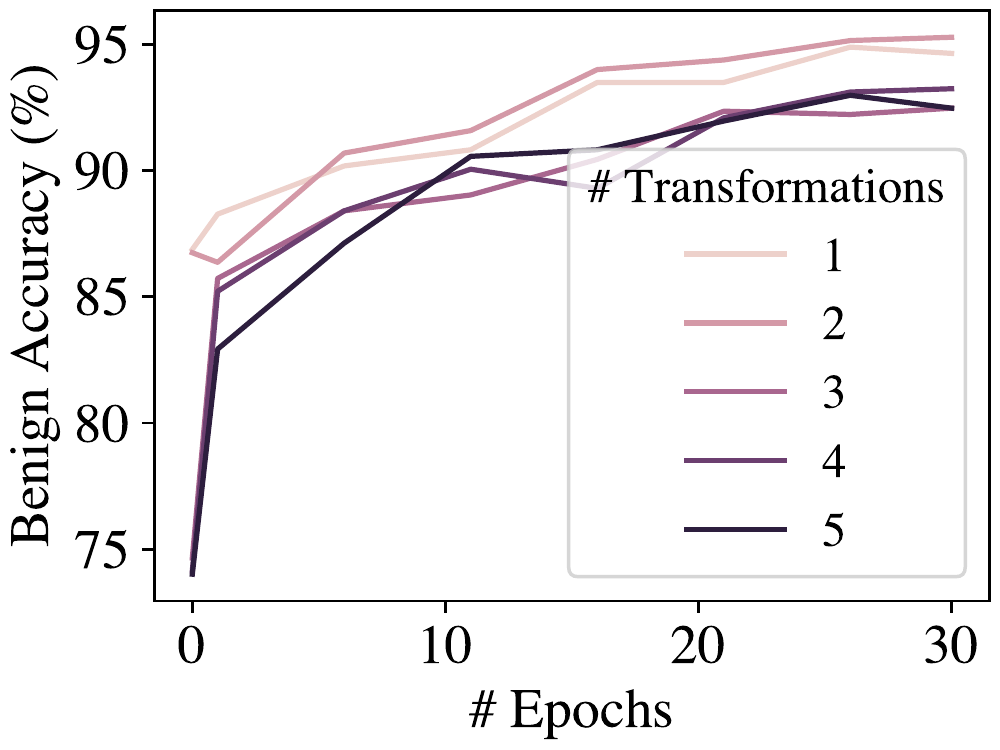}
        \caption{No Attacks}
    \end{subfigure}
%    \hspace{1em}
    \begin{subfigure}[t]{0.32\textwidth}
        \includegraphics[width=\linewidth]{imgs/invariance_bart_targeted_by_epoch.pdf}
        \caption{Targeted Attacks}
    \end{subfigure}
%    \hspace{1em}
    \begin{subfigure}[t]{0.32\textwidth}
        \includegraphics[width=\linewidth]{imgs/invariance_bart_untargeted_by_epoch.pdf}
        \caption{Untargeted Attacks}
    \end{subfigure}
    \caption{Performance of the BaRT defense on ImageNette and models during fine-tuning.}
    \label{fig:app:fine-tune}
\end{figure}

 Some recent works also suggest that one could gain robustness by removing invariance that does not hurt the utility~\cite{shift-invariance}. However, this may not be the case for defenses with a larger randomization space. For example, we evaluate the performance of BaRT when it applies to models during the fine-tuning process (same experiment as in \Cref{sec:exp:tradeoff}). As shown in \Cref{fig:app:fine-tune}, the robustness has already dropped significantly before the model achieves invariance that preserves most of the utility. 

%\clearpage

\subsection{Visualization of Adversarial Perturbation}
\label{app:exp:vis}

\begin{figure}[h]
    \centering
    \begin{subfigure}[t]{0.34\textwidth}
        \centering
        \includegraphics[width=\linewidth]{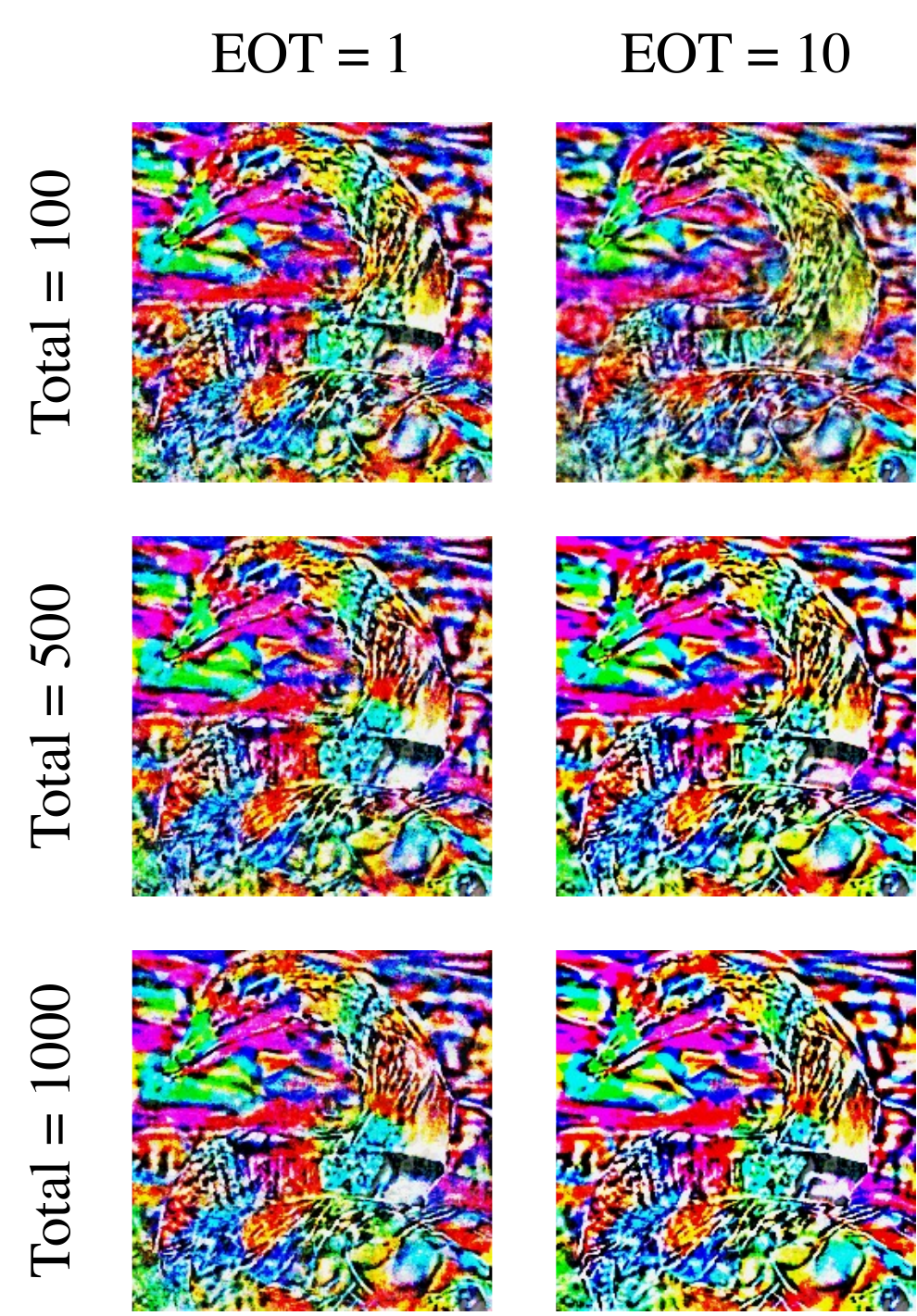}
        \caption{$\sigma=0.25$}
    \end{subfigure}
    \quad
    \begin{subfigure}[t]{0.293\textwidth}
        \centering
        \includegraphics[width=\linewidth]{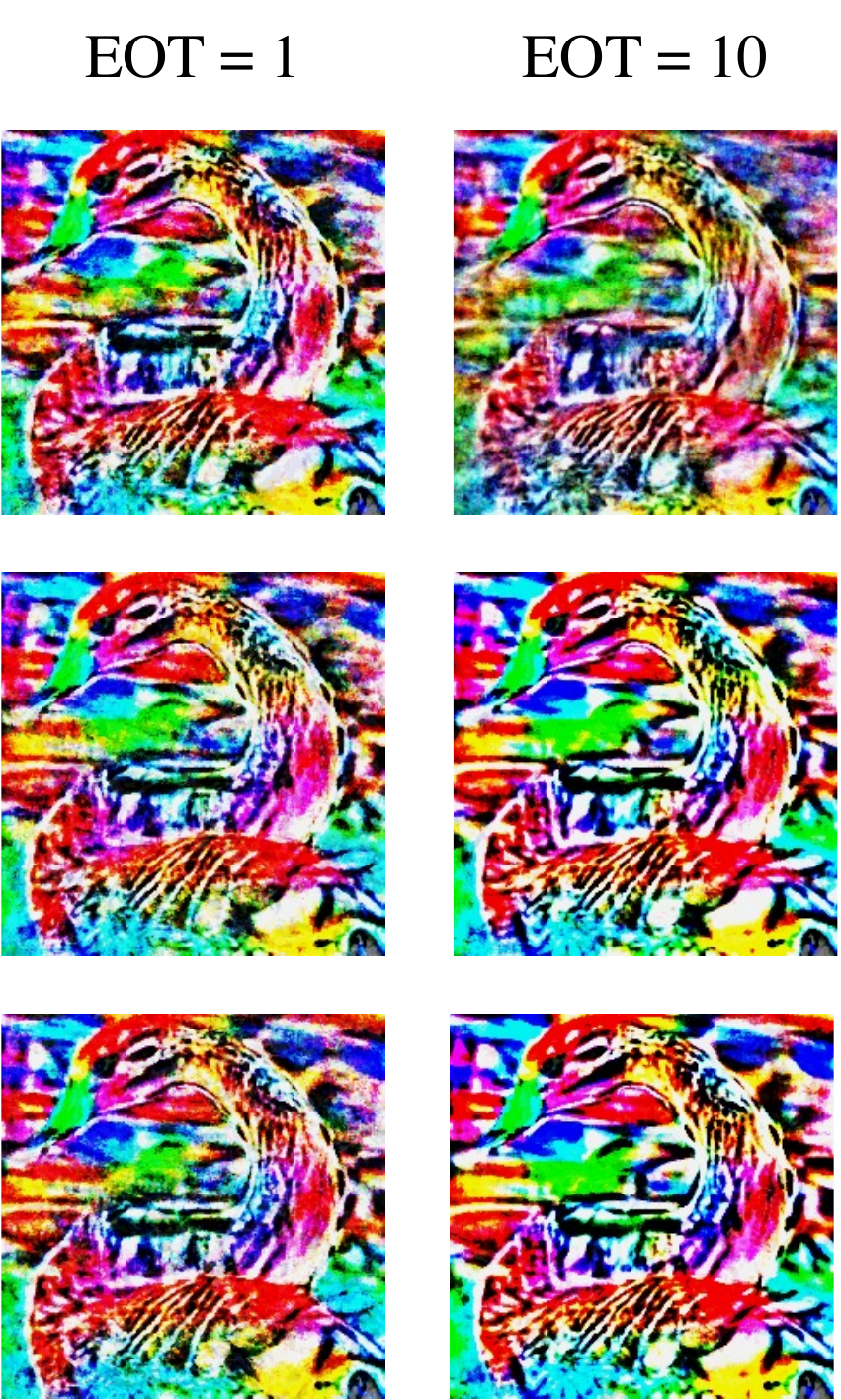}
        \caption{$\sigma=0.50$}
    \end{subfigure}
    \quad
    \begin{subfigure}[t]{0.293\textwidth}
        \centering
        \includegraphics[width=\linewidth]{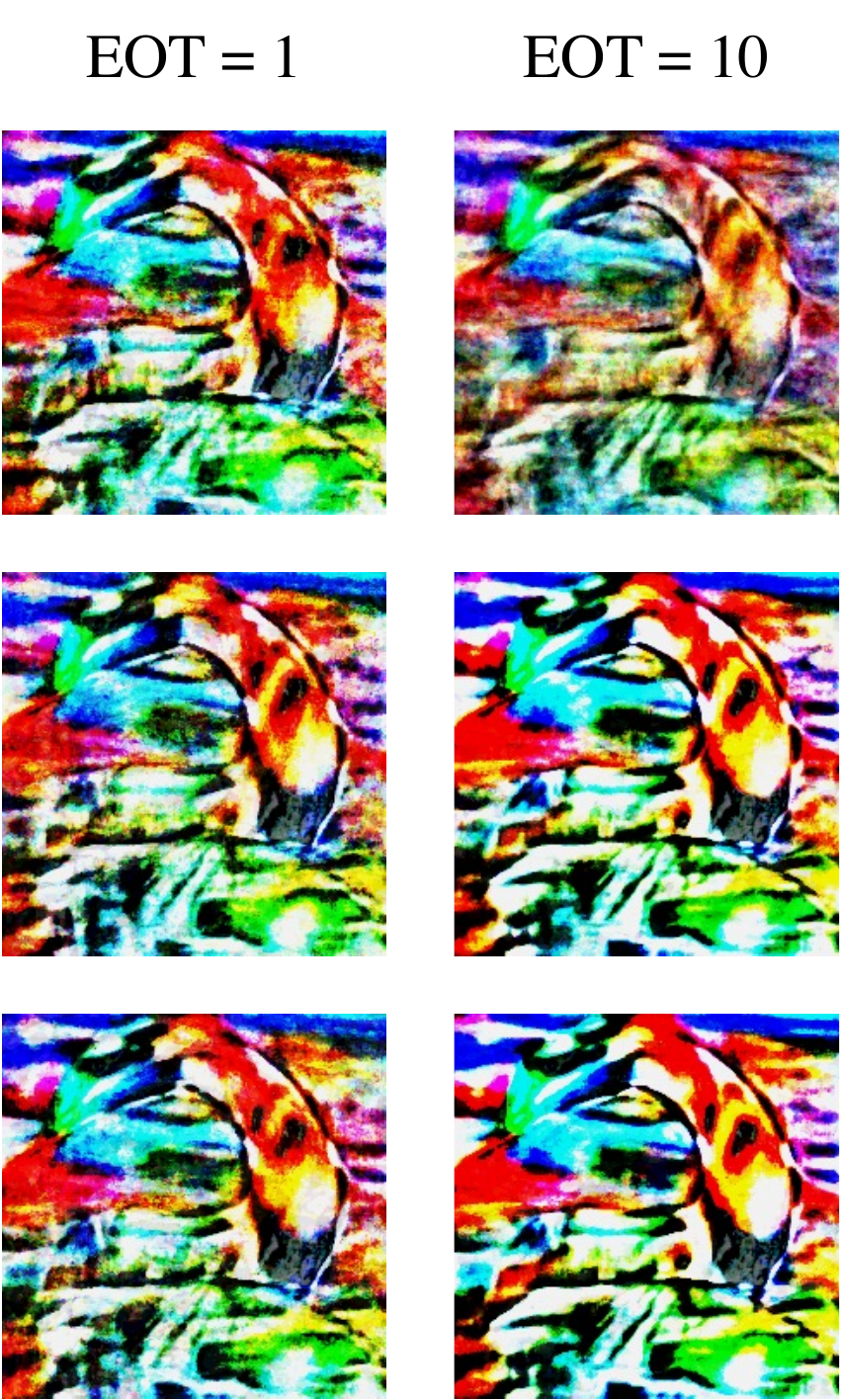}
        \caption{$\sigma=1.00$}
    \end{subfigure}
    
    \caption{Adversarial perturbation created by \emph{untargeted} PGD attacks with and without EOT.}
    \label{fig:vis1}
\end{figure}

\begin{figure}[h]
    \centering

    \begin{subfigure}[t]{0.34\textwidth}
        \centering
        \includegraphics[width=\linewidth]{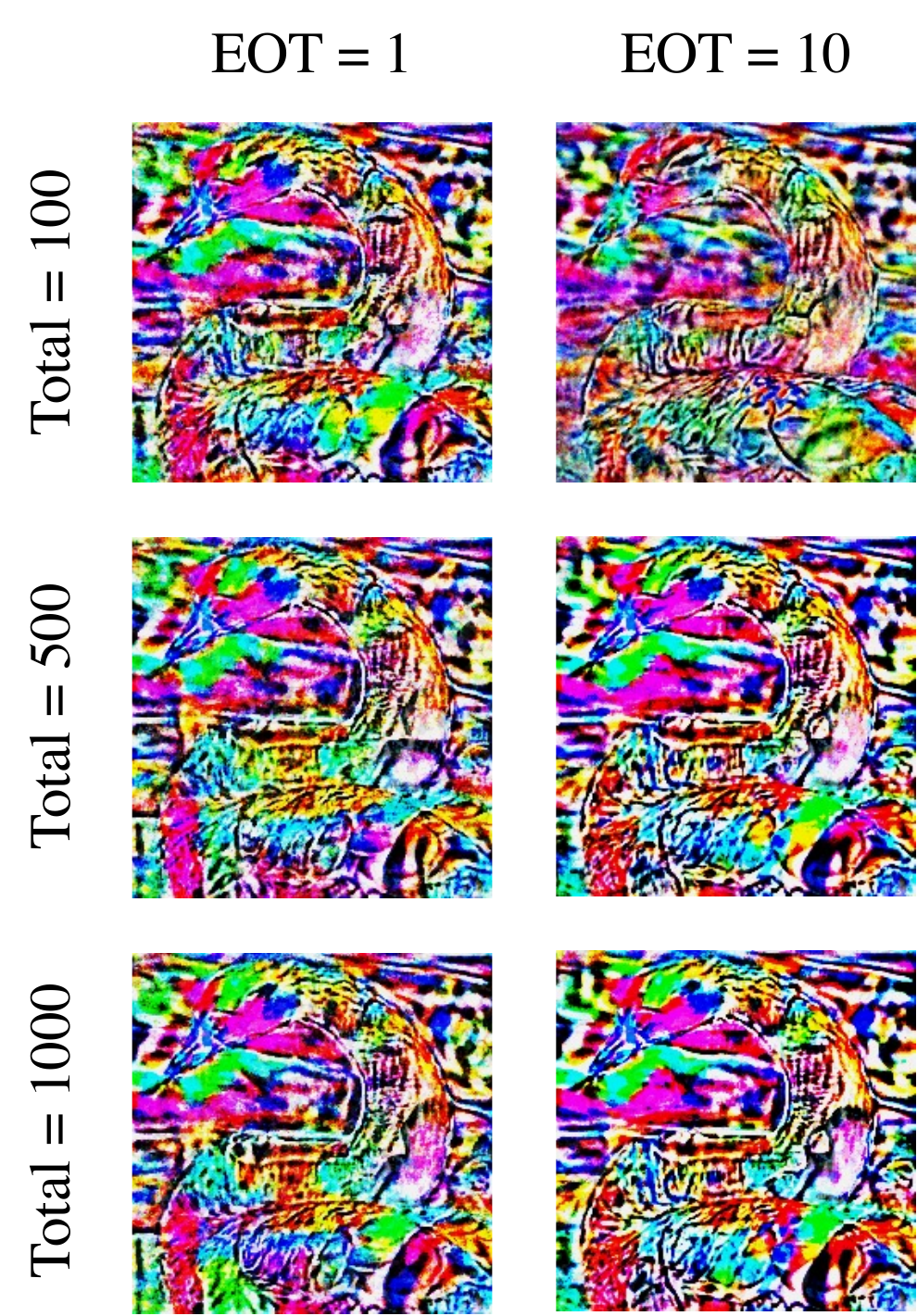}
        \caption{$\sigma=0.25$}
    \end{subfigure}
    \quad
    \begin{subfigure}[t]{0.293\textwidth}
        \centering
        \includegraphics[width=\linewidth]{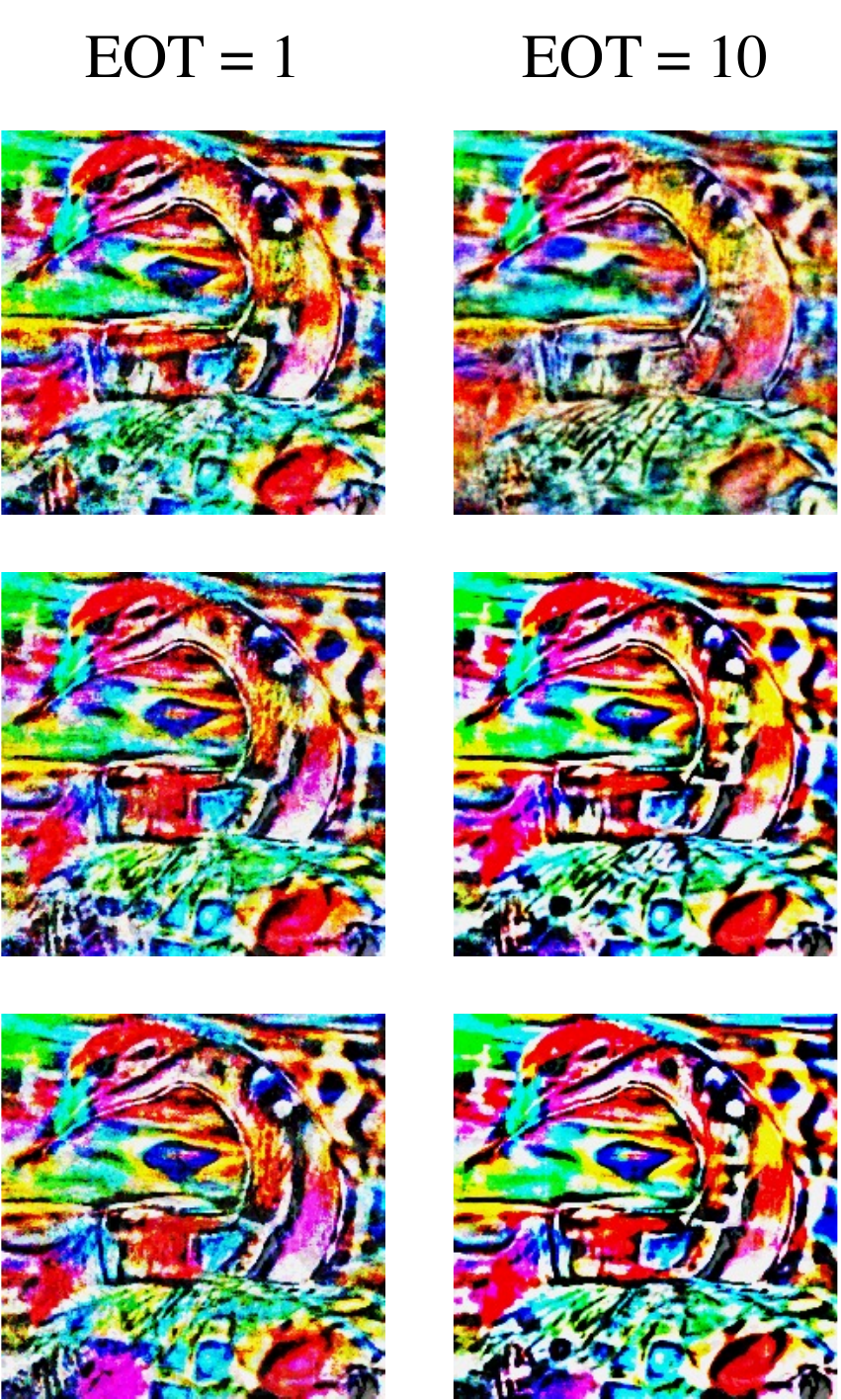}
        \caption{$\sigma=0.50$}
    \end{subfigure}
    \quad
    \begin{subfigure}[t]{0.293\textwidth}
        \centering
        \includegraphics[width=\linewidth]{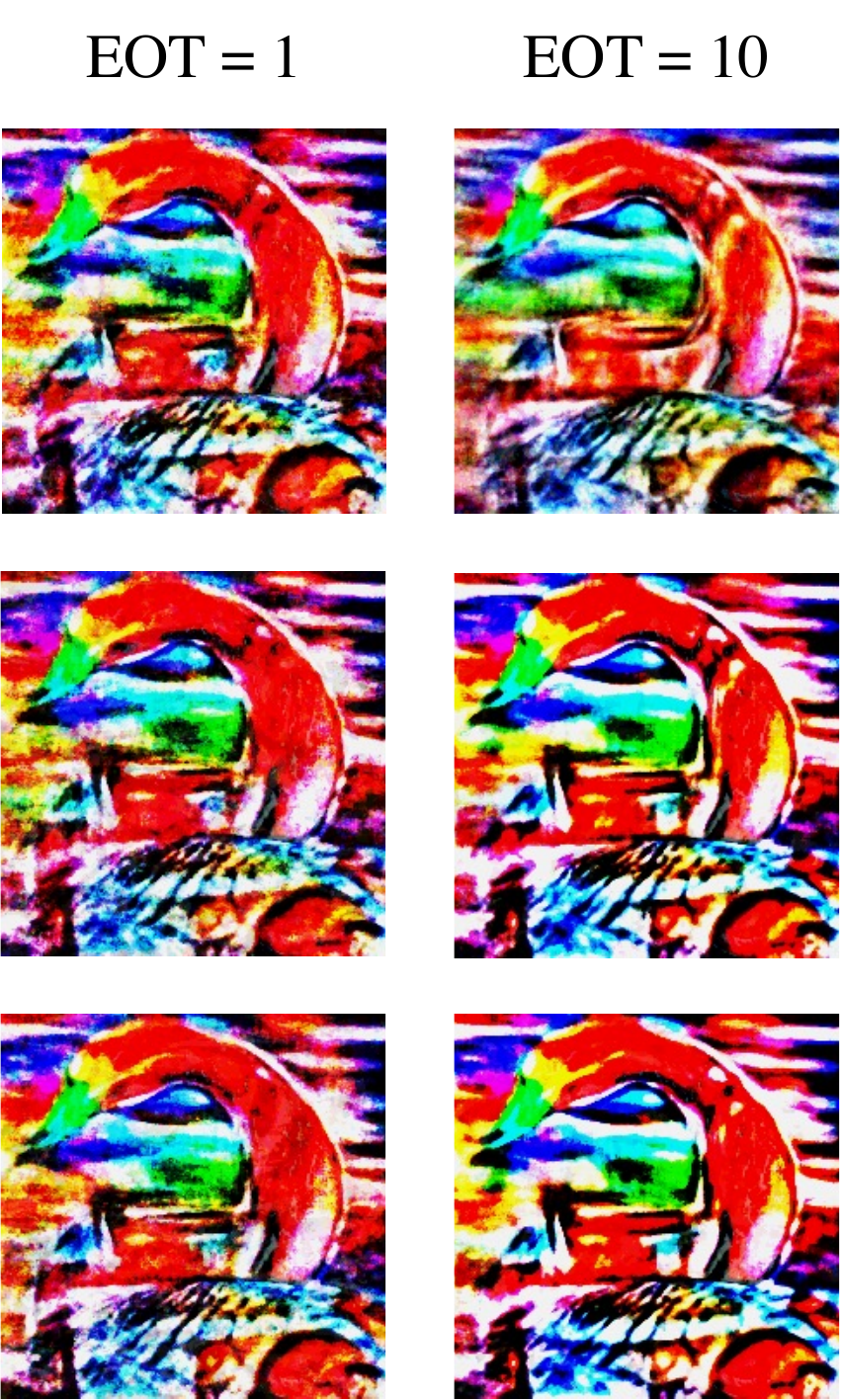}
        \caption{$\sigma=1.00$}
    \end{subfigure}
    
    \caption{Adversarial perturbation created by \emph{targeted} PGD attacks with and without EOT.}
    \label{fig:vis2}

\end{figure}

\Cref{fig:vis1,fig:vis2} show the adversarial perturbation created by PGD attacks with and without EOT. 

\textbf{Settings.} When given $C$ gradient queries in total, we run PGD for (1) $C$ steps without EOT and (2) $C/10$ steps with EOT of 10 samples. All attacks use \LL{\infty}-norm budget $\epsilon=8/255$ and step size $\alpha=1/255$. The target model is a ResNet-50 defended by randomized smoothing with Gaussian noise $\sigma\in\s{0.25, 0.50, 1.00}$. We adopt pre-trained models from \citet{smoothing}.

\textbf{Visualization.} We randomly choose an image (id 5000) from the ImageNet validation set. For the benign image $x$ and its adversarial example $x^\prime$, the perturbation is written as $\delta\coloneqq x^\prime-x\in[-\epsilon, \epsilon]$. We normalize it to $\delta^\prime\coloneqq\delta / (2\epsilon) + 0.5\in[0, 1]$ and multiply it by 0.95 for better visualization.

\textbf{Compare PGD with and without EOT.} For models of the same noise level, applying EOT leads to slightly smoother (or less noisy) adversarial perturbation. This observation shows that EOT computes more stable gradients. Besides, the above effect becomes more significant when (1) the model has a higher level of randomness (i.e., large $\sigma$), or (2) the attack runs in the targeted mode. These are the scenarios where applying EOT benefits more, which correspond to our findings in \Cref{sec:exp:eot}.

%\clearpage

\textbf{Compare PGD on models with different degrees of randomness.} If we compare the visualization across different models, we can observe that models with a higher level of randomness produce smoother adversarial perturbation (even without applying EOT). While this observation seems counterintuitive, we note that these models are all fine-tuned on noisy data. As a result, making the model invariant to randomness also smoothes out the gradient, which removes the expected robustness provided by randomness. This observation corresponds to our theoretical model in \Cref{sec:tradeoff} and empirical findings in \Cref{sec:exp:tradeoff}.

\subsection{Additional Experiments on CIFAR10}
\label{app:exp:cifar10}

In \Cref{tab:previous-defenses}, we evaluate a few defenses on CIFAR10, including randomized activation pruning~\cite{activation-pruning} and discontinuous activation~\cite{kwinners}. Such defenses cannot defend against standard PGD attacks without applying EOT. This shows that our findings hold on small and large input spaces.

In this section, we add an experiment to show that our findings in \Cref{sec:exp:tradeoff} also hold on CIFAR10. To this end, we evaluate the randomized smoothing defense~\cite{smoothing} on CIFAR10 with the more challenging targeted attack. Specifically, we run the standard PGD attack for 100 steps, with budget $\epsilon=8/255$, step size $\alpha=1/255$, target label 9, and EOT of $m=20$ samples. The target models are ResNet-110 pre-trained on noisy data by \citet{smoothing}. We evaluate four different noise levels $\sigma\in\s{0.12, 0.25, 0.50, 1.00}$.
For each noise level, we run the \emph{same} attack on models before and after fine-tuning on data perturbed by such noise. The results are shown in \Cref{tab:cifar_acc,tab:cifar_asr}.

For all noise levels, fine-tuning models to obtain invariance improves the benign accuracy as expected. During this procedure, however, we can observe that the defense becomes less effective when the model recovers more invariance. In particular, the attack is nearly ineffective for $\sigma\in\s{0.25, 0.50, 1.00}$ when the model has low invariance, yet starts to work as the model recovers invariance. This observation is consistent with our findings on ImageNet in \Cref{sec:exp:tradeoff}.

\begin{table}[h]
\centering
\caption{Benign accuracy of models with low and high invariance to the defense's randomness.}
\label{tab:cifar_acc}
\begin{tabular}{@{}rrrrr@{}}
\toprule
                   & \multicolumn{1}{c}{$\sigma=0.12$} & \multicolumn{1}{c}{$\sigma=0.25$} & \multicolumn{1}{c}{$\sigma=0.50$} & \multicolumn{1}{c}{$\sigma=1.00$} \\ \midrule
Before Fine-tuning (Low Invariance) & 23.4\%                            & 14.7\%                            & 12.3\%                            & 10.1\%                            \\
After Fine-tuning (High Invariance)  & \textbf{83.6}\%                            & \textbf{77.9}\%                            & \textbf{71.1}\%                            & \textbf{56.7}\%                            \\ \bottomrule
\end{tabular}
\end{table}

\begin{table}[h]
\centering
\caption{Attack success rate on models with low and high invariance to the defense's randomness.}
\label{tab:cifar_asr}
\begin{tabular}{@{}rrrrr@{}}
\toprule
                   & \multicolumn{1}{c}{$\sigma=0.12$} & \multicolumn{1}{c}{$\sigma=0.25$} & \multicolumn{1}{c}{$\sigma=0.50$} & \multicolumn{1}{c}{$\sigma=1.00$} \\ \midrule
Before Fine-tuning (Low Invariance) & 52.1\%                            & 1.1\%                             & 0.0\%                             & 0.0\%                             \\
After Fine-tuning (High Invariance) & \textbf{63.1}\%                            & \textbf{29.5}\%                            & \textbf{18.1}\%                            & \textbf{12.3}\%                           \\ \bottomrule
\end{tabular}
\end{table}

%\clearpage
\section{More Discussions}
\label{app:discuss}

\subsection{Discussions about DiffPure~\cite{diffpure}}
\label{app:discuss:diffpure}

In parallel to our work, DiffPure~\cite{diffpure} adopts a complicated stochastic diffusion process to purify the input images. This defense belongs to an existing line of research that leverages generative models to pre-process input images and hence removing the potential adversarial perturbation~\cite{defense-gan,pixeldefend,gen-robustness}. In this section, we elaborate on the implications of our work for DiffPure.

\textbf{\emph{Firstly, DiffPure is the defense that our work expects to avoid.}}
As we indicated in \Cref{sec:intro}, a thorough evaluation of stochastic pre-processing defenses typically requires significant modeling and computational efforts. DiffPure is a new example of such defenses --- it has a complicated solver of stochastic differential equations (SDE) and requires ``high-end NVIDIA GPUs with 32 GB of memory\footnote{\url{https://github.com/NVlabs/DiffPure}}.'' Our initial experiment shows that it takes several hours to attack even one batch of 8 CIFAR10 images on an Nvidia RTX 2080 Ti GPU with 11 GB of memory, and we received an out-of-memory error when attempting ImageNet with batch size 1. Because of these complications and computational costs, fully understanding its robustness requires substantially more effort than a previous stochastic pre-processing defense BaRT~\cite{bart}.

Given this challenging arms race between attacks and defenses, our work provides empirical and theoretical evidence to show that stochastic pre-processing defenses are fundamentally flawed. They cannot provide inherent robustness (like that from adversarial training) to prevent the existence of adversarial examples. Hence, future attacks may break it. As a result of these findings, future research should look for new ways of using randomness, such as those discussed in \Cref{sec:discussions}.

\textbf{\emph{Secondly, DiffPure matches our theoretical model.}}
DiffPure has two consecutive steps:
\begin{enumerate}
	\item Forward SDE adds noise to the image to decrease invariance like \Cref{eq:linear-defended}. The model becomes more robust because the input distribution is shifted.
	\item Reverse SDE removes noise from the image to recover invariance like \Cref{eq:linear-trained}. The model becomes less robust because the shifted input distribution is recovered.
\end{enumerate}
These two steps are consistent with our characterization of stochastic pre-processing defenses in \Cref{sec:tradeoff}. While our work mainly focuses on trained invariance (through model fine-tuning), an auxiliary denoiser (like Reverse SDE) can achieve a similar notion of invariance. Hence, we expect our arguments about the robustness-invariance trade-off to hold here as well.

\textbf{\emph{Finally, Our findings raise concerns with the way DiffPure claims to obtain robustness.}}
The above discussion finds no evident difference between DiffPure and our model in \Cref{sec:tradeoff}. When the Reverse SDE is perfect, we should achieve full invariance in \Cref{eq:linear-reduction} and expect no improved robustness --- attacking the whole procedure is equivalent to attacking the original model (if non-differentiable and randomized components are handled correctly). Hence, our findings raise concerns with the way DiffPure claims to obtain robustness.

Driven by the above concerns, we carefully review DiffPure’s evaluation and identify red flags:
\begin{enumerate}
	\item They only used 100 PGD steps and 20--30 EOT samples in AutoAttack~\cite{autopgd}. This setting is potentially inadequate based on our empirical results in \Cref{tab:previous-defenses}. Even breaking a less complicated defense requires far more steps and samples.
	\item Previous purification defenses cannot prevent adversarial examples on the manifold of their underlying generative model or denoiser~\cite{bpda}. However, DiffPure did not discuss this attack, i.e., whether it is possible to find an adversarial example of the diffusion model such that it remains adversarial (to the classifier) after the diffusion process. This strategy is different from its current evaluation, which attacks the whole pipeline with BPDA and EOT.
\end{enumerate}
These red flags suggest that there is still room for improving DiffPure’s evaluation.

\textbf{\emph{Summary.}} DiffPure matches our theoretical characterization of previous stochastic pre-processing defense. Thus, we expect our findings to hold here as well. Unfortunately, we cannot finish the evaluation of the above discussions due to their high computational costs. However, this challenge is exactly what our work aims to mitigate — we can identify concerns with the way robustness is achieved without needing to design adaptive attacks, and our findings have motivated us to identify red flags in their evaluation. We hope our work can increase the confidence of future research towards understanding the robustness of defenses sharing a similar assumption.

\subsection{Insights for Designing Attacks and Defenses Regarding Randomness}
\label{app:discuss:guidance}

While systematic guidance for designing defenses (and their attacks) remains an open question, we attempt to summarize some critical insights for this direction as follows.

\textbf{Guidance for Attacks.}
\begin{enumerate}
	\item Attackers aiming to evaluate defenses (i.e., not merely breaking them) should start with standard attacks before resorting to more involved attack strategies like EOT. This helps form a better understanding of the defense’s fundamental weakness.
	\item Stochastic pre-processors cannot provide inherent robustness, so an effective attack should exist. Although there has not been a systematic way to design or find such attacks, our work provides general guidelines to help with this task.
	\item Stochastic pre-processors provide robustness by invariance, so attackers can examine the model invariance to check the room for improvements.
\end{enumerate}

\textbf{Guidance for Defenses.}
\begin{enumerate}
	\item The current use of randomness is not promising. Defenses should decouple robustness and invariance; otherwise, future attacks may break them.
	\item Defenses should look for new ways of using randomness, such as those below or beyond the input space. Below-input randomness divides the input into orthogonal components, like modalities~\cite{singlesource} and independent patches~\cite{certifiedpatch}. Beyond-input randomness routes the input to separate components, like non-transferable models~\cite{trs}.
	\item Randomness should force the attack to target all possible (independent) subproblems, where the model performs well on each (independent and) non-transferable subproblem. In this case, defenses can decouple robustness and invariance, hence avoiding the pitfall of previous randomized defenses.
	\item Randomness alone does not provide robustness. Defenses must combine randomness with other inherently robust concepts to improve robustness.
\end{enumerate}

\subsection{Limitations and Potential Negative Societal Impacts}
\label{app:limitation}

Finally, we discuss the limitations and potential negative societal impacts of this work.

\paragraph{Limitations.}
This paper mainly focuses on stochastic \emph{pre-processing} defenses, thus we cannot comment on the effectiveness of stochastic defenses that are not based on input transformations. However, we do evaluate a few such defenses in \Cref{tab:previous-defenses} and observe similar results for our own interests, such as randomized activation pruning~\cite{activation-pruning} and discontinued activation~\cite{kwinners}. Given this observation, we believe our findings on stochastic pre-processing defenses are potentially generalizable to all stochastic defenses. We leave this exploration to future work.

Due to the limitation of computing resources, we are unable to evaluate the full BaRT~\cite{bart} defense on ImageNet. To mitigate this problem, we evaluate a subset of BaRT on the smaller ImageNette dataset. Since the primary objective of this work is to study the limitations of such defenses but not to break them, we believe the limitations that we observe on a subset of BaRT are reasonably generalizable to the full set of BaRT. Other work studying this defense made a similar choice~\cite{aggmopgd}.

We are unable to evaluate the parallel defense DiffPure~\cite{diffpure} due to its significantly high computational requirements. Given this limitation, we provide a thorough discussion in \Cref{app:discuss:diffpure} and explain that DiffPure is consistent with our model, hence we expect our findings to hold here as well.

\paragraph{Potential Negative Societal Impacts.}
This paper investigates the limitations of stochastic pre-processing defenses against adversarial examples. While the publication of this research may be used by attackers to create stronger attacks, we argue such considerations are out-weighted by the benefits of enabling defenders to understand the weaknesses of existing defenses. Moreover, our evaluation mainly involves existing attacks and previously broken defenses, thus we do not observe novel negative societal impacts. Our main objective is to uncover the fundamental weaknesses of such defenses, both empirically and theoretically, thereby raising the awareness of how to design proper stochastic defenses that avoid inadvertently weak evaluations and overestimated security.

\end{document}